\documentclass[opre,nonblindrev]{informs3}

\OneAndAHalfSpacedXI 

\usepackage{endnotes}
\let\footnote=\endnote
\let\enotesize=\normalsize
\def\notesname{Endnotes}%
\def\makeenmark{\hbox to1.275em{\theenmark.\enskip\hss}}
\def\enoteformat{\rightskip0pt\leftskip0pt\parindent=1.275em
  \leavevmode\llap{\makeenmark}}

\RequirePackage[numbers]{natbib}
\RequirePackage[colorlinks,citecolor=blue,linkcolor=blue,urlcolor=blue,hypertexnames=false]{hyperref}
\RequirePackage{graphicx}
\usepackage{float}
\usepackage{caption}
\usepackage{subcaption}
\usepackage{pdfpages}
\usepackage{xcolor}
\usepackage{booktabs}
\usepackage{algorithm}
\usepackage{algpseudocode}

\newcommand{\floor}[1]{\left\lfloor #1 \right\rfloor}
\newcommand\abs[1]{\left|#1\right|}
\def\indic#1{\mathbb{I}\left({#1}\right)}
\def\E{\mathbb{E}}
\def\P{\mathbb{P}}
\providecommand{\argmax}{\mathop\mathrm{arg max}}
\providecommand{\argmin}{\mathop\mathrm{arg min}}
\newcommand{\overbar}[1]{\mkern 1.5mu\overline{\mkern-1.5mu#1\mkern-1.5mu}\mkern 1.5mu}

\usepackage{natbib}
 \bibpunct[, ]{(}{)}{,}{a}{}{,}%
 \def\bibfont{\small}%
\TheoremsNumberedThrough     
\ECRepeatTheorems

\EquationsNumberedThrough    

\begin{document}


\RUNAUTHOR{Fan and Glynn}

\RUNTITLE{The Fragility of Optimized Bandit Algorithms}

\TITLE{The Fragility of Optimized Bandit Algorithms}

\ARTICLEAUTHORS{%
\AUTHOR{Lin Fan}
\AFF{Kellogg School of Management, Northwestern University, Evanston, IL 60208, \EMAIL{lin.fan@kellogg.northwestern.edu}} 
\AUTHOR{Peter W. Glynn}
\AFF{Department of Management Science and Engineering, Stanford University, Stanford, CA 94305, \EMAIL{glynn@stanford.edu}}
} 

\ABSTRACT{%
Much of the literature on optimal design of bandit algorithms is based on minimization of expected regret. 
It is well known that algorithms that are optimal over certain exponential families can achieve expected regret that grows logarithmically in the
number of trials, at a rate specified by the Lai-Robbins lower bound. 
In this paper, we show that when one uses such optimized algorithms, the resulting regret distribution necessarily has a very heavy tail, specifically, that of a truncated Cauchy distribution.
Furthermore, for $p>1$, the $p$’th moment of the regret distribution grows much faster than poly-logarithmically, in particular as a power of the total number of trials. We show that optimized UCB algorithms are also fragile in an additional sense, namely when the problem is even slightly mis-specified, the regret can grow much faster than the conventional theory suggests. 
Our arguments are based on standard change-of-measure ideas, and indicate that the most likely way that regret becomes larger than expected is when the optimal arm returns below-average rewards in the first few arm plays, thereby causing the algorithm to believe that the arm is sub-optimal.
To alleviate the fragility issues exposed, we show that UCB algorithms can be modified so as to ensure a desired degree of robustness to mis-specification.
In doing so, we also show a sharp trade-off between the amount of UCB exploration and the heaviness of the resulting regret distribution tail.
}%


\HISTORY{To appear in Operations Research (first version on arXiv: Sept 28, 2021)}

\maketitle
%



\section{Introduction} \label{introduction}

The multi-armed bandit (MAB) problem is a widely studied model that is both useful in practical applications and is a valuable theoretical paradigm exhibiting the exploration-exploitation trade-off that arises in sequential decision-making under uncertainty. More specifically, the goal in a MAB problem is to maximize the expected reward derived from playing, at each time step, one of $K$ bandit arms. Each arm has its own unknown reward distribution, so that playing a particular arm both provides information about that arm’s reward distribution (exploration) and provides an associated random reward (exploitation). One measure of the quality of a MAB algorithm is the (pseudo-)regret $R(T)$, which is essentially the number of times the sub-optimal arms are played over a time horizon $T$, as compared to an oracle that acts optimally with knowledge of the means of all arm reward distributions; a precise definition will be given in Section \ref{setup}.

There is an enormous literature on this problem, with much of the research having been focused on algorithms that attempt to minimize expected regret. In this regard, a fundamental result is the Lai-Robbins lower bound that establishes that the expected regret $\E[R(T)]$ grows logarithmically in $T$, with a multiplier that depends on the Kullback-Leibler (KL) divergences between the optimal arm and each of the sub-optimal arms; see \cite{lai_etal1985}. A predominant focus in the bandit literature is on designing algorithms that attain the Lai-Robbins lower bound over particular exponential families of distributions; see \cite{lai_etal1985} and \cite{burnetas_etal1996}. We call such algorithms \textit{optimized}. Among the many optimized algorithms in the literature, two prominent examples are the KL-upper confidence bound (KL-UCB) algorithm and Thompson sampling (TS); see \cite{cappe_etal2013} (and earlier work: \cite{garivier_etal2011}, \cite{maillard_etal2011}) for KL-UCB, and \cite{korda_etal2013} for TS (originally proposed by \cite{thompson_1933}).
Earlier optimized UCB-type algorithms can be found in, for example, \cite{lai_1987} and \cite{agrawal_1995}.

In this paper, we show that any optimized algorithm necessarily has the undesirable property that the tail of $R(T)$ is very heavy. In particular, because $\E[R(T)]$ is $O(\log(T))$ (where $O(a_T)$ is any sequence having the property that its absolute value is dominated by a constant multiple of $a_T$), Markov’s inequality implies that for $c > 0$, $\P(R(T) > c T) = O(\log(T)/T)$ as $T \to \infty$. One of our central results is a lower bound characterization of $\P(R(T) > c T)$ that roughly establishes that this probability is attained, namely it is roughly of order $T^{-1}$ for optimized algorithms. More precisely, our Theorem \ref{thm1} shows that optimized MAB algorithms automatically have the property that 
\begin{align*}
    \P(R(T) > x) \asymp \frac{1}{x}
\end{align*}
as $T \to \infty$, uniformly in $x$ with $T^a \le x \le c T$, for any $0 < a < 1$ and suitable $c > 0$. (We write $a_T \asymp b_T$ as $T \to \infty$ whenever $\log(a_T)/\log(b_T)$ converges to $1$ as $T \rightarrow \infty$.) In other words, the tail of the regret $R(T)$ looks, in logarithmic scale, like that of a \textit{truncated Cauchy distribution} (truncated due to the time horizon $T$). Thus, such algorithms fail to produce logarithmic regret with large probability, and when they fail to produce such regret, the magnitude of the regret can be very large. This is one sense in which bandit algorithms optimized for expected regret can be fragile.

An additional sense in which such optimized bandit algorithms are fragile is their sensitivity to model mis-specification. By this, we mean that if an algorithm has been optimized to attain the Lai-Robbins lower bound over a particular class of bandit environments (e.g., with the arm distributions belonging to a specific exponential family), then we can see much worse regret behavior when the environment presented to the algorithm does not belong to the class. For example, we show that for the KL-UCB algorithm designed for Gaussian environments with known and equal variances but unknown means, the expected regret for Gaussian environments can grow as a power $T^{r}$ when the variance of the optimal arm's rewards is larger than the variance built into the algorithm's design. In fact, $r$ can be made arbitrarily close to $1$ depending on how large the optimal arm's variance is, relative to the variance of the algorithm's design (Corollary \ref{cor1}). In other words, even when the mis-specification remains Gaussian, the expected regret can grow at a rate close to linear in the time horizon $T$.
Besides mis-specification of the bandits' marginal reward distributions, optimized algorithms are equally susceptible to mis-specification of the serial dependence structure of rewards. For example, expected regret deteriorates similarly as reward processes (e.g., evolving as Markov chains) become more autocorrelated (Corollary \ref{cor2}, Proposition \ref{markovian} and Example \ref{ex4}).

A final sense in which such optimized algorithms are fragile is that when one only slightly modifies the objective, the regret behavior of the algorithm can look much worse. In particular, suppose that we consider minimizing $\E[R(T)^p]$ for some $p > 1$, rather than $\E[R(T)]$. This objective would arise naturally, for example, in the presence of risk aversion to high regret. One might reasonably expect that algorithms optimized for $\E[R(T)]$ would have the property that $\E[R(T)^p]$ would then grow poly-logarithmically in $T$. However, the Cauchy-type tails discussed earlier imply that $(R(T)/\log(T))^p$ is not a uniformly integrable sequence. We show in Corollary \ref{cor3} that for optimized algorithms, $\E[R(T)^p]$ grows roughly at least as fast as $T^{p-1}$ as $T \to \infty$.

Our proofs rely on change-of-measure arguments that also provide insight into how algorithms optimized for expected regret can fail to identify the optimal arm, thereby generating large regret. For example, we show that conditional on large regret, the sample means of sub-optimal arms obey laws of large numbers that indicate that they continue to behave in their usual way; see Proposition \ref{prop6}. This suggests that the most likely way that large regret occurs for such optimized algorithms is when the optimal arm under-performs in the exploration phase at the start of the experiment, after which it is played infrequently, thereby generating large amounts of regret. This intuitive scenario has been heuristically considered several times in the literature (see, e.g., \cite{audibert_etal2009}), but this paper provides the theoretical justification for its central role in generating large regret.

To mitigate some of the fragility issues we expose, we show how to modify UCB algorithms so that their regret tails are lighter.
By suitably increasing the rate of UCB exploration, we can achieve any polynomial or exponential rate of tail decay; see Proposition \ref{prop5}.
For example, in well-specified settings, if one increases the nominal amount of exploration by a factor of $1+b$ times for any desired $b > 0$, then the tail of the resulting regret distribution will have an exponent of $-(1+b)$ (or less).
In particular, $\P(R(T) > x) \asymp x^{-(1+b)}$ as $T \to \infty$, uniformly in $x$ with $\log^a(T) \le x \le c T$, for any $a > 1$ and suitable $c > 0$.
By lightening the regret distribution tail to a given exponent, we also create a prescribed margin of safety against model mis-specification.
The modified UCB algorithm becomes more robust to mis-specification of the reward distribution (Corollary \ref{cor4}) and of the serial dependence structure of rewards (Corollary \ref{cor5}).
Of course, these benefits must come at the cost of greater expected regret.
We study (sharp) trade-offs between the regret tail and expected regret in Proposition \ref{prop5} and Theorem \ref{thm0}.

Our study of the tail of the regret distribution of MAB algorithms, and our uncovering of the above fragility phenomena, underscore the value of understanding the regret distribution beyond the expected regret performance measure that the literature focuses overwhelmingly on.
Despite the fundamental role of expected regret for sequential decision-making under uncertainty, as we show, important insights can be missed and severe fragility can result from optimizing for expected regret alone.
Our work thus provides a novel and useful complement to the by-now mature theory of expected regret in the bandit literature.
See also Section \ref{relatedwork} for recent related work concerning the regret distribution.

The rest of the paper is structured as follows. 
After discussing related work in Section \ref{relatedwork}, we introduce the setup for the paper in Section \ref{setup}.
In Section \ref{generalchar}, we establish our main result, Theorem \ref{thm1}, that optimized algorithms have regret distributions for which the tails are truncated Cauchy.
This result requires a technical condition (Definition \ref{A2}), which holds essentially for all continuous reward distributions.
To illustrate the key ideas behind Theorem \ref{thm1}, we prove a simplified version of the result in Section \ref{sketch}.
We develop in Section \ref{upperbound} tight upper bounds characterizing the regret tail for KL-UCB in settings where the regret tail is lighter than truncated Cauchy (because the condition in Definition \ref{A2} does not hold); see Theorem \ref{generalupperbound}.
In Section \ref{overview_fragility}, we discuss the connections between the heavy regret tails of optimized algorithms and their susceptibility to model mis-specification.
Afterwards, we show in Sections \ref{mis-specification} and \ref{dependent} that the performance of optimized algorithms can deteriorate sharply under the slightest amount of mis-specification of the distribution or the serial dependence structure of the rewards.
These insights make use of results from Section \ref{lowerbound}, where we establish general lower bounds for the regret tail of algorithms such as KL-UCB when the rewards come from stochastic processes; see Theorem \ref{gartnerellis}.
Moreover, we show in Section \ref{highermoments} that such optimized algorithms offer no control over the $p$'th moment of regret for any $p > 1$.
We extend the regret tail characterizations for exponential family models in Theorem \ref{thm1} to models with general reward distributions in Section \ref{generalmodels}.
In Section \ref{supportingresults}, we develop a trade-off in Theorem \ref{thm0} showing that lighter regret tails come at the cost of greater expected regret.
Our result significantly generalizes the Lai-Robbins lower bound for expected regret (as well as the Burnetas-Katehakis extension).
In Section \ref{simple}, building upon Section \ref{upperbound}, we discuss how to modify UCB algorithms to achieve any desired regret tail, with polynomial or exponential rates of decay, by suitably increasing the rate of exploration.
We then discuss how the modifications provide protection against mis-specification of the distribution of rewards and of the serial dependence structure of rewards in Sections \ref{robust1} and \ref{robust2}, respectively.
In Section \ref{numerical}, we examine some numerical experiments.
We conclude with the proofs of Theorems \ref{thm1}, \ref{generalupperbound}, \ref{gartnerellis} and \ref{thm0} in Appendices \ref{thm1_proof}, \ref{generalupperbound_proof}, \ref{gartnerellis_proof} and \ref{tradeoff_proof}, respectively.

\subsection{Related Work} \label{relatedwork}

In terms of related work, \cite{audibert_etal2009,salomon_etal2011} study concentration properties of the regret distribution.
In particular, \cite{audibert_etal2009} develop a finite-time upper bound on the tail of the regret distribution for a particular version of UCB in bounded reward settings.
Their upper bound has polynomial rates of tail decay, which are adjustable depending on algorithm settings.
One of their motivations for developing regret tail bounds is to establish a trade-off between the rate of exploration and the resulting heaviness of the regret tail.
However, it is lower bounds on the regret tail that are needed to conclusively establish the trade-off and confirm that the regret distribution is heavy-tailed.
Our lower bounds turn out to be frequently tight.

The regret distribution tail approximations developed in the current work are complementary to the strong laws of large numbers (SLLN's) and central limit theorems (CLT's) developed for bandit algorithms in instance-dependent settings in \cite{fan_etal2022}.
For example, in the Gaussian bandit setting (with unit variances for simplicity), for both TS and UCB, the regret satisfies the SLLN:
\begin{align*}
    \frac{R(T)}{\log(T)} \overset{\text{a.s.}}{\to} \sum_{i \ne i^*} \frac{2}{\Delta_i}
\end{align*}
and the CLT:
\begin{align*} 
    \frac{R(T) - \sum_{i \ne i^*} \frac{2}{\Delta_i} \log(T)}{\sqrt{\sum_{i \ne i^*} \frac{8}{\Delta_i^2} \log(T)}} \Rightarrow N(0,1), 
\end{align*}
where $\Delta_i > 0$ is the difference between the mean of the optimal arm $i^*$ and that of sub-optimal arm $i$, and $\Rightarrow$ denotes convergence in distribution.
These results can be viewed as describing the typical behavior and fluctuation of regret when $T$ is large.
This stands in contrast to the results in the current work, which describe the tail behavior of the regret.
Tails are generally affected by atypical behavior.
As noted above, our arguments show that the regret tail is impacted by trajectories on which the algorithm mis-identifies the optimal arm.
The mean and the variance in the CLT both scale as $\log(T)$ with the time horizon $T$. By analogy with the large deviations theory for sums of independent, identically distributed (iid) random variables, this suggests that large deviations of regret correspond to deviations from the expected regret that are of order $\log(T)$. We characterize the tail of the regret beyond $\log^{1+\gamma}(T)$ for small $\gamma > 0$, and we save the analysis of deviations on the $\log(T)$ scale for future work.

The regret distribution of MAB algorithms has also been studied in an asymptotic regime different from the Lai and Robbins regime that this paper and \cite{fan_etal2022} focus on.
\cite{kuang_etal2021}, \cite{fan_etal2021} and \cite{kalvit_etal2021} obtain diffusion approximations for the regret distribution of MAB algorithms (including for TS and UCB) in the setting where the gap between arm means scales like $1/\sqrt{T}$, with the time horizon $T \to \infty$.

Recently, \cite{ashutosh_etal2021} showed that for an algorithm to achieve expected regret growing logarithmically in the time horizon across a collection of environment instances, the distributional class of arm rewards cannot be too large.
For example, if the rewards are known to be sub-Gaussian, then an upper bound restriction on the variance proxy is required.
They conclude that if such a restriction is mis-specified, then the worst case (for some environment instance) expected regret could grow polynomially in the time horizon.
Their result provides no information about algorithm behavior for any particular environment instance, nor does it cover narrower classes of distributions (e.g., Gaussian).

Our results on model mis-specification are significantly differentiated from other results on model mis-specification in the bandit literature.
We consider the effect of mis-specifying, for example, the distribution or serial dependence structure of rewards, on the (frequentist) regret tail (and as a direct consequence, also the expected regret).
From the perspective of Bayesian expected regret, \cite{simchowitz_etal2021} studies mis-specification of the prior distribution used by algorithms, and \cite{liu_etal2022} studies mis-specified Bernoulli bandits for algorithms based on Gaussian prior and likelihood structure.
From the perspective of frequentist expected regret, existing literature, including \cite{ghosh_etal2017}, \cite{lattimore_generative_etal2020}, \cite{foster_etal2020}, \cite{takemura_etal2021} and \cite{krishnamurthy_etal2021}, study the effect of mis-specifying a linear regression structure when the true regression function is not exactly linear.

There is also a growing literature on risk-averse formulations of the MAB problem, with a non-comprehensive list being: \cite{sani_etal2012, maillard_2013, galichet_etal2013, zimin_etal2014, szorenyi_etal2015, vakili_etal2016, cassel_etal2018, tamkin_etal2019, szorenyi_etal2020, prashanth_etal2020, baudry_etal2021, khajonchotpanya_etal2021}. As noted earlier, risk-averse formulations involve defining arm optimality using criteria other than the expected reward. These papers consider mean/variance criteria, value-at-risk, or conditional value-at-risk measures, and develop algorithms which achieve good (or even optimal in some cases) regret performance relative to their chosen criterion. Our results serve as motivation for these papers, and highlight the need to consider robustness in many MAB problem settings.
We believe it would be interesting to investigate in follow-up studies the fragility issues exposed in our paper through the lens of these risk-averse MAB formulations.

\section{Model and Preliminaries} \label{setup}

\subsection{The Multi-armed Bandit Framework} \label{mabintro}
A $K$-armed MAB evolves within a bandit environment $\nu = (Q_1,\dots,Q_K)$, where each $Q_i$ is a distribution on $\mathbb{R}$.
At time $t$, the decision-maker selects an arm $A(t) \in [K] := \{1,\dots,K\}$ to play.
Upon selecting the arm $A(t)$, a reward $Y(t)$ from arm $A(t)$ is received as feedback.
The conditional distribution of $A(t)$ given $A(1),Y(1),\dots,A(t-1),Y(t-1)$ is $\pi_t(\cdot \mid A(1),Y(1),\dots,A(t-1),Y(t-1))$, where $\pi = (\pi_t, t \ge 1)$ is a sequence of probability kernels, which constitutes the bandit algorithm.
For each $t \ge 1$, $\pi_t$ is a probability kernel from $([K] \times \mathbb{R})^{t-1}$ to $[K]$. 
The conditional distribution of $Y(t)$ given $A(1),Y(1),\dots,A(t-1),Y(t-1),A(t)$ is $Q_{A(t)}(\cdot)$.
We write $X_i(s)$ to denote the reward received when arm $i$ is played for instance $s$, so that $Y(t) = X_{A(t)}(N_{A(t)}(t))$, where $N_i(t) = \sum_{s=1}^t \indic{A(s) = i}$ denotes the number of plays of arm $i$ up to and including time $t$.

For any time $t$, the interaction between the algorithm $\pi$ and the environment $\nu$ induces a unique probability $\P_{\nu \pi}(\cdot)$ on $([K] \times \mathbb{R})^\infty$ for which
\begin{align*}
    \P_{\nu \pi}(A(1) = a_1, Y(1) \in dy_1,\dots, A(t) = a_t, Y(t) \in dy_t) = \prod_{s=1}^t \pi_s(a_s \; | \; a_1,y_1,\dots,a_{s-1},y_{s-1}) Q_{a_s}(dy_s). 
\end{align*}
Here, $dy$ denotes an infinitesimal set containing $y$.
For $t \ge 1$, we write $\E_{\nu \pi}[\cdot]$ to denote the expectation associated with $\P_{\nu \pi}(\cdot)$.

The quality of an algorithm $\pi$ operating in an environment $\nu = (Q_1,\dots,Q_K)$ is measured by the (pseudo-)regret (at time $T$):
\begin{align*}
    R(T) = \sum_{i=1}^K N_i(T)\Delta_i,
\end{align*}
where $\Delta_i = \mu_*(\nu) - \mu(Q_i)$ and $\mu_*(\nu) = \max_{Q \in \nu} \mu(Q)$.
(For any distribution $Q$, we use $\mu(Q)$ to denote its mean.)
An arm $i$ is called optimal if $\Delta_i = 0$, and sub-optimal if $\Delta_i > 0$.
The goal in most settings is to find an algorithm $\pi$ which minimizes the expected regret $\E_{\nu \pi}[R(T)]$, i.e., plays the optimal arm(s) as often as possible in expectation.

When discussing the regret distribution tail in multi-armed settings, we will often reference (for any given environment) the $i$-th-best arm (with the $i$-th largest mean).
For each $i=1,\dots,K$, we will use $r(i) \in [K]$ to denote the index/label of the $i$-th-best arm.
To keep our discussions and derivations streamlined, unless specified otherwise, throughout the paper we will only consider environments where for each $i = 1,\dots,K$, the $i$-th-best arm is unique.

\subsection{Optimized Algorithms} \label{optimized}

In order to discuss optimized algorithms, we consider arm reward distributions from a one-dimensional exponential family, parameterized by mean, of the form:
\begin{align}
P^z(dx) = \exp\bigl( \theta_P(z) \cdot x - \Lambda_P(\theta_P(z)) \bigr) P(dx), \quad z \in \mathcal{I}_P. \label{model}
\end{align}
Here, $P$ is a base distribution with cumulant generating function (CGF) $\Lambda_P$. We use $\mathcal{I}_P$ to denote the set of all possible means for distributions $P^z$ of the form in (\ref{model}), with $\theta_P(z)$ being any real number in the set $\Theta_P = \{\theta \in \mathbb{R} : \Lambda_P(\theta) < \infty\}$.
Moreover, for each $z \in \mathcal{I}_P$, we use $\theta_P(z)$ to denote the unique value for which $\mu(P^z) = z$.
(Also recall that $\Lambda_P'(\theta_P(z)) = z$.)
Throughout the paper, we will always work with base distributions $P$ such that $\Theta_P$ contains a neighborhood of zero.

For a base distribution $P$, we denote the mean-parameterized model in (\ref{model}) via:
\begin{align}
    \mathcal{M}_P = \left\{ P^z : z \in \mathcal{I}_P \right\}, \label{modelp}
\end{align}
which induces a class $\mathcal{M}_P^K$ of $K$-armed bandit environments, where each environment consists of a $K$-tuple of distributions from $\mathcal{M}_P$.
The KL divergence between distributions in $\mathcal{M}_P$ with means $z_1, z_2 \in \mathcal{I}_P$ is denoted by $d_P(z_1,z_2)$, and can be expressed as:
\begin{align}
d_P(z_1,z_2) 
& = \int \log \frac{dP^{z_1}}{dP^{z_2}}(x) \; P^{z_1}(dx) \nonumber \\
& = \Lambda_P(\theta_P(z_2)) - \Lambda_P(\theta_P(z_1)) - \Lambda_P'(\theta_P(z_1)) \cdot \bigl(\theta_P(z_2) - \theta_P(z_1)\bigr), \label{bregman}
\end{align}
where $dP^{z_1}/dP^{z_2}$ denotes the likelihood ratio of $P^{z_1}$ to $P^{z_2}$.

From the seminal work of \cite{lai_etal1985}, there is a precise characterization of the minimum possible growth rate of expected regret for an algorithm $\pi$ designed for $\mathcal{M}_P^K$.
We start with the notion of \textit{consistency} in Definition \ref{A00} below, which restricts the types of algorithms considered in order to formulate a theory of optimality.
This notion rules out unnatural algorithms that over-specialize and perform very well in particular environment instances within a class, but very poorly in other instances.
\begin{definition}[$\mathcal{M}_P$-Consistent Algorithm] \label{A00} 
An algorithm $\pi$ is $\mathcal{M}_P$-consistent if for any $a > 0$, any environment $\nu \in \mathcal{M}_P^K$, and each sub-optimal arm $i$:
\begin{align}
    \lim_{T \to \infty} \frac{\E_{\nu \pi}[N_i(T)]}{T^a} = 0. \label{lairobbins_consistency}
\end{align}
\end{definition}
The Lai-Robbins lower bound is then formulated for consistent algorithms.
In particular, for any $\mathcal{M}_P$-consistent algorithm $\pi$, any environment $\nu = (P^{\mu_1},\dots,P^{\mu_K}) \in \mathcal{M}_P^K$ and each sub-optimal arm $i$,
\begin{align}
    \liminf_{T \to \infty} \frac{\E_{\nu \pi}[N_i(T)]}{\log(T)} \ge \frac{1}{d_P(\mu_i,\mu_*(\nu))}. \label{lairobbins}
\end{align}
We say that an $\mathcal{M}_P$-consistent algorithm $\pi$ is \textit{$\mathcal{M}_P$-optimized} if the lower bound in (\ref{lairobbins}) is achieved, as in the following Definition \ref{A1}.

\begin{definition}[$\mathcal{M}_P$-Optimized Algorithm] \label{A1} 
An algorithm $\pi$ is $\mathcal{M}_P$-optimized if for any environment $\nu = (P^{\mu_1},\dots,P^{\mu_K}) \in \mathcal{M}_P^K$ and each sub-optimal arm $i$,
\begin{align*}
\lim_{T \to \infty} \frac{\E_{\nu \pi}[N_i(T)]}{\log(T)} = \frac{1}{d_P(\mu_i,\mu_*(\nu))}.
\end{align*}
\end{definition}

\section{Characterization of the Regret Tail} \label{main}

\subsection{Truncated Cauchy Tails} \label{generalchar}

In this section, we show that for many classes of exponential family bandit environments, the tail of the regret distribution of optimized algorithms is essentially that of a truncated Cauchy distribution.
Moreover, for such classes, the tail is truncated Cauchy for \textit{every environment} within the class.
This is established in Theorem \ref{thm1}.
As we will see, this truncated Cauchy tail property always holds when the exponential family is continuous with left tails that are lighter than exponential (possessing CGF's that are finite on the negative half of the real line).
When the exponential family is discrete or has exponential left tails, the regret distribution tail is generally lighter than truncated Cauchy, but still heavy and decaying at polynomial rates.

As discussed in the Introduction, the regret tail characterization that we develop here reveals several important insights about the fragility of optimized bandit algorithms.
For example, when the regret tail is truncated Cauchy, as is generally the case for continuous exponential families, the slightest degree of mis-specification of the marginal distribution (see Section \ref{mis-specification}) or serial dependence structure (see Section \ref{dependent}) of arm rewards can cause optimized algorithms to lose the basic consistency property and suffer expected regret growing polynomially in the time horizon.
Moreover, in such settings there is no control over any higher moment of the regret beyond the first moment (see Section \ref{highermoments}).
It is furthermore striking that every environment instance within such classes of bandit environments suffers from these fragility issues, not just some worst case instances within such classes.

Theorem \ref{thm1} relies in part on the notion of \textit{discrimination equivalence}, as stated in Definition \ref{A2} below.
This property can be readily verified from (\ref{bregman}).
Following the statement of the theorem, we will provide an easier-to-verify equivalent characterization (Lemma \ref{lem1}) as well as simple sufficient conditions for this property (Propositions \ref{prop3} and \ref{prop4}).
We will then explain the choice of terminology, ``discrimination equivalence'', and provide examples for intuition. 
\begin{definition}[Discrimination Equivalence] \label{A2}
A distribution $P$ is \textit{discrimination equivalent} if for any $z_1,z_2 \in \mathcal{I}_P$ with $z_1 > z_2$,
\begin{align}
    \inf_{z \in \mathcal{I}_P \, : \, z < z_2} \frac{d_P(z,z_1)}{d_P(z,z_2)} = 1. \label{klratio}
\end{align}
\end{definition}

For an algorithm $\pi$ operating in an environment $\nu$, we say that the resulting distribution of regret $R(T)$ has a \textit{tail exponent} of $-c$ if $\P_{\nu \pi}(R(T) > x) \asymp x^{-c}$ as $T \to \infty$, uniformly in $x$ with $T^a \le x \le a' T$, for any $0 < a < 1$ and suitable $a' > 0$.
Intuitively, the regret tail exponent is determined by the tail exponent of the distribution of $N_{r(2)}(T)$, the number of plays of the second-best arm $r(2)$.
(So it suffices to consider the tail exponent of $N_{r(2)}(T)$ when discussing the regret tail exponent.)
In Theorem \ref{thm1}, (\ref{conclusion0}) and (\ref{conclusion1}) are reflective of this intuition, since achieving logarithmic expected regret means the regret tail exponent cannot be greater than $-1$.
(See Theorem \ref{generalupperbound} in Section \ref{upperbound}, where we fully establish this intuition by specializing the analysis from general optimized algorithms to the KL-UCB algorithm.)

The full proof of Theorem \ref{thm1} is given in Appendix \ref{thm1_proof}. 
In Section \ref{sketch}, we prove a simplified version of Theorem \ref{thm1}, along with a discussion to highlight the intuition behind this result.
Through simulation studies (see Figures \ref{figure1}-\ref{figure3} in Section \ref{numerical}), we verify that the result provides accurate approximations over reasonably short time horizons.
\begin{theorem} \label{thm1}
Let $\pi$ be $\mathcal{M}_P$-optimized.
Then for any environment $\nu = (P^{\mu_1},\dots,P^{\mu_K}) \in \mathcal{M}_P^K$ and the $i$-th-best arm $r(i)$,
\begin{align}
\liminf_{T \to \infty} \inf_{x \in B_\gamma(T)} \frac{\log \P_{\nu \pi}(N_{r(i)}(T) > x)}{\log(x)} \ge - \sum_{j = 1}^{i-1} \inf_{z \in \mathcal{I}_P \, : \, z < \mu_{r(i)}} \frac{d_P(z,\mu_{r(j)})}{d_P(z,\mu_{r(i)})}, \label{conclusion0}
\end{align}
with $B_\gamma(T) = [\log^{1+\gamma}(T), (1-\gamma)T]$ and any $\gamma \in (0,1)$. \\
If in addition, $P$ is discrimination equivalent, then for the second-best arm $r(2)$,
\begin{align}
\lim_{T \to \infty} \frac{\log \P_{\nu \pi}(N_{r(2)}(T) > x)}{\log(x)} = -1 \label{conclusion1}
\end{align}
uniformly for $x \in [T^\gamma, (1-\gamma)T]$ for any $\gamma \in (0,1)$ as $T \to \infty$. 
Moreover, for $i \ge 3$, (\ref{conclusion0}) holds with the right side equal to $-(i-1)$.
\end{theorem}

In Lemma \ref{lem1} below, we provide an equivalent characterization of discrimination equivalence. 
This characterization implies that each summand on the right side of (\ref{conclusion0}) is equal to $-1$.
(Note that for any $z,z_1,z_2 \in \mathcal{I}_P$ with $z < z_2 < z_1$, we always have $d_P(z,z_1)/d_P(z,z_2) \ge 1$.)
In light of the exact $-1$ tail exponent for the second-best arm $r(2)$ in (\ref{conclusion1}), we might conjecture that the lower bounds in (\ref{conclusion0}) are tight in general without discrimination equivalence.
We will rigorously establish this fact for a particular choice of algorithm (KL-UCB) in Section \ref{upperbound}.
The proof of Lemma \ref{lem1} is given in Appendix \ref{generalcharproofs}.
\begin{lemma} \label{lem1}
$P$ is discrimination equivalent if and only if $\; \inf \Theta_P = -\infty$ and
\begin{align}
    \lim_{\theta \to -\infty} \theta \Lambda_P'(\theta) - \Lambda_P(\theta) = \infty. \label{equivalentcondition}
\end{align}
((\ref{equivalentcondition}) can also be expressed as: $\lim_{\theta \to -\infty} \Lambda_P^*(\Lambda_P'(\theta)) = \infty$, where $\Lambda_P^*$ is the convex conjugate of $\Lambda_P$.)
\end{lemma}

In Proposition \ref{prop3}, we give a simple sufficient condition for discrimination equivalence that applies to reward distributions with support that is unbounded to the left on the real line.
The requirement is that the CGF of the distribution is finite on the negative half of the real line.
In Proposition \ref{prop4}, we provide simple conditions to determine whether or not discrimination equivalence holds for distributions with support that is bounded to the left on the real line.
When the support is bounded to the left, discrimination equivalence holds for continuous distributions, but generally not for discrete distributions.
The proofs of Propositions \ref{prop3} and \ref{prop4} can be found in Appendix \ref{generalcharproofs}. 

\begin{proposition} \label{prop3}
If the support of $P$ is unbounded to the left, and $\inf \Theta_P = -\infty$, then $P$ is discrimination equivalent.
\end{proposition}

\begin{proposition} \label{prop4}
If the support of $P$ is bounded to the left with no point mass at the infimum of the support, then $P$ is discrimination equivalent.
But if there is a positive point mass at the infimum of the support, then $P$ is not discrimination equivalent. 
\end{proposition}

It can be verified that for fixed $z_1 > z_2$,
\begin{align}
    \inf_{z \in \mathcal{I}_P \, : \, z < z_2} \frac{d_P(z,z_1)}{d_P(z,z_2)} = \lim_{z \downarrow \inf \mathcal{I}_P} \frac{d_P(z,z_1)}{d_P(z,z_2)}. \label{klequivalence}
\end{align}
The KL divergence $d_P(z,z')$ can be thought of as the mean information for discriminating between $P^z$ and $P^{z'}$, given a sample from $P^z$.
Since $d_P(z,z_1) = d_P(z,z_2)$ if and only if $z_1 = z_2$, the ratio $d_P(z,z_1)/d_P(z,z_2)$ can be thought of as a measure of the difficulty of discriminating between $P^z$ and $P^{z_1}$ relative to that between $P^z$ and $P^{z_2}$, given a sample from $P^z$ in both cases.
The more similar the difficulty in the two cases, the closer the ratio is to one. 
In such cases, as suggested by Theorem \ref{thm1}, the regret tail will be heavier/closer to being truncated Cauchy.
(In Theorem \ref{generalupperbound} in Section \ref{upperbound}, we provide matching upper bounds for (\ref{conclusion0}) for the KL-UCB algorithm, thereby providing validation for this way of thinking.)
With this interpretation, we review in the following examples some of the settings covered by Propositions \ref{prop3} and \ref{prop4} above.
We provide derivations for these examples at the end of Appendix \ref{generalcharproofs}.

\begin{example} \label{ex0}
Suppose in (\ref{model}) that the base distribution $P$ is the Gaussian distribution with mean $0$ and variance $\sigma^2$.
Then,
\begin{align*}
    d_P(z,z') = \frac{(z - z')^2}{2\sigma^2}.
\end{align*}
Hence, in this setting, (\ref{klequivalence}) is always equal to $1$, and $P$ is discrimination equivalent.
\end{example}

\begin{example} \label{ex1}
Suppose in (\ref{model}) that the base distribution $P$ is the uniform distribution on $[0,1]$.
Then, the CGF is:
\begin{align*}
    \Lambda_P(\theta) = \log\left( \frac{e^\theta - 1}{\theta} \right).
\end{align*}
It can be verified from the identity (\ref{bregman}) that in this setting, (\ref{klequivalence}) is always equal to $1$, and so $P$ is discrimination equivalent.
\end{example}

\begin{example} \label{ex2}
Suppose in (\ref{model}) that the base distribution $P$ is the Bernoulli distribution with mean $1/2$.
It can be verified from the identity (\ref{bregman}) that in this setting,
\begin{align*}
    \lim_{z \downarrow 0} \frac{d_P(z,z_1)}{d_P(z,z_2)} = \frac{\log(1-z_1)}{\log(1-z_2)}.
\end{align*}
Hence, in this setting, (\ref{klequivalence}) is always strictly greater than $1$ for $0 < z_2 < z_1 < 1$, and so $P$ is not discrimination equivalent.

A similar behavior arises whenever $P$ puts positive mass at the left endpoint of its support, which we denote by $L$.
From the perspective of the distribution $P^{z}$ (which becomes a unit point mass at $L$ as $z \downarrow L$), the different point masses at $L$ associated with $P^{z_1}$ and $P^{z_2}$ can be discriminated at different rates.
Hence, in such settings, $P$ is not discrimination equivalent.
\end{example}

\begin{example} \label{ex3}
Suppose in (\ref{model}) that the base distribution $P(dx) = e^x \cdot \indic{x \le 0} dx$ for $x \in \mathbb{R}$, so $P$ is a negatively supported exponential distribution, and $\Theta_P = (-1,\infty)$.
It can be verified from the identity (\ref{bregman}) that 
\begin{align*}
    \lim_{z \to -\infty} \frac{d_P(z,z_1)}{d_P(z,z_2)} = \frac{z_2}{z_1}.
\end{align*}
Hence, in this setting, (\ref{klequivalence}) is always strictly greater than $1$ for $z_2 < z_1 < 0$, and so $P$ is not discrimination equivalent.
Intuitively, this behavior arises because $P^{z_1}$ is a scale change of $P^{z_2}$ (as opposed to a location change, as in the setting of Example \ref{ex0}).
So the ability to discriminate from the perspective of $P^{z}$ (as $z \to -\infty$), differs in the two cases, regardless of how negative $z$ is.
\end{example}

As noted earlier, Theorem \ref{thm1} establishes under $P$-discrimination equivalence that the regret tail of an $\mathcal{M}_P$-optimized algorithm is truncated Cauchy for every environment in $\mathcal{M}_P^K$.
However, regardless of whether or not discrimination equivalence holds, there always exist some environments for which the regret tail of optimized algorithms is arbitrarily close to being truncated Cauchy (with a tail exponent arbitrarily close to $-1$).
This is the content of Corollary \ref{cor0} below, which follows immediately from (\ref{conclusion0}) in Theorem \ref{thm1} by taking the difference $\mu_{r(1)} - \mu_{r(2)}$ to be sufficiently small and using the relevant continuity property of the ratio of KL divergences on the right side of (\ref{conclusion0}).
This result highlights a universal fragility property of algorithms optimized for any exponential family class of environments.
However, compared to the fragility implications from Theorem \ref{thm1} which pertain to \textit{all environment instances} within a class, Corollary \ref{cor0} is weaker as it pertains only to \textit{some environment instances} within a class.

\begin{corollary} \label{cor0}
Let $\pi$ be $\mathcal{M}_P$-optimized.
Then for any $\epsilon > 0$, there exists $\delta > 0$ such that for any environment $\nu = (P^{\mu_1},\dots,P^{\mu_K}) \in \mathcal{M}_P^K$ with $0 < \mu_{r(1)} - \mu_{r(2)} < \delta$,
\begin{align*}
\liminf_{T \to \infty} \inf_{x \in B_\gamma(T)} \frac{\log \P_{\nu \pi}(N_{r(2)}(T) > x)}{\log(x)} \ge -(1+\epsilon),
\end{align*}
with $B_\gamma(T) = [\log^{1+\gamma}(T), (1-\gamma)T]$ and any $\gamma \in (0,1)$.
\end{corollary}

\subsection{Key Ideas Behind Theorem \ref{thm1} and Further Results} \label{sketch}

Below we provide a proof of a simplified version of Theorem \ref{thm1}, focusing on the two-armed bandit setting.
As we will see, the key idea behind our proof is a change of measure argument in which the reward distribution of the optimal arm is tilted so that its mean becomes less than that of the sub-optimal arm.
Then, within the new environment resulting from the change of measure, we require control over the number of plays of the new sub-optimal arm.
Lemma \ref{lemma2} below provides such control through a weak law of large numbers (WLLN) for the number of sub-optimal arm plays of optimized algorithms.
Lemma \ref{lemma2} follows immediately for optimized algorithms due to a ``one-sided'' WLLN in Theorem \ref{thm0} in Section \ref{supportingresults}, which is developed for completely general (possibly non-exponential family) models.
Moreover, Lemma \ref{lemma_klinf_wlln} from Section \ref{generalmodels} extends Lemma \ref{lemma2} to such general models.
For further discussion, see Remark \ref{remark99} in Section \ref{supportingresults}.
\begin{lemma} \label{lemma2}
Let $\pi$ be $\mathcal{M}_P$-optimized.
Then for any environment $\nu = (P^{\mu_1},\dots,P^{\mu_K}) \in \mathcal{M}_P^K$ and each sub-optimal arm $i$,
\begin{align*}
    \frac{N_i(T)}{\log(T)} \to \frac{1}{d_P(\mu_i,\mu_*(\nu))}
\end{align*}
in $\P_{\nu \pi}$-probability as $T \to \infty$.
\end{lemma}

We will show the following simplified version of Theorem \ref{thm1}. 
Let $a \in (0,1)$, and $\nu = (P^{\mu_1},P^{\mu_2}) \in \mathcal{M}_P^2$ such that (without loss of generality) $\mu_1 > \mu_2$, i.e., arm 1 is optimal in $\nu$.
For any $\mathcal{M}_P$-optimized algorithm $\pi$, we will first obtain:
\begin{align}
    \liminf_{T \to \infty} \frac{\log \P_{\nu \pi}(N_2(T) > a T)}{\log(T)} \ge - \inf_{z \in \mathcal{I}_P \, : \, z < \mu_2} \frac{d_P(z,\mu_1)}{d_P(z,\mu_2)}. \label{sketch_liminf}
\end{align}
If additionally $P$ is discrimination equivalent, then
\begin{align}
    \lim_{T \to \infty} \frac{\log \P_{\nu \pi}(N_2(T) > a T)}{\log(T)} = -1. \label{sketch_lim}
\end{align}
\proof{Proof for (\ref{sketch_liminf}) and (\ref{sketch_lim}).}
To obtain (\ref{sketch_liminf}), consider a new environment $\widetilde{\nu} = (P^{\widetilde{\mu}_1},P^{\mu_2}) \in \mathcal{M}_P^2$ with $\widetilde{\mu}_1 < \mu_2$, i.e., arm $1$ is sub-optimal in $\widetilde{\nu}$.
By a change of measure from $\nu$ to $\widetilde{\nu}$,
\begin{align}
    \P_{\nu \pi}\left( N_2(T) > a T \right) & = \E_{\widetilde{\nu} \pi}\Biggl[ \indic{N_2(T) > a T} \underbrace{\prod_{t = 1}^{N_1(T)} \frac{dP^{\mu_1}}{dP^{\widetilde{\mu}_1}}(X_1(t))}_{\textstyle \text{\small $:= L_T(\mu_1,\widetilde{\mu}_1)$}} \Biggr]. \label{com111}
\end{align}
Note that
\begin{align*}
    \log L_T(\mu_1,\widetilde{\mu}_1) = N_1(T) \cdot \frac{1}{N_1(T)} \sum_{t=1}^{N_1(T)} \log \frac{dP^{\mu_1}}{dP^{\widetilde{\mu}_1}}(X_1(t)).
\end{align*}
Under $\widetilde{\nu}$, by Lemma \ref{lemma2},
\begin{align}
     \frac{N_1(T)}{\log(T)} \to \frac{1}{d_P(\widetilde{\mu}_1,\mu_2)} \label{com1}
\end{align}
in $\P_{\widetilde{\nu} \pi}$-probability as $T \to \infty$.
Under $\widetilde{\nu}$, by (\ref{com1}) and the WLLN,
\begin{align}
    \frac{1}{N_1(T)} \sum_{t=1}^{N_1(T)} \log \frac{dP^{\mu_1}}{dP^{\widetilde{\mu}_1}}(X_1(t)) \to -d_P(\widetilde{\mu}_1,\mu_1) \label{com11}
\end{align}
in $\P_{\widetilde{\nu} \pi}$-probability as $T \to \infty$.
The WLLN's (\ref{com1}) and (\ref{com11}) then imply that for $\epsilon > 0$,
\begin{align}
    \log L_T(\mu_1,\widetilde{\mu}_1) \ge -(1+\epsilon) \frac{d_P(\widetilde{\mu}_1,\mu_1)}{d_P(\widetilde{\mu}_1,\mu_2)} \log(T) \label{com1111}
\end{align}
with $\P_{\widetilde{\nu} \pi}$-probability converging to 1 as $T \to \infty$.
Since (under $\widetilde{\nu}$) $\P_{\widetilde{\nu} \pi}(N_2(T) > a T) \to 1$, using (\ref{com111}) and (\ref{com1111}), we obtain:
\begin{align}
    \liminf_{T \to \infty} \frac{\log \P_{\nu \pi}(N_2(T) > a T)}{\log(T)} \ge - \frac{d_P(\widetilde{\mu}_1,\mu_1)}{d_P(\widetilde{\mu}_1,\mu_2)}. \label{com11111}
\end{align}
Note that $\widetilde{\mu}_1$ is a free variable that we can optimize over, subject to the constraints: $\widetilde{\mu}_1 < \mu_2$ and $\widetilde{\mu}_1 \in \mathcal{I}_P$.
Doing so yields (\ref{sketch_liminf}).
The right side of (\ref{sketch_liminf}) equals $-1$ if $P$ is discrimination equivalent.
As noted in the Introduction, for an $\mathcal{M}_P$-optimized algorithm $\pi$,
\begin{align*}
    \limsup_{T \to \infty} \frac{\log \P_{\nu \pi}(N_2(T) > a T)}{\log(T)} \le -1. 
\end{align*}
So if $\pi$ is $\mathcal{M}_P$-optimized and $P$ is discrimination equivalent, we obtain (\ref{sketch_lim}).
\halmos 
\endproof

To obtain (\ref{sketch_liminf}), the ``optimal'' change of measure from $\nu$ to $\widetilde{\nu}$ in (\ref{com111}) essentially involves sending $\widetilde{\mu}_1 \downarrow \inf \mathcal{I}_P$, which can be quite extreme.
For example, under the conditions of Proposition \ref{prop3}, $\inf \mathcal{I}_P = -\infty$ and the optimal change of measure would involve sending the optimal arm $1$ mean $\widetilde{\mu}_1 \to -\infty$.
This suggests that the primary way that large regret arises is when the mean of the optimal arm $1$ is under-estimated to be below that of the sub-optimal arm $2$, likely due to receiving some unlucky rewards early on in the bandit experiment.
Arm $1$ is then mis-labeled as sub-optimal, and the mis-labeling is not corrected for a long time, resulting in large regret.

To obtain, for example, a regret of $O(T)$ when the optimal arm $1$ is mis-labeled as sub-optimal, there effectively needs to be $O(\log(T))$ unusually low rewards from arm $1$.
The probability of such a scenario is exponential in the number of arm $1$ plays.
So the probability decays as an inverse power of $T$.

One might also consider a different change of measure, where the distribution of the sub-optimal arm $2$ is tilted so that its mean is above that of the optimal arm $1$.
This corresponds to the scenario where the mean of arm $2$ is over-estimated to be above that of arm $1$, and so arm $2$ is mis-labeled as optimal.

To obtain, for example, a regret of $O(T)$ when the sub-optimal arm $2$ is mis-labeled as optimal, there effectively needs to be $O(T)$ unusually high rewards from arm $2$.
The probability of such a scenario is exponential in the number of arm $2$ plays.
So the probability decays exponentially with $T$.

To accompany Theorem \ref{thm1}, we show in Proposition \ref{prop6} that large regret is not due to over-estimation of sub-optimal arm means, but must therefore be due to under-estimation of the optimal arm mean.
The proof of Proposition \ref{prop6} is given in Appendix \ref{sketchproofs}.
(Here, we use $\widehat{\mu}_i(t) = \frac{1}{N_i(t)} \sum_{s=1}^{N_i(t)} X_i(s)$ to denote the sample mean of arm $i$ rewards up to time $t$.)

\begin{proposition} \label{prop6}
Let $\pi$ be $\mathcal{M}_P$-optimized.
Then for any environment $\nu = (P^{\mu_1},\dots,P^{\mu_K}) \in \mathcal{M}_P^K$, any sub-optimal arm $i$, and any $\epsilon > 0$,
\begin{align*}
    \lim_{T \to \infty} \P_{\nu \pi}\left( \abs{\widehat{\mu}_i(T) - \mu_i} \le \epsilon \mid N_i(T) > x \right) = 1
\end{align*}
uniformly for $x \in [\log^{1+\gamma}(T), (1-\gamma)T]$ for any $\gamma \in (0,1)$ as $T \to \infty$.
\end{proposition}

It is straightforward to obtain results such as (\ref{sketch_liminf}) and (\ref{sketch_lim}) in multi-armed settings.
To obtain lower bounds on the distribution tail of the number of plays $N_{r(i)}(T)$ of arm $r(i)$ (the $i$-th-best arm, for $i \ge 2$), we tilt the reward distributions of arms $r(1),\dots,r(i-1)$ so that their means become less than that of arm $r(i)$.
We choose the new environment $\widetilde{\nu}$ with the new arm parameter values, so that arm $r(i)$ becomes the optimal arm.
The change of measure from $\nu$ to $\widetilde{\nu}$ then results in the product of $i-1$ likelihood ratios corresponding to the arms $r(1),\dots,r(i-1)$.
Subsequently, each of the tilted parameter values for arms $r(1),\dots,r(i-1)$ can be optimized separately to yield, for example, (\ref{conclusion0}).
We refer the reader to the full proof of Theorem \ref{thm1} in Appendix \ref{thm1_proof}.

\subsection{Tail Probability Upper Bounds} \label{upperbound}

In Theorem \ref{thm1} from Section \ref{generalchar}, we developed a lower bound (\ref{conclusion0}) for the distribution tail of the number of plays $N_{r(i)}(T)$ of the $i$-th-best arm $r(i)$ (for $i \ge 2$) by an optimized algorithm.
In the presence of discrimination equivalence, we showed in (\ref{conclusion1}) that the tail exponent for $R(T)$, as determined by $N_{r(2)}(T)$, is exactly equal to $-1$.
The lower bound part of this result is obtained using (\ref{conclusion0}) and discrimination equivalence.
The upper bound part follows directly from Markov's inequality, as discussed in the Introduction.

However, when discrimination equivalence does not hold, the upper bound derived from Markov's inequality does not match the lower bounds.
As part of Theorem \ref{generalupperbound}, we develop refined upper bounds for the tail of $N_{r(i)}(T)$ for all $i \ge 2$, for the KL-UCB algorithm (Algorithm 2 and Theorem 1 of \cite{cappe_etal2013}).
These refined upper bounds exactly match the lower bounds in (\ref{conclusion0}), thereby providing strong evidence that the lower bounds in (\ref{conclusion0}) are tight more generally, regardless of whether or not discrimination equivalence holds.
The proof of Theorem \ref{generalupperbound} is given in Appendix \ref{generalupperbound_proof}.

\begin{theorem} \label{generalupperbound}
Let $\pi$ be $\mathcal{M}_P$-optimized KL-UCB.
Then for any environment $\nu = (P^{\mu_1},\dots,P^{\mu_K}) \in \mathcal{M}_P^K$ and the $i$-th-best arm $r(i)$,
\begin{align}
\lim_{T \to \infty} \frac{\log \P_{\nu \pi}(N_{r(i)}(T) > x)}{\log(x)} = - \sum_{j = 1}^{i-1} \inf_{z \in \mathcal{I}_P \, : \, z < \mu_{r(i)}} \frac{d_P(z,\mu_{r(j)})}{d_P(z,\mu_{r(i)})} \label{klucbtailexponent}
\end{align}
uniformly for $x \in [\log^{1+\gamma}(T), (1-\gamma)T]$ for any $\gamma \in (0,1)$ as $T \to \infty$.
\end{theorem}

From (\ref{klucbtailexponent}), we see that the tail exponents for the distributions of $N_{r(i)}(T)$, $i \ge 3$ are always strictly less than that of $N_{r(2)}(T)$.
So $N_{r(2)}(T)$ determines the exponent of the distribution tail of the regret $R(T)$; see also Remark \ref{remark1} below.
Indeed, when $P$ is discrimination equivalent, Lemma \ref{lem1} implies that the right side of (\ref{klucbtailexponent}) is exactly $-(i-1)$ for $i \ge 3$, which can be compared to (\ref{conclusion1}).
Whenever $P$ is not discrimination equivalent (for example, for all discrete distributions with support bounded to the left and strictly positive mass on the infimum of the support; see Proposition \ref{prop4}), the right side of (\ref{klucbtailexponent}) is always strictly less than $-1$ for the second-best arm $r(2)$.
So the regret tail is always strictly lighter than truncated Cauchy in such settings.
We confirm this fact for Bernoulli environments through numerical simulations in Figure \ref{figure3} in Section \ref{numerical}.

This indicates that an algorithm optimized for (and operating within) an environment class $\mathcal{M}_P^K$, when $P$ is a discrete distribution, is in general less fragile than when $P$ is a continuous distribution.
However, recall from Corollary \ref{cor0} that regardless of whether the reward distributions are discrete or continuous, there always exist environments in $\mathcal{M}_P^K$ for which the regret tail is arbitrarily close to being truncated Cauchy.
Optimized algorithms universally suffer from this weaker sense of fragility.
In fact, as we will see in Section \ref{supportingresults}, this is a key characteristic of optimized algorithms that, together with our change of measure argument, leads to a new proof of a generalized version of the Lai-Robbins lower bound. (See Theorem \ref{thm0} and Theorem \ref{thm0}.)

\begin{remark} \label{remark1}
In the setting of Theorem \ref{generalupperbound}, the distribution tail of the regret $R(T)$, as determined by that of $N_{r(2)}(T)$, depends only on the top two arm reward distributions.
(In this case, the KL divergences in (\ref{klucbtailexponent}) only involve $P^{\mu_{r(1)}}$ and $P^{\mu_{r(2)}}$.)
In contrast, all sub-optimal arms contribute to the expected regret.
\end{remark}

We also point out that (\ref{klucbtailexponent}) in Theorem \ref{generalupperbound} holds uniformly over a greater range $[\log^{1+\gamma}(T), (1-\gamma)T]$ than the range $[T^\gamma, (1-\gamma)T]$ of (\ref{conclusion1}) in Theorem \ref{thm1}.
As discussed in the Introduction, in reference to the CLT's for regret developed in \cite{fan_etal2022}, the large deviations of regret correspond to deviations from the expected regret that are of order $\log(T)$.
While we do not analyze deviations on such a scale in this paper, we do interpolate between the $\log(T)$ and poly-$T$ regions by considering the poly-$\log(T)$ region of the regret tail.
Since we simply relied on logarithmic expected regret and Markov's inequality in Theorem \ref{thm1} to establish the upper bound part of (\ref{conclusion1}), there we could not make conclusions about the poly-$\log(T)$ regions.
Here in Theorem \ref{generalupperbound}, however, we perform careful analysis to establish a more informative upper bound, which gives us insight about the poly-$\log(T)$ regions.

In Sections \ref{additional} and \ref{improvement}, we will frequently use the KL-UCB algorithm and general UCB algorithms as examples to illustrate fragility issues and modifications to alleviate fragility issues.
In Theorem \ref{generalupperbound} above, we characterized the regret tail of $\mathcal{M}_P$-optimized KL-UCB operating within environments from $\mathcal{M}_P^K$, i.e., the environment is well-specified.
Later in Proposition \ref{wrongdistribution}, we develop a result for general UCB algorithms operating in essentially arbitrary environments, including mis-specified ones.

\section{Illustrations of Fragility} \label{additional}

\subsection{Overview of Results} \label{overview_fragility}
Throughout Section \ref{additional}, we highlight several ways in which optimized algorithms are fragile.
Whereas previously we developed regret tail characterizations for optimized algorithms in \textit{well-specified} settings, we now consider \textit{mis-specified} settings.
By mis-specified, we mean that an algorithm $\pi$ is designed (possibly optimized) for some class of environments $\mathcal{M}^K$, but $\pi$ operates in an environment $\nu \notin \mathcal{M}^K$. 
In real world settings, there is generally some degree of model mis-specification.
So it is important to understand the sensitivity of algorithmic performance to mis-specification, of which there are many different forms.

Recall that in the formulation of instance-based asymptotic optimality of bandit algorithms, we first restrict consideration to the class of consistent algorithms, as in Definition \ref{A00}.
Then, within the class of consistent algorithms, we seek optimized algorithms that achieve the Lai-Robbins lower bound, as in Definition \ref{A1}.
From (\ref{conclusion1}) in Theorem \ref{thm1}, in every (well-specified) environment instance, optimized algorithms have extremely heavy regret tails, essentially truncated Cauchy with tail exponent exactly equal to $-1$ (under the discrimination equivalence condition).
If the tail exponent is at all $> -1$, then expected regret grows polynomially in the time horizon, i.e., consistency is lost.
Thus, in every single (well-specified) environment instance, optimized algorithms just barely maintain consistency.
Moreover, this suggests that the slightest degree of model mis-specification can cause optimized algorithms to not only suffer greater expected regret, but altogether lose the basic consistency property by suffering expected regret growing polynomially in the time horizon.

Our results in Sections \ref{mis-specification}-\ref{dependent} indicate that the hypothesis of the previous sentence is true for optimized algorithms in great generality.
It is useful to highlight two particular examples involving KL-UCB (an optimized algorithm in well-specified settings), which were touched upon in the Introduction and will be discussed in more detail in the sections to follow.
First, consider KL-UCB designed for iid Gaussian rewards with variance $\sigma^2$, and operating in an environment with iid Gaussian rewards with variance $\sigma_0^2 > \sigma^2$.
Second, consider the same KL-UCB algorithm designed for iid Gaussian rewards, but operating in an environment with rewards evolving according to a stationary AR(1) process with positive AR coefficient and Gaussian marginal distributions matching the algorithm's Gaussian design (with the same variance $\sigma^2$).
So the first case corresponds to mis-specified marginal distributions, whereas the second case corresponds to mis-specified correlation structure.
In both cases, no matter how close $\sigma_0^2$ is to $\sigma^2$, or how close the AR(1) coefficient is to zero, the KL-UCB algorithm will have a regret tail that is heavier than truncated Cauchy, with tail exponent $> -1$.
Thus, KL-UCB is inconsistent in these settings, with expected regret growing polynomially in the time horizon.
Strikingly, these results hold no matter how big the separation is between the arm means.

In Section \ref{mis-specification}, we study mis-specification of the marginal distributions of rewards in iid settings.
Then, in Section \ref{lowerbound}, we develop lower bounds on the regret tail for general reward processes, which are subsequently applied to study mis-specification of the serial dependence structure of rewards in Section \ref{dependent}.
Our theory in these sections will primarily be developed for $\mathcal{M}_P$-optimized KL-UCB for any chosen base distribution $P$, and operating in an environment $\nu \notin \mathcal{M}_P^K$.
To avoid pathological/trivial situations and ensure that the KL divergence function $d_P$ remains well-defined in mis-specified settings, we will assume that $\mathcal{I}_P$ is an interval (possibly infinite) that contains the range of all possible values of rewards for each arm of the true environment $\nu$.

To conclude our illustrations of fragility, we examine in Section \ref{highermoments} the higher moments (beyond the first moment) of regret for optimized algorithms operating in well-specified environments.
Under the assumptions of Theorem \ref{thm1}, we will see that optimizing for expected regret provides no control (uniform integrability) over any higher power of regret.
Higher moments grow as powers of the time horizon $T$ instead of as poly-$\log(T)$.

\subsection{Mis-specified Reward Distribution} \label{mis-specification}

In this section, we examine the regret tail behavior of optimized algorithms under mis-specification of marginal reward distributions.
We begin with Proposition \ref{wrongdistribution}, which is a characterization of the regret tail of (possibly) mis-specified KL-UCB operating in an environment $\nu = (Q_1,\dots,Q_K)$, where arm $i$ yields independent rewards from some distribution $Q_i$.
We can compare the right side of (\ref{wrongdistributionratio}) in Proposition \ref{wrongdistribution} to the right side of (\ref{klucbtailexponent}) in Theorem \ref{generalupperbound}.
In well-specified settings, Theorem \ref{generalupperbound} and Proposition \ref{wrongdistribution} are the same result.
In mis-specified settings, which is covered by Proposition \ref{wrongdistribution}, the KL divergences $d_{Q_{r(j)}}$ in the numerator do not match the KL divergence $d_P$ in the denominator.

The proof of Proposition \ref{wrongdistribution} is given in Appendix \ref{mis-specificationproofs}.
The proof uses a LLN (Lemma \ref{stronglaw}) for the regret of (possibly) mis-specified KL-UCB, and a general tail probability lower bound (Theorem \ref{gartnerellis}), which are deferred to Section \ref{lowerbound}. 
These supporting results are developed for more general (possibly non-iid) reward processes.
They are useful for establishing the results in Section \ref{dependent}, but they are stronger than needed in the current section.

\begin{proposition} \label{wrongdistribution}
   Let $\pi$ be $\mathcal{M}_P$-optimized KL-UCB. Let the environment $\nu = (Q_1,\dots,Q_K)$, where arm $i$ yields independent rewards from some distribution $Q_i$ such that its CGF $\Lambda_{Q_i}(\theta) < \infty$ for $\theta$ in a neighborhood of zero.
    Then for the $i$-th-best arm $r(i)$,
\begin{align}
\lim_{T \to \infty} \frac{\log \P_{\nu \pi}(N_{r(i)}(T) > x)}{\log(x)} = - \sum_{j=1}^{i-1} \inf_{z \in \mathcal{I}_{Q_{r(j)}} \, : \, z < \mu(Q_{r(i)})} \frac{d_{Q_{r(j)}}(z, \mu(Q_{r(j)}))}{d_P(z,\mu(Q_{r(i)}))} \label{wrongdistributionratio}
\end{align}
uniformly for $x \in [\log^{1+\gamma}(T),(1-\gamma)T]$ for any $\gamma \in (0,1)$ as $T \to \infty$.
\end{proposition}

In Corollary \ref{cor1} below, we show that for Gaussian KL-UCB operating in environments with iid Gaussian rewards, if the actual variance is just slightly greater than the variance specified in the algorithm design, then the expected regret will grow at a rate that is a power of $T$.
The proof details simplify significantly in this Gaussian setting, and for future reference, we provide a stand-alone proof of Corollary \ref{cor1} in Appendix \ref{mis-specificationproofs}.
See Figure \ref{figure1} in Section \ref{numerical} for numerical simulations illustrating (\ref{varianceratio}).

\begin{corollary} \label{cor1}
Let $\pi$ be KL-UCB optimized for iid Gaussian rewards with variance $\sigma^2 > 0$.
Then for any two-armed environment $\nu$ yielding iid Gaussian rewards with actual variance $\sigma_0^2 > 0$,
\begin{align}
\lim_{T \to \infty} \frac{\log \P_{\nu \pi}(N_{r(2)}(T) > x)}{\log(x)} = - \frac{\sigma^2}{\sigma_0^2} \label{varianceratio}
\end{align}
uniformly for $x \in [\log^{1+\gamma}(T),(1-\gamma)T]$ for any $\gamma \in (0,1)$ as $T \to \infty$.
So if $\sigma_0^2 > \sigma^2$, then for any $a \in (\sigma^2/\sigma_0^2,1]$,
\begin{align*}
\liminf_{T \to \infty} \frac{\E_{\nu \pi}[N_{r(2)}(T)]}{T^{1 - a}} \ge 1.
\end{align*}
\end{corollary}
Corollary \ref{cor1} also holds with $\pi$ as TS designed for iid Gaussian rewards with variance $\sigma^2 > 0$ (and with Gaussian priors on the arm means).
In particular, we can obtain this result by using the SLLN's developed in \cite{fan_etal2022}.

\subsection{Tail Probability Lower Bounds for General Reward Processes} \label{lowerbound}

In this section, we develop supporting results, which are needed in Section \ref{dependent} to establish regret tail characterizations in settings where the dependence structures of rewards are mis-specified. 
These supporting results can also be used to derive the results in Section \ref{mis-specification} in settings where the marginal reward distributions are mis-specified.
Lemma \ref{stronglaw} is a SLLN for the regret of KL-UCB operating in an environment with general (possibly non-iid) reward processes that satisfy Assumptions \ref{A3}-\ref{A4} below.
KL-UCB is an example of an algorithm that is so-called $\mathcal{M}_P$-\textit{pathwise convergent}, a notion that we introduce in Definition \ref{A10} below.
In Theorem \ref{gartnerellis}, we apply our change-of-measure argument to establish lower bounds for the regret tail of such algorithms when operating in an environment with reward processes satisfying Assumptions \ref{A3}-\ref{A4}.

We first state a few definitions and assumptions for the reward processes $X_i(t)$, $i \in [K]$, $t \ge 1$ that we will work with.
For each arm $i$ and sample size $n \ge 1$, define the re-scaled CGF of the sample mean of arm rewards:
\begin{align}
    \overbar{\Lambda}_i^n(\theta) = \frac{1}{n} \log \; \E\left[ \exp\left( \theta \cdot \sum_{t = 1}^n X_i(t) \right) \right]. \nonumber
\end{align}
We will assume the following for each arm $i$. 
\begin{assumption} \label{A3}
The limit $\overbar{\Lambda}_i(\theta) = \lim_{n \to \infty} \overbar{\Lambda}_i^n(\theta)$ exists (possibly infinite) for each $\theta \in \mathbb{R}$, and $0 \in \overbar{\Theta}_i := \textnormal{interior} \{\theta \in \mathbb{R} : \overbar{\Lambda}_i(\theta) < \infty \}$.
\end{assumption}
\begin{assumption} \label{A4}
$\overbar{\Lambda}_i(\cdot)$ is differentiable throughout $\overbar{\Theta}_i$, and either $\overbar{\Theta}_i = \mathbb{R}$ or $\lim_{m \to \infty} \abs{\overbar{\Lambda}_i'(\theta^m)} = \infty$ for any sequence $\theta^m \in \overbar{\Theta}_i$ converging to a boundary point of $\overbar{\Theta}_i$.
\end{assumption}
These are the conditions ensuring that the G\"{a}rtner-Ellis Theorem holds for the sample means of arm rewards (see, for example, Theorem 2.3.6 of \cite{dembo_etal1998}).
In the context of Assumption \ref{A3}, we refer to the limit $\overbar{\Lambda}_i$ as the \textit{limiting CGF} for arm $i$.
In the context of Assumption \ref{A4}, $\overbar{\Lambda}_i'(0)$, the derivative of limiting CGF evaluated at zero, is the long-run mean reward for arm $i$.
Indeed, by the G\"{a}rtner-Ellis Theorem and the Borel-Cantelli Lemma,
\begin{align}
    \frac{1}{n} \sum_{t=1}^n X_i(t) \to \overbar{\Lambda}_i'(0) \label{slln1}
\end{align}
almost surely as $n \to \infty$ for each arm $i$.
The optimal arm $r(1)$ is such that $\overbar{\Lambda}_{r(1)}'(0) = \max_{i \in [K]} \overbar{\Lambda}_i'(0)$.

In the current section and in Section \ref{dependent}, we also assume for simplicity that the reward process for each arm only evolves forward in time when the arm is played.
This ensures that the serial dependence structures of the reward processes are not interrupted in a complicated way by an algorithm's adaptive sampling schedule, and allows us to determine the limit in Assumption \ref{A3} for various processes of interest such as Markov chains. 
Regardless of the specifics of the serial dependence structure of rewards for each arm, we will always assume that there is no dependence between rewards of different arms.

Before stating Lemma \ref{stronglaw} and Theorem \ref{gartnerellis}, we introduce the following notion, which can be compared to the notion of an $\mathcal{M}_P$-optimized algorithm in Definition \ref{A1}.
\begin{definition}[Pathwise Convergent Algorithm] \label{A10}
An algorithm $\pi$ is $\mathcal{M}_P$-\textit{pathwise convergent} if for any environment $\nu$ yielding arm reward sequences $X_i(t)$, $i \in [K]$, $t \ge 1$,
\begin{align}
    \left\{ \omega : \lim_{n \to \infty} \frac{1}{n} \sum_{t=1}^n X_i(t) = c_i, \; i \in [K] \right\} \subset \left\{ \omega : \lim_{T \to \infty} \frac{N_{i}(T)}{\log(T)} = \frac{1}{d_P(c_i,\max_j c_j)}, \; \forall \; i \ne \argmax_j c_j \right\}. \label{samplepathconsistent}
\end{align}
\end{definition}
We conjecture that, in general, $\mathcal{M}_P$-optimized algorithms are also $\mathcal{M}_P$-pathwise convergent.
This is directly supported by Lemma \ref{stronglaw} below, as well as by the SLLN developed for Gaussian TS in \cite{fan_etal2022}. 
It is also suggested by the analysis for developing SLLN's for non-optimized forced sampling-based algorithms and other UCB algorithms in \cite{cowan_etal2019}.
The proof of Lemma \ref{stronglaw} is based on the arguments in \cite{cowan_etal2019}, and can be found in Appendix \ref{lowerboundproofs}.

\begin{lemma} \label{stronglaw}
    $\mathcal{M}_P$-optimized KL-UCB is $\mathcal{M}_P$-pathwise convergent.
\end{lemma}

We now introduce Theorem \ref{gartnerellis}, whose proof can be found in Appendix \ref{gartnerellis_proof}.
For arm $i$, we use $\overbar{\Lambda}_i^*$ to denote the convex conjugate of the limiting CGF $\overbar{\Lambda}_i$, and we define $\overline{\mathcal{I}}_i = \textnormal{interior}\{z \in \mathbb{R} : \overbar{\Lambda}_i^*(z) < \infty \}$.
As mentioned previously, to avoid pathological/trivial situations, we will always assume for each arm $i$ that $\overline{\mathcal{I}}_i \subset \mathcal{I}_P$ for the chosen base distribution $P$.
(We also recall that the convex conjugate of the limiting CGF is the \textit{rate function} in the G\"{a}rtner-Ellis Theorem.)
\begin{theorem} \label{gartnerellis}
    Let $\pi$ be $\mathcal{M}_P$-pathwise convergent.
    Let the $K$-armed environment $\nu$ yield rewards for each arm that evolve according to any process satisfying Assumptions \ref{A3}-\ref{A4}.
    Then for the $i$-th-best arm $r(i)$,
\begin{align}
\liminf_{T \to \infty} \inf_{x \in B_\gamma(T)} \frac{\log \P_{\nu \pi}(N_{r(i)}(T) > x)}{\log(x)} \ge - \sum_{j=1}^{i-1} \inf_{ z \in \overline{\mathcal{I}}_{r(j)} \, : \, z < \overbar{\Lambda}_{r(i)}'(0)} \frac{\overbar{\Lambda}^*_{r(j)}(z)}{d_P(z,\overbar{\Lambda}_{r(i)}'(0))}, \label{gartnerellisratio}
\end{align}
with $B_\gamma(T) = [\log^{1+\gamma}(T),(1-\gamma)T]$ and any $\gamma \in (0,1)$.
\end{theorem}

\begin{remark} \label{remark100}
Whenever we can establish a WLLN for the $N_i(T)$ (e.g., as in Lemma \ref{lemma2}), then our change-of-measure approach can be used to obtain lower bounds on the tail probabilities of the $N_i(T)$ (as in Theorems \ref{thm1} and \ref{gartnerellis}).
The almost sure convergence of the $N_i(T)$, as provided by Assumptions \ref{A3}-\ref{A4} (leading to (\ref{slln1})) together with pathwise convergence (in Definition \ref{A10}), is sufficient but not necessary.
\end{remark}

\subsection{Mis-specified Reward Dependence Structure} \label{dependent}

Even if the marginal distributions of the arm rewards are correctly specified, optimized algorithms such as KL-UCB (designed for iid rewards) can still be susceptible to mis-specification of the serial dependence structure.
In Corollary \ref{cor2}, we provide a lower bound characterization of the regret tail for Gaussian KL-UCB applied to bandits with rewards evolving as Gaussian AR(1) processes.
Specifically, for each arm $i$, we assume the rewards evolve as an AR(1) process: 
\begin{align}
    X_i(t) = \alpha_i + \beta_i X_i(t-1) + W_i(t), \label{arprocess}
\end{align}
where the $\beta_i \in (0,1)$ and the $W_i(t)$ are iid $N(0,\sigma_i^2)$.
The equilibrium distribution for arm $i$ is then $N(\alpha_i/(1-\beta_i),\sigma_i^2/(1-\beta_i^2))$.
For simplicity, we assume that the AR(1) reward process for each arm is initialized in equilibrium.
So the marginal mean (also the long-run mean as in (\ref{slln1})) for arm $i$ is $\overbar{\Lambda}_i'(0) = \alpha_i/(1-\beta_i)$.
The proof of Corollary \ref{cor2} follows from a straightforward verification of Assumptions \ref{A3}-\ref{A4}, which is omitted, and then a direct application of Theorem \ref{gartnerellis}.

\begin{corollary} \label{cor2}
Let $\pi$ be KL-UCB optimized for iid Gaussian rewards with variance $\sigma^2 > 0$.
Then for any two-armed environment $\nu$ yielding rewards that evolve as AR(1) processes (as in (\ref{arprocess})),
\begin{align*}
\liminf_{T \to \infty} \inf_{x \in B_\gamma(T)} \frac{\log \P_{\nu \pi}(N_{r(2)}(T) > x)}{\log(x)} \ge - \frac{\sigma^2}{\sigma_{r(1)}^2}(1-\beta_{r(1)})^2, 
\end{align*}
with $B_\gamma(T) = [\log^{1+\gamma}(T),(1-\gamma)T]$ and any $\gamma \in (0,1)$.
\end{corollary}

To see the effect of mis-specifying the dependence structure, suppose $\sigma_1^2 = \sigma_2^2 = \sigma_0^2$ and $\beta_1 = \beta_2 = \beta_0$, for some $\sigma_0^2 > 0$ and $\beta_0 \in (0,1)$, so that the equilibrium distributions for the rewards of both arms are Gaussian with variance $\sigma_0^2/(1-\beta_0^2)$.
Then, even if we specify the same variance $\sigma^2 = \sigma_0^2/(1-\beta_0^2)$ in Gaussian KL-UCB, so that the marginal distribution of rewards is correctly specified, we still end up with a tail exponent that is strictly greater than $-1$.
This is due to the mis-specification of the serial dependence structure.
Specifically, using Corollary \ref{cor2},
\begin{align}
\liminf_{T \to \infty} \frac{\log \P_{\nu \pi}(N_{r(2)}(T) > T/2)}{\log(T)} \ge - \frac{1-\beta_0}{1+\beta_0}, \label{arvarianceratio2}
\end{align}
and so for any $a \in ((1-\beta_0)/(1+\beta_0),1]$,
\begin{align*}
\liminf_{T \to \infty} \frac{\E_{\nu \pi}[N_{r(2)}(T)]}{T^{1 - a}} \ge 1. 
\end{align*}
We verify (\ref{arvarianceratio2}) through numerical simulations in Figure \ref{figure2} in Section \ref{numerical}.
The simulations suggest that the lower bound in (\ref{arvarianceratio2}) is tight.

In Proposition \ref{markovian} below, we develop a characterization of the regret tail of KL-UCB operating in an environment $\nu$ with rewards evolving as finite state Markov chains.
For each arm $i$, we assume that the rewards evolve as an irreducible Markov chain on a common, finite state space $S \subset \mathbb{R}$, with transition matrix $H_i$.
For any $\theta \in \mathbb{R}$ and transition matrix $H$, we use $\phi_{H}(\theta)$ to denote the logarithm of the Perron-Frobenius eigenvalue of the corresponding tilted transition matrix:
\begin{align}
    \bigl(\exp(\theta \cdot y) H(x,y), \; x,y \in S\bigr). \label{Qtilted1}
\end{align}
So in the context of Assumptions \ref{A3}-\ref{A4}, $\overbar{\Lambda}_i(\theta) = \phi_{H_i}(\theta)$ for each arm $i$. 
(Note that the convex conjugate $\overbar{\Lambda}_i^*$ of $\overbar{\Lambda}_i$ plays the same role in Proposition \ref{markovian} as it does in Theorem \ref{gartnerellis}.)
For simplicity, we assume that the Markov chain reward process for each arm is initialized in equilibrium.
So the marginal mean (also the long-run mean as in (\ref{slln1})) for arm $i$ is $\overbar{\Lambda}_i'(0) = \phi_{H_i}'(0)$.
Lastly, we wish to ensure that any equilibrium mean between $s_{\text{min}} := \min S$ and $s_{\text{max}} := \max S$ can be realized through tilting the transition matrices as in (\ref{Qtilted1}).
This provides technical convenience, and allows us to use Chernoff-type bounds for Markov chains from the existing literature to derive upper bounds on the regret tail.
So we introduce the following notion.
We say that a transition matrix $H$ on $S$ satisfies the \textit{Doeblin Condition} if we have $H(x,s_{\text{min}}) > 0$ for each $x \ne s_{\text{min}}$, and $H(x,s_{\text{max}}) > 0$ for each $x \ne s_{\text{max}}$.

The lower bound part of Proposition \ref{markovian} follows from a straightforward verification of Assumptions \ref{A3}-\ref{A4}, which is omitted, and then a direct application of Theorem \ref{gartnerellis}. To establish the upper bound part, we can again use the proof of Theorem \ref{generalupperbound} (in Appendix \ref{generalupperbound_proof}) and substitute in, where appropriate (in (\ref{term3}) and (\ref{term7bound})), a Chernoff-type bound for additive functionals of finite-state Markov chains.
One version of such a result that is convenient for our purposes is established in Theorem 1 of \cite{moulos_etal2019}. 
(Earlier and more general results can be found in \cite{miller_1961} and \cite{kontoyiannis_etal2003}, respectively.)

\begin{proposition} \label{markovian}
    Let $\pi$ be $\mathcal{M}_P$-optimized KL-UCB.
    Let the $K$-armed environment $\nu$ yield rewards for each arm that evolve according to an irreducible Markov chain with a finite state space (with $\overbar{\Lambda}_i$ as defined above for each arm $i$), and suppose that the transition matrix for each arm satisfies the Doeblin Condition.
    Then for the $i$-th-best arm $r(i)$,
\begin{align}
\lim_{T \to \infty} \frac{\log \P_{\nu \pi}(N_{r(i)}(T) > x)}{\log(x)} = - \sum_{j=1}^{i-1} \inf_{ z \in \overline{\mathcal{I}}_{r(j)} \, : \, z < \overbar{\Lambda}_{r(i)}'(0)} \frac{\overbar{\Lambda}_{r(j)}^*(z)}{d_P(z,\overbar{\Lambda}_{r(i)}'(0))} \label{markovianratio}
\end{align}
uniformly for $x \in [\log^{1+\gamma}(T),(1-\gamma)T]$ for any $\gamma \in (0,1)$ as $T \to \infty$.
\end{proposition}

\begin{example} \label{ex4}
For the state space $S = \{0,1\}$ (binary rewards), we can examine some numerical values for the right side of (\ref{markovianratio}).
Here, we take $d_P(z,z')$ to be the KL divergence between Bernoulli distributions with means $z$ and $z'$.
We assume the arm rewards evolve as Markov chains on $S$.
So the marginal distributions of the arm rewards are well-specified.
Suppose the best arm $r(1)$ evolves according to a transition matrix of the form:
\begin{align}
    H_{r(1)} = \begin{bmatrix}
    1-q \;\;\;\;\; & q \\
    1-w(q) \;\;\;\;\; & w(q)
    \end{bmatrix}. \label{q1}
\end{align}
For any $q \le 0.8$, we set $w(q) \ge 0.8$ such that the chain evolving on $S$ according to $H_{r(1)}$ has equilibrium mean equal to $0.8$.
Suppose also that the gap between the equilibrium means of the top two arms, $r(1)$ and $r(2)$, is $\Delta > 0$.
In Table \ref{table1} below, we provide numerical values for the right side of (\ref{markovianratio}) for the case $i=2$ and for different values of $q$ and $\Delta$.
As $q$ becomes smaller relative to $0.8$, the autocorrelation in the rewards for arm $r(1)$ becomes more positive, and the resulting regret distribution tail becomes heavier.
As the gap $\Delta$ shrinks, the resulting regret tail also becomes heavier.
We can see from Table \ref{table1} that it is fairly easy (for reasonable values of $q$ and $\Delta$) to obtain regret tails that are heavier than truncated Cauchy (the right side of (\ref{markovianratio}) is greater than $-1$).
\begin{table}[H]
\centering
\begin{tabular}{ccccccccccc}
\toprule
\multicolumn{2}{c}{} & \multicolumn{9}{c}{$\Delta$} \\
\cmidrule(rl){3-11}
$q$ & $w(q) \quad$ & 0.12 & 0.11 & 0.10 & 0.09 & 0.08 & 0.07 & 0.06 & 0.05 & 0.04 \\
\midrule
$0.8$ & $0.8 \quad$ & -1.41 & -1.37 & -1.34 & -1.30 & -1.26 & -1.23 & -1.19 & -1.16 & -1.13 \\
$0.7$ & $0.825 \quad$ & -1.06 & -1.03 & -1.00 & -0.97 & -0.95 & -0.92 & -0.89 & -0.87 & -0.84 \\
$0.6$ & $0.85 \quad$ & -0.80 & -0.78 & -0.76 & -0.74 & -0.72 & -0.70 & -0.68 & -0.66 & -0.64 \\
$0.5$ & $0.875 \quad$ & -0.61 & -0.59 & -0.58 & -0.56 & -0.54 & -0.53 & -0.51 & -0.50 & -0.49 \\
\bottomrule
\end{tabular}
\caption{\label{table1} For arm rewards evolving as Markov chains on the state space $S = \{0,1\}$, and with $d_P$ being the Bernoulli KL divergence, we provide numerical values for the right side of (\ref{markovianratio}).
Here, $i=2$, and we consider different values of $q$ (as used in the best arm's transition matrix $H_{r(1)}$ in (\ref{q1})) and $\Delta$ (the gap between the equilibrium means of the two best arms, $r(1)$ and $r(2)$).}
\end{table}
\end{example}

\subsection{Higher Moments} \label{highermoments}

In this section, we point out that the $1+\delta$ moment of regret for any $\delta > 0$ must grow roughly as $T^\delta$.
Contrary to what one might conjecture in light of the WLLN that we saw in Lemma \ref{lemma2}, the $1+\delta$ moment of regret is not poly-logarithmic. 
In Corollary \ref{cor3} below, which is a direct consequence of Theorem \ref{thm1}, we show that expected regret minimization does not provide any help in controlling higher moments of regret.
It forces the tail of the regret distribution to be as heavy as possible while ensuring the expected regret scales as $\log(T)$ (as we saw in Theorem \ref{thm1} and Corollary \ref{cor0}).
Consequently, there is no control over the distribution tails of $1+\delta$ powers of regret, and thus no uniform integrability of $1+\delta$ powers of regret (normalized by $\log^{1+\delta}(T)$).

\begin{corollary} \label{cor3}
Let $\pi$ be $\mathcal{M}_P$-optimized.
Suppose also that $P$ is discrimination equivalent.
Then for any environment $\nu \in \mathcal{M}_P^K$, and any $\delta > 0$ and $\delta' \in (0,\delta)$,
\begin{align*}
\liminf_{T \to \infty} \frac{\E_{\nu \pi}[N_{r(2)}(T)^{1+\delta}]}{T^{\delta'}} \ge 1.
\end{align*}
\end{corollary}

\section{A Generalization and a Trade-off} \label{newlyadded}

In this section, we consider well-specified settings and develop further results regarding the regret tail.
We consider a general model denoted by $\mathcal{M}$, which is allowed to be an arbitrary collection of distributions with finite means. 
In Section \ref{generalmodels}, Proposition \ref{prop_klinf} is a lower bound for the regret tail for such general models $\mathcal{M}$, which generalizes the lower bound in (\ref{conclusion0}) from Theorem \ref{thm1} (developed for exponential family models).
In Section \ref{supportingresults}, in the context of a general model $\mathcal{M}$, Theorem \ref{thm0} exposes a trade-off between the heaviness of the regret tail and the growth rate of expected regret, with lighter regret tails implying larger growth rates of expected regret.

\subsection{Tail Probability Lower Bounds for General Models} \label{generalmodels}

Before establishing Proposition \ref{prop_klinf}, which is a lower bound for the regret tail of optimized algorithms in a general model $\mathcal{M}$ (to be defined below), we first discuss some concepts.
We will use the following quantity frequently:
\begin{align}
    D_{\text{inf}}(Q,y,\mathcal{M}) := \inf \{D(Q \: \lVert \: Q') : Q' \in \mathcal{M}, \; \mu(Q') > y \}, \label{generalizedkl}
\end{align}
which is defined for a general model $\mathcal{M}$, an arbitrary distribution $Q$, and $y \in \mathbb{R}$.
Here, $D(Q \: \lVert \: Q')$ denotes the KL divergence between the distributions $Q$ and $Q'$, and we take the infimum of the empty set to be $+\infty$.
When $\mathcal{M} = \mathcal{M}_P$, an exponential family with base distribution $P$, then $D_{\text{inf}}(P^{\mu},\mu',\mathcal{M}_P) = d_P(\mu,\mu')$ for $\mu < \mu'$.

Next, we have the definition of optimized algorithm for a general model $\mathcal{M}$ (similar to Defintion \ref{A1} for exponential family models).
This notion of optimized algorithm comes from the generalized version (due to \cite{burnetas_etal1996}) of the Lai-Robbins lower bound for expected regret (from (\ref{lairobbins})).
\begin{definition}[$\mathcal{M}$-Optimized Algorithm] \label{A99} 
For a general model $\mathcal{M}$, an algorithm $\pi$ is $\mathcal{M}$-optimized if for any environment $\nu = (P_1,\dots,P_K) \in \mathcal{M}^K$ and each sub-optimal arm $i$,
\begin{align*}
\lim_{T \to \infty} \frac{\E_{\nu \pi}[N_i(T)]}{\log(T)} = \frac{1}{D_{\text{inf}}(P_i,\mu_*(\nu),\mathcal{M})}.
\end{align*}
\end{definition}

The above definition of optimized algorithm leads to the regret tail lower bound in Proposition \ref{prop_klinf} below.
This result generalizes the regret tail lower bound in (\ref{conclusion0}) from Theorem \ref{thm1}.
Its proof, which we omit, follows from a direct adaptation of the proof of (\ref{conclusion0}) from Theorem \ref{thm1} (see Appendix \ref{thm1_proof}).
As in the proof of (\ref{conclusion0}) from Theorem \ref{thm1}, we require a WLLN for the number of sub-optimal arm plays of $\mathcal{M}$-optimized algorithms.
This is provided in Lemma \ref{lemma_klinf_wlln} below, which is an extension of the earlier Lemma \ref{lemma2} to general models $\mathcal{M}$.
The WLLN in Lemma \ref{lemma_klinf_wlln} follows immediately from (\ref{generalinproblowerbound}) in Theorem \ref{thm0} with the choice $f(T) = \log(T)$ (as discussed in Remark \ref{remark99} in the next section), together with the definition of an $\mathcal{M}$-optimized algorithm.

\begin{lemma} \label{lemma_klinf_wlln}
Let $\pi$ be $\mathcal{M}$-optimized.
Then for any environment $\nu = (P_1,\dots,P_K) \in \mathcal{M}^K$ and each sub-optimal arm $i$,
\begin{align*}
    \frac{N_i(T)}{\log(T)} \to \frac{1}{D_{\textnormal{inf}}(P_i,\mu_*(\nu),\mathcal{M})}
\end{align*}
in $\P_{\nu \pi}$-probability as $T \to \infty$.
\end{lemma}

\begin{proposition} \label{prop_klinf}
Let $\pi$ be $\mathcal{M}$-optimized.
Then for any environment $\nu = (P_1,\dots,P_K) \in \mathcal{M}^K$ and the $i$-th best arm $r(i)$,
\begin{align}
\liminf_{T \to \infty} \inf_{x \in B_\gamma(T)} \frac{\P_{\nu \pi}(N_{r(i)}(T) > x)}{\log(x)} \ge - \sum_{j=1}^{i-1} \inf_{Q \in \mathcal{M} \, : \, \mu(Q) < \mu(P_{r(i)})} \frac{D(Q \: \lVert \: P_{r(j)})}{D_{\textnormal{inf}}(Q,\mu(P_{r(i)}),\mathcal{M})}, \label{lowerbound_klinf}
\end{align}
with $B_\gamma(T) = [\log^{1+\gamma}(T),(1-\gamma)T]$ and any $\gamma \in (0,1)$.
\end{proposition}

When Proposition \ref{prop_klinf} is specialized to a model $\mathcal{M}_P$ as in (\ref{model})-(\ref{modelp}), we recover (\ref{conclusion0}) in Theorem \ref{thm1}.
In this case, the infima of the individual optimization problems on the right side of (\ref{lowerbound_klinf}) are attained or approached by taking $Q = P^z$ with $z \downarrow \inf \mathcal{I}_P$.
In other words, the optimal change of measure is to exponentially tilt the mean of the optimal arm to be as small as possible; recall the discussion following (\ref{klequivalence}) in Section \ref{generalchar}.
In practical terms, as discussed in Section \ref{sketch}, this suggests that large regret arises due to the mean of the optimal arm being severely under-estimated to be below the means of the sub-optimal arm(s).
However, for general models $\mathcal{M}$, the distribution(s) $Q \in \mathcal{M}$ that attain or approach the infima for each of the individual optimization problems on the right side of (\ref{lowerbound_klinf}) may not be straightforward to determine.
In general, the optimal change of measure will not be a simple exponential tilt to change the mean.

We can already see interesting distinctions in the following example, where $\mathcal{M}$ is the set of all Gaussian distributions with unknown means and also unknown variances.
Note that this is a strictly larger model than any model $\mathcal{M}_P$ with Gaussian base distribution $P$ having a particular variance that is known (so that $\mathcal{M}_P$ is parameterized only by the unknown means).

\begin{example} \label{ex9}
Let $\mathcal{M}$ be the collection of all Gaussian distributions with all possible means and variances.
This model corresponds to the setting with Gaussian rewards, where both the mean and variance are unknown.
For any distribution $Q = N(z,\sigma^2)$ and $z < y$,
\begin{align*}
    D_{\textnormal{inf}}(Q,y,\mathcal{M}) = \frac{1}{2} \log\left( 1 + \frac{(y-z)^2}{\sigma^2} \right).
\end{align*}
Let $\nu = (P_1,\dots,P_K) \in \mathcal{M}^K$, with $P_k = N(\mu_k,\sigma_k^2)$ and $\sigma_k^2 > 0$.
Then, for $i = 2$ and $j = 1$, the optimization problem on the right side of (\ref{lowerbound_klinf}) can be expressed as:
\begin{align*}
    - \inf_{Q \in \mathcal{M} \, : \, \mu(Q) < \mu(P_{r(2)})} \frac{D(Q \: \lVert \: P_{r(1)})}{D_{\textnormal{inf}}(Q,\mu(P_{r(2)}),\mathcal{M})} = - \inf_{ (z,\sigma^2) \, : \, z < \mu_{r(2)}, \, \sigma^2 > 0} \frac{\log\left( \frac{\sigma_{r(1)}^2}{\sigma^2} \right) + \frac{\sigma^2 + (\mu_{r(1)}-z)^2}{\sigma_{r(1)}^2} - 1}{\log\left( 1 + \frac{(\mu_{r(2)}-z)^2}{\sigma^2} \right)} = -1.
\end{align*}
So, the regret tail in this setting is also truncated Cauchy.
Here, the optimal value (tail exponent) of $-1$ is achieved by fixing any $z < \mu_{r(2)}$ and then sending $\sigma^2 \downarrow 0$.
In other words, an optimal change of measure is to tilt the mean of the best arm to be any finite amount lower than the mean of the second-best arm, and then send the variance of the best arm to zero.
By contrast, for a Gaussian model $\mathcal{M}_P$ parameterized by mean with known variance (which is strictly smaller than $\mathcal{M}$), recall that the optimal change of measure is to tilt the mean of the best arm all the way to $-\infty$.
This type of change of measure (a simple exponential tilting) is inadequate for achieving the $-1$ tail exponent when the model is $\mathcal{M}$. 
\end{example}

\subsection{Trade-off: Regret Tail vs. Regret Expectation} \label{supportingresults}

In this section, we establish a trade-off between the heaviness (decay rate) of the regret tail and the growth rate of expected regret.
In particular, an upper bound on the decay rate of the regret tail implies a lower bound on the growth rate of expected regret.
And a greater decay rate of the regret tail in (\ref{tailconsistency}) must be accompanied by a greater growth rate of expected regret.
This holds for a range of different rates as captured through the choice of function $f$, from logarithmic to polynomial (in $T$), with the latter corresponding to exponential regret tails.

The trade-off we establish generalizes the asymptotic lower bounds for expected regret, first developed by \cite{lai_etal1985} for exponential families $\mathcal{M}_P$ and then extended by \cite{burnetas_etal1996} to general models $\mathcal{M}$.
To see this, take $f(T) = \log(T)$ in Theorem \ref{thm0}.
Then, the assumption of consistency of algorithms used in these papers implies the regret tail upper bound in (\ref{tailconsistency}), which is the starting assumption of Theorem \ref{thm0}.
More specifically, \cite{lai_etal1985} assumes $\mathcal{M}_P$-consistency, as in Definition \ref{A00} for exponential family models, and \cite{burnetas_etal1996} assumes $\mathcal{M}$-consistency, which is defined in the same way except for general models $\mathcal{M}$.
These consistency assumptions ensure, by Markov's inequality, that the regret tail cannot be heavier than truncated Cauchy in all such environments, i.e., the tail exponent must be $\le -1$, which is precisely the content of (\ref{tailconsistency}) for the choice $f(T) = \log(T)$.

We will study the tightness of the trade-off in Theorem \ref{thm0} by analyzing an extension of the KL-UCB algorithm (see Algorithm \ref{alg1}) in Section \ref{simple} below.
Specifically, see Proposition \ref{prop5} and Remark \ref{trade-off-tightness}.
As will be discussed, a sufficient condition for the trade-off to be tight is when $f$ is a so-called slowly varying function (intuitively, with growth slower than any polynomial).
When $f$ exhibits faster growth (for example, polynomial), then the lower bound for expected regret in (\ref{generallowerbound}) still has the correct asymptotic dependence on $T$, but the resulting constant on the right side may not be tight.

The proof of Theorem \ref{thm0} can be found in Appendix \ref{tradeoff_proof}.
The environments in the general environment class $\mathcal{M}^K$ are allowed to be arbitrary (provided the distributions have finite mean).
The best arm(s), second-best arm(s), etc., do not need to be unique.

\begin{theorem} \label{thm0}
Let $f : (1,\infty) \to (0,\infty)$ be an increasing function satisfying $\liminf_{t \to \infty} f(t)/\log(t) \ge 1$ and $f(t) = o(t)$.
For every environment $\nu = (P_1,\dots,P_K) \in \mathcal{M}^K$, suppose $\pi$ satisfies for each sub-optimal arm $i$: 
\begin{align}
    \limsup_{T \to \infty} \frac{\log \P_{\nu \pi}(N_i(T) > T/K)}{f(T)} \le -1. \label{tailconsistency}
\end{align}
Then for any such environment $\nu = (P_1,\dots,P_K) \in \mathcal{M}^K$, each sub-optimal arm $i$, and any $\epsilon \in (0,1)$,
\begin{align}
\lim_{T \to \infty} \P_{\nu \pi}\left( \frac{N_i(T)}{f(T)} \ge \frac{1-\epsilon}{D_{\textnormal{inf}}(P_i,\mu_*(\nu),\mathcal{M})} \right) = 1, \label{generalinproblowerbound}
\end{align}
and thus,
\begin{align}
\liminf_{T \to \infty} \frac{\E_{\nu \pi}[N_i(T)]}{f(T)} \ge \frac{1}{D_{\textnormal{inf}}(P_i,\mu_*(\nu),\mathcal{M})}. \label{generallowerbound}
\end{align}
\end{theorem}

\begin{remark} \label{remark99}
To see that Theorem \ref{thm0}, with the choice $f(T) = \log(T)$, implies Lemma \ref{lemma_klinf_wlln}, note the following.
For $\mathcal{M}$-optimized algorithms satisfying Definition \ref{A99}, (\ref{tailconsistency}) is automatically satisfied, since the regret tail of such algorithms (which are, of course, $\mathcal{M}$-consistent) cannot be heavier than truncated Cauchy, as discussed above.
Thus, the ``one-sided'' WLLN in (\ref{generalinproblowerbound}) holds for $\mathcal{M}$-optimized algorithms, which implies the full ``two-sided'' WLLN in Lemma \ref{lemma_klinf_wlln}.
In the special case that $\mathcal{M} = \mathcal{M}_P$ is an exponential family, these arguments apply verbatim to justify Lemma \ref{lemma2} for $\mathcal{M}_P$-optimized algorithms satisfying Definition \ref{A1}.
\end{remark}

\section{Improvement of the Regret Tail} \label{improvement}

In Sections \ref{main} and \ref{additional}, we have seen that the regret tails of algorithms optimized for restrictive model classes like exponential families are extremely heavy, to the point that such optimized algorithms are barely able to maintain consistency in well-specified settings.
In particular, the slightest degree of model mis-specification can result in the loss of consistency, with expected regret growing polynomially in the time horizon.
In Section \ref{simple}, we discuss a simple approach to make the regret tail lighter for UCB-type algorithms.
Our analysis establishes an explicit trade-off between the amount of UCB exploration (and corresponding expected regret) and the resulting heaviness of the regret tail (see Proposition \ref{prop5}, which also validates and strengthens the trade-off in Theorem \ref{thm0}).
Moreover, we show in Section \ref{robust1} that the modification to obtain lighter regret tails provides protection against mis-specification of the arm reward distributions in iid settings.
We show in Section \ref{robust2} that the modification also provides protection against Markovian departures from independence of the arm rewards.

\subsection{A Simple Approach to Obtain Lighter Regret Tails} \label{simple}

In this section, and in Sections \ref{robust1} and \ref{robust2}, we focus on extensions of the KL-UCB algorithm of \cite{cappe_etal2013}; see Algorithm \ref{alg1} below.
Like KL-UCB, Algorithm \ref{alg1} is defined for any exponential family $\mathcal{M}_P$, as in (\ref{model})-(\ref{modelp}).
However, in contrast to KL-UCB, we will allow for more flexibility in the rate of exploration, as specified by the ``exploration'' function $f$.
Whereas KL-UCB is developed for certain $f$ satisfying $\lim_{t \to \infty} f(t)/\log(t) = 1$, we will allow $f$ to have faster growth rates, including poly-logarithmic and polynomial rates. 

\begin{algorithm}
\caption{(from Algorithm 2 of \cite{cappe_etal2013})} \label{alg1}
\hspace{2.5mm} \textbf{input:} divergence $d_P : \mathcal{I}_P \times \mathcal{I}_P \to [0,\infty)$, increasing ``exploration'' function $f : (1,\infty) \to (0,\infty)$ 

\hspace{2.5mm} \textbf{initialize:} Play each arm $1,\dots,K$ once
\begin{algorithmic}
\For{$t \ge K$}
    \State Play the arm (with ties broken arbitrarily):
    \begin{align*}
        A(t+1) = \argmax_{i \in [K]} \; \sup\left\{ z \in \mathcal{I}_P : d_P(\widehat{\mu}_i(t),z) \le \frac{f(t)}{N_i(t)} \right\}
    \end{align*}
\EndFor
\end{algorithmic}
\end{algorithm}

Below, we have Proposition \ref{prop5}, the main result of the section.
In (\ref{psitail1}), we characterize the regret tail for Algorithm \ref{alg1} under different choices of the exploration function $f$.
We require some regularity conditions on $f$ in order to be able to obtain the exact asymptotic limit in (\ref{psitail1}).
We will use the notions of regularly varying and slowly varying functions (see, for example, \cite{embrechts_etal1997}).
A function $h : \mathbb{R} \to \mathbb{R}$ is regularly varying if $h(x) = x^a l(x)$ for some $a \in \mathbb{R}$ (which is allowed to be zero), with $l$ being a slowly varying function.
And a function $l : \mathbb{R} \to \mathbb{R}$ is slowly varying if $\lim_{x \to \infty} l(cx)/l(x) = 1$ for any $c > 0$.
Slowly varying functions can be intuitively thought of as functions with growth (or decay) that is slower than any polynomial (or inverse polynomial) (for example, any function with logarithmic or poly-logarithmic growth/decay).

The regret tail characterization in (\ref{psitail1}) is accompanied by an asymptotic characterization of expected regret in (\ref{asymptotic_expected_regret}).
To upper bound expected regret for Algorithm \ref{alg1}, a non-asymptotic minimum growth rate is required for $f$.
Here, we use the rate: $f(t) \ge \log(1+t\log^2(t))$ (for sufficiently large $t$).
This is motivated by the development of KL-UCB in Chapter 10 of \cite{lattimore_etal2020} (see their Algorithm 8), where the exploration function is chosen to be $t \mapsto \log(1+t\log^2(t))$.
Other choices are possible, including $t \mapsto \log(t) + 3\log\log(t)$, as used by \cite{cappe_etal2013} in the original development of KL-UCB.
Such non-asymptotic minimum growth rates for $f$ imply the asymptotic rate: $\liminf_{t \to \infty} f(t)/\log(t) \ge 1$, which is used to obtain the regret tail characterization in (\ref{psitail1}).

To establish (\ref{psitail1}) in Proposition \ref{prop5}, we adapt the proofs of Theorems \ref{thm1} and \ref{generalupperbound}. 
To establish (\ref{asymptotic_expected_regret}), the asymptotic upper bound for expected regret follows from standard techniques found in, for example, Chapter 10 of \cite{lattimore_etal2020}. 
The matching asymptotic lower bound for expected regret can be deduced using a LLN for the regret of Algorithm \ref{alg1} (which follows directly from the proof of Lemma \ref{stronglaw} in Appendix \ref{lowerboundproofs}), together with Markov's inequality. 
The complete proof details for Proposition \ref{prop5} can be found in Appendix \ref{simpleproofs}.

\begin{proposition} \label{prop5}
Let $\pi$ be Algorithm \ref{alg1}, with divergence $d_P$ and exploration function $f$. 

(i) 
Let $f$ satisfy $\liminf_{t \to \infty} f(t)/\log(t) \ge 1$ and $f(t) = o(t^\lambda)$ for some $\lambda \in (0,1)$,
and also let $f$ be regularly varying and strictly increasing.
Then, for any environment $\nu = (P^{\mu_1},\dots,P^{\mu_K}) \in \mathcal{M}_P^K$ and the $i$-th-best arm $r(i)$,
\begin{align}
\lim_{T \to \infty} \frac{\log \P_{\nu \pi}(N_{r(i)}(T) > x)}{f(x)} 
& = - \sum_{j=1}^{i-1} \inf_{z \in \mathcal{I}_P \, : \, z < \mu_{r(i)}} \frac{d_P(z,\mu_{r(j)})}{d_P(z,\mu_{r(i)})} \label{psitail1} \\
& \le -(i-1) \nonumber
\end{align}
uniformly for $x \in [f^{1+\gamma}(T), (1-\gamma)T]$ for any $\gamma \in (0,(1/\lambda)-1)$ as $T \to \infty$.

(ii) Let $f$ satisfy $f(t) \ge \log(1+t\log^2(t))$ for sufficiently large $t$, and also $f(t) = o(t)$.
Then, for any environment $\nu = (P^{\mu_1},\dots,P^{\mu_K}) \in \mathcal{M}_P^K$ and any sub-optimal arm $i$,
\begin{align}
    \lim_{T \to \infty} \frac{\E_{\nu \pi}\left[ N_i(T) \right]}{f(T)} = \frac{1}{d_P(\mu_i,\mu_*(\nu))}. \label{asymptotic_expected_regret}
\end{align}
\end{proposition}

From (\ref{psitail1}), we see there is an explicit correspondence between the rate of exploration and the resulting heaviness of the regret distribution tail.
For KL-UCB, the exploration function $f$ satisfies $\lim_{t \to \infty} f(t)/\log(t) = 1$.
In Algorithm \ref{alg1}, if we increase the rate of exploration to $1+b$ times the nominal rate of KL-UCB, so that $\lim_{t \to \infty} f(t)/\log(t) = 1+b$ (with $b > 0$), then we obtain a regret tail exponent $\le -(1+b)$.
(See Figure \ref{figure4} in Section \ref{numerical} for numerical illustrations with different values of $b > 0$ when $\mathcal{M}_P$ is a Gaussian family.)
If we further increase the rate of exploration so that $\liminf_{t \to \infty} f(t)/t^a > 0$ for some $a \in (0,1)$, then the regret tail will decay at an exponential rate, as indicated by (\ref{psitail1}).

In summary, we are able to achieve essentially any desired regret tail using Algorithm \ref{alg1} by suitably adjusting the rate of exploration via the exploration function $f$.
Moreover, we will show in Sections \ref{robust1}-\ref{robust2} that an increased rate of exploration/a lighter regret tail provides increased robustness to model mis-specification.
Of course, as indicated by (\ref{asymptotic_expected_regret}), the price to pay is that the expected regret will increase accordingly as we use $f$ with larger growth rates.

\begin{remark} \label{trade-off-tightness}
Together, (\ref{psitail1})-(\ref{asymptotic_expected_regret}) in Proposition \ref{prop5} provide a setting in which the trade-off in Theorem \ref{thm0} between the regret tail in (\ref{tailconsistency}) and expected regret in (\ref{generallowerbound}) is tight.
Specifically, tightness is guaranteed when the model is an exponential family ($\mathcal{M} = \mathcal{M}_P$), and $f$ is a slowly varying and strictly increasing function satisfying a minimum growth rate such as: $f(t) \ge \log(1+t\log^2(t))$ (used to upper bound the expected regret).
In this setting, the exact limit in (\ref{asymptotic_expected_regret}) indicates that the asymptotic lower bound in (\ref{generallowerbound}) is not vacuous.

When $f$ is regularly varying, but not slowly varying, then the trade-off in Theorem \ref{thm0} is no longer tight.
In this case, $\lim_{T \to \infty} f(cT)/f(T) \ne 1$ for any positive $c \ne 1$.
This causes the regret tail upper bound condition in (\ref{tailconsistency}) to be sensitive to any multiplicative scaling of $T$.
Consequently, it is harder to establish a sharp correspondence between a condition such as (\ref{tailconsistency}) and a resulting lower bound for expected regret such as (\ref{generallowerbound}).

Nevertheless, Proposition \ref{prop5} suggests that Theorem \ref{thm0} exhibits the correct dependence on $T$ in the trade-off. 
In particular, given (\ref{tailconsistency}), the left side of (\ref{generallowerbound}) has the correct normalization dependence on $T$, but the resulting constant on the right side may not be tight.
To see this, consider an exponential family model $\mathcal{M}_P$ with base distribution $P$, and let $f(t) = t^a$ with $a \in (0,1)$ (which is regularly varying, but not slowly varying).
Then, Proposition \ref{prop5} indicates that (\ref{psitail1}) holds and also,
\begin{align*}
    \lim_{T \to \infty} \frac{\E_{\nu \pi}\left[ N_i(T) \right]}{T^a} = \frac{1}{d_P(\mu_i,\mu_*(\nu))},
\end{align*}
for each sub-optimal arm $i$ in environment $\nu = (P^{\mu_1},\dots,P^{\mu_K})$.
However, given (\ref{psitail1}), in this setting Theorem \ref{thm0} only indicates that
\begin{align*}
    \liminf_{T \to \infty} \frac{\E_{\nu \pi}\left[ N_i(T) \right]}{T^a} \ge \frac{1}{K^a} \frac{1}{d_P(\mu_i,\mu_*(\nu))},
\end{align*}
which is off by a factor of $K^{-a}$, where $K$ is the number of arms.
\end{remark}

\begin{remark} \label{remark8}
While studying a related problem, \cite{audibert_etal2009} developed finite-time upper bounds on the tail of the regret distribution for the UCB1 algorithm (due to \cite{auer_etal2002}) in the bounded rewards setting, which are suggestive of the exploration-regret tail trade-off that we provide in (\ref{psitail1}).
They study the case where the amount of UCB exploration is increased by a $1+b$ multiplicative factor (similar to setting $f$ such that $\lim_{t \to \infty} f(t)/\log(t) = 1+b$ in Algorithm \ref{alg1}), so that the regret tail becomes lighter with more negative tail exponent $-c \cdot (1+b)$ for some fixed $c > 0$.
However, they do not develop matching lower bounds for the regret tail. 
Such lower bounds are a fundamental ingredient in establishing the nature of the trade-off.
\end{remark}

\begin{remark} \label{remark200}
Furthermore, we can ensure the same regret tail guarantees as in (\ref{psitail1}) for all environments in a class $\mathcal{M}_{P,0}^K$ that is larger than $\mathcal{M}_P^K$.
Here, $\mathcal{M}_{P,0}$ is a general family of distributions, whose CGF's are dominated by those of $\mathcal{M}_P$:
\begin{align}
    \mathcal{M}_{P,0} = \left\{ Q : \mu(Q) \in \mathcal{I}_P, \; \Lambda_Q(\theta) \le \Lambda_{P^{\mu(Q)}}(\theta) \;\; \forall \; \theta \in \mathbb{R} \right\}. \label{domination}
\end{align}
(Recall that $\Lambda_{P^{\mu(Q)}}$ is the CGF of $P^{\mu(Q)}$, the distribution resulting from tilting $P$ to have mean $\mu(Q)$, as in (\ref{model}).)
Below are two examples of $\mathcal{M}_P$ and $\mathcal{M}_{P,0}$.
\end{remark}
\begin{example} \label{ex5}
Let $\mathcal{M}_P$ be the Gaussian family with variance $\sigma^2$.
Then $\mathcal{M}_{P,0}$ is the family of all sub-Gaussian distributions with variance proxy $\sigma^2$.
(We say $Z$ is sub-Gaussian with variance proxy $\sigma^2$ if $\E[e^{\theta (Z - \E[Z])}] \le e^{\sigma^2 \theta^2/2}$ for all $\theta \in \mathbb{R}$.)
\end{example}
\begin{example} \label{ex6}
Let $\mathcal{M}_P$ be the Bernoulli family.
Then $\mathcal{M}_{P,0}$ is the family of all distributions supported on a subset of $[0,1]$.
\end{example}

\subsection{Robustness to Mis-specified Reward Distribution} \label{robust1}

For an $\mathcal{M}_P$-optimized algorithm, if the true reward distributions do not belong in $\mathcal{M}_P$, then the regret tails can be heavier than truncated Cauchy, resulting in expected regret that grows polynomially in the time horizon.
As we saw in Section \ref{mis-specification} via Proposition \ref{wrongdistribution} and Corollary \ref{cor1}, one example of this is when the variance in the design of KL-UCB for Gaussian bandits is just slightly under-specified relative to the true variance.

To alleviate such issues, we can use Algorithm \ref{alg1} with a suitably increased rate of exploration through the choice of $f$, which leads to a lighter regret tail in well-specified settings, as previously shown in Proposition \ref{prop5}.
We will see in Corollary \ref{cor4} below that this also provides protection against distributional mis-specification of the arm rewards.
In particular, by increasing $f$ to be $1+b$ times the nominal rate $\log(1+t\log^2(t))$ (minimum growth rate of $f(t)$) used in part (ii) of Proposition \ref{prop5}, we can preserve logarithmic expected regret for environments from an enlarged class $\mathcal{M}_{P,b}^K$, which is defined in (\ref{mdelta}) below.
Moreover, from part (i) of Proposition \ref{prop5}, in the well-specified case that the environment is from $\mathcal{M}_P^K$, the regret tail will have an exponent $\le -(1+b)$.

The enlarged family of distributions is
\begin{align}
\mathcal{M}_{P,b} = \left\{ Q \, : \, \mu(Q) \in \mathcal{I}_P, \; \Lambda_{Q}(\theta) \le \Psi_{P,b}(\mu(Q),\theta) \;\; \forall \; \theta \in \mathbb{R} \right\}, \label{mdelta}
\end{align}
where for any distribution $Q$ and $z \in \mathcal{I}_Q$, we define:
\begin{align}
    \Psi_{Q,b}(z,\theta) = \frac{\Lambda_{Q^z}((1+b)\theta)}{1+b}, \quad \theta \in \mathbb{R}. \label{psibz}
\end{align}
Setting $b = 0$ recovers $\mathcal{M}_{P,0}$ as in (\ref{domination}).
Using Jensen's inequality and the definition in (\ref{psibz}), it is straightforward to see that $\mathcal{M}_P \subsetneqq \mathcal{M}_{P,b}$ for $b > 0$.
Moreover, with $\mathcal{M}_{P,0}$ from (\ref{domination}) and any $b' > b > 0$, we have $\mathcal{M}_{P,0} \subsetneqq \mathcal{M}_{P,b} \subsetneqq \mathcal{M}_{P,b'}$.
An example of $\mathcal{M}_P$ and $\mathcal{M}_{P,b}$ is the following.
\begin{example} \label{ex7} 
Let $\mathcal{M}_P$ be the Gaussian family with variance $\sigma^2$.
Then $\mathcal{M}_{P,b}$ is the family of all sub-Gaussian distributions with variance proxy $\sigma^2(1+b)$.
(Also, $\mathcal{M}_{P,0}$ is the family of all sub-Gaussian distributions with variance proxy $\sigma^2$, as we saw in Example \ref{ex5}.)
\end{example}

The proof of Corollary \ref{cor4} follows from a straightforward adaptation of the proof of (\ref{asymptotic_expected_regret}) in Proposition \ref{prop5}.
The details are provided in Appendix \ref{robustnessproofs}.

\begin{corollary} \label{cor4}
Let $\pi$ be Algorithm \ref{alg1}, with divergence $d_P$ and exploration function $f$.
Let $f$ satisfy $f(t) \ge (1+b) \log(1+t\log^2(t))$ with $b \ge 0$ for sufficiently large $t$, and also $f(t) = o(t)$.
Then, for any environment $\nu = (Q_1,\dots,Q_K) \in \mathcal{M}_{P,b}^K$ and each sub-optimal arm $i$,
\begin{align}
\lim_{T \to \infty} \frac{\E_{\nu \pi}\left[ N_i(T) \right]}{f(T)} = \frac{1}{d_P(\mu(Q_{i}),\mu_*(\nu))}. \label{psiregret2}
\end{align}
\end{corollary}

\subsection{Robustness to Mis-specified Reward Dependence Structure} \label{robust2}

In this section, we consider arm rewards taking values in a finite set $S \subset \mathbb{R}$.
Let $P$ be a distribution on $S$.
Even if the marginal distributions of arm rewards belong in the exponential family $\mathcal{M}_P$, the serial dependence structure of rewards could be mis-specified, which can result in regret tails that are heavier than truncated Cauchy, and expected regret that grows polynomially in the time horizon.
We saw an example of this in Section \ref{dependent} via Proposition \ref{markovian}, particularly via Example \ref{ex4}.

To alleviate such issues, we can use Algorithm \ref{alg1} with a suitably increased rate of exploration through the choice of $f$, which leads to a lighter regret tail in well-specified settings, as previously shown in Proposition \ref{prop5}.
We will see in Corollary \ref{cor5} below that this also provides protection against Markovian departures from independence of the arm rewards.
In particular, by increasing $f$ to be $1+b$ times the nominal rate $\log(1+t\log^2(t))$ (minimum growth rate of $f(t)$) used in part (ii) of Proposition \ref{prop5}, we can preserve logarithmic expected regret when the arm rewards evolve as Markov chains with transition matrices from a set $\widetilde{\mathcal{M}}_{P,b}$, which is defined in (\ref{widetildeMb}) below.
Moreover, as in the previous section, in the well-specified case that the environment is from $\mathcal{M}_P^K$, the regret tail will have an exponent $\le -(1+b)$.

Let $\mathcal{S}_{\abs{S}}$ denote the set of $\abs{S} \times \abs{S}$ irreducible stochastic matrices satisfying the Doeblin Condition (as discussed in Section \ref{dependent} in the context of Proposition \ref{markovian}).
We define
\begin{align}
    \widetilde{\mathcal{M}}_{P,b} = \left\{ H \in \mathcal{S}_{\abs{S}} \, : \, \phi_H(\theta) \le \Psi_{P,b}(\phi_H'(0),\theta) \;\; \forall \; \theta \in \mathbb{R} \right\}, \label{widetildeMb}
\end{align}
and we recall that $\phi_H(\theta)$ is the logarithm of the Perron-Frobenius eigenvalue of the tilted version (as in (\ref{Qtilted1})) of transition matrix $H$, and $\phi_H'(0)$ is the equilibrium  mean of a chain with transition matrix $H$.
Of course, the exponential family $\mathcal{M}_P$ is equivalent to a strict subset of the collection of transition matrices with identical rows in $\widetilde{\mathcal{M}}_{P,b}$, for any $b > 0$.
Also, for any $b' > b > 0$, $\widetilde{\mathcal{M}}_{P,b} \subsetneqq \widetilde{\mathcal{M}}_{P,b'}$.
In Example \ref{ex8}, which is given after Corollary \ref{cor5}, we examine the degree to which $\widetilde{\mathcal{M}}_{P,b}$ is ``larger'' than $\mathcal{M}_P$ when $S = \{0,1\}$ and $\mathcal{M}_P$ is the Bernoulli family.

Like for Corollary \ref{cor4}, the proof of Corollary \ref{cor5} also follows from a straightforward adaptation of the proof of (\ref{asymptotic_expected_regret}) in Proposition \ref{prop5}.
The details are provided in Appendix \ref{robustnessproofs}.

\begin{corollary} \label{cor5}
Let $\pi$ be Algorithm \ref{alg1}, with divergence $d_P$ and exploration function $f$.
Let $f$ satisfy $f(t) \ge (1+b) \log(1+t\log^2(t))$ with $b \ge 0$ for sufficiently large $t$, and also $f(t) = o(t)$.
For the $K$-armed environment $\nu$, suppose arm $i$ yields rewards that evolve according to a Markov chain with transition matrix $H_i \in \widetilde{\mathcal{M}}_{P,b}$.
Then, for any sub-optimal arm $i$,
\begin{align}
\lim_{T \to \infty} \frac{\E_{\nu \pi}\left[ N_{i}(T) \right]}{f(T)} = \frac{1}{d_P(\phi_{H_{i}}'(0),\phi_{H_{r(1)}}'(0))}. \label{psiregret4}
\end{align}
\end{corollary}

\begin{example} \label{ex8}
Let the state space $S = \{0,1\}$, and let $\mathcal{M}_P$ be the Bernoulli family of distributions.
Consider transition matrices on $S$ of the form:
\begin{align}
    H = \begin{bmatrix}
    1-q \;\;\;\;\; & q \\
    1-q' \;\;\;\;\; & q'
    \end{bmatrix}.
    \label{q2}
\end{align}
The more positive the difference $q' - q$, the more positive the autocorrelation between the rewards.
In Table \ref{table2} below, for different values of $b > 0$, we examine how positive the difference $q' - q$ can be in order for $H$ to still belong in $\widetilde{\mathcal{M}}_{P,b}$, and thus for Corollary \ref{cor5} to be applicable.
As the targeted regret tail exponent $-(1+b)$ is made more negative, the algorithm can withstand more positive autocorrelation between the rewards and still maintain logarithmic expected regret.
\begin{table}[H]
\centering
\begin{tabular}{ccccccccccc}
\toprule
$-(1+b) \qquad$ & -2 & -3 & -4 & -5 & -6 & -7 & -8 & -9 & -10 & -11 \\
$\text{max allowed } q' - q \qquad$ & 0.18 & 0.36 & 0.49 & 0.59 & 0.65 & 0.70 & 0.74 & 0.77 & 0.80 & 0.82 \\
\bottomrule
\end{tabular}
\caption{\label{table2} For particular $-(1+b)$ values (upper bound on the regret tail exponent), and for the restriction $q, q' \in [0.05, 0.95]$, we give the maximum allowed difference $q' - q$ that ensures the transition matrix $H$ in (\ref{q2}) belongs in $\widetilde{\mathcal{M}}_{P,b}$, as defined in (\ref{widetildeMb}) (and (\ref{psibz})).}
\end{table}
\end{example}

\begin{remark} \label{remark6}
For general reward processes satisfying Assumptions \ref{A3}-\ref{A4}, e.g., general Markov processes, there are no finite-sample concentration bounds.
So there does not seem to be a universal way to obtain an upper bound on the regret tail.
For such reward processes, there also does not seem to be a universal way to obtain upper bounds on expected regret such as in Corollary \ref{cor5}, and thus there are no provable robustness benefits for our procedure to lighten the regret tail.
Nevertheless, our simulations in Figure \ref{figure5} in Section \ref{numerical} suggest that we can still ensure the regret tail is lighter to a desired level using our procedure.
(So, the lower bound in Theorem \ref{gartnerellis} seems to be tight in greater generality than what we are able to provably show.)

\end{remark}

\section{Numerical Experiments} \label{numerical}

In this section, we use numerical experiments to verify that our asymptotic approximations for the regret distribution tail hold over finite time horizons.
For each experiment, we perform (statistically) independent simulation runs of the bandit system, and we compute empirical probabilities for the regret tail.

\begin{figure}
    \centering
    \hspace{7mm}
    \includegraphics[width=0.85\textwidth, trim={0cm 5cm 0cm 5.5cm}, clip]{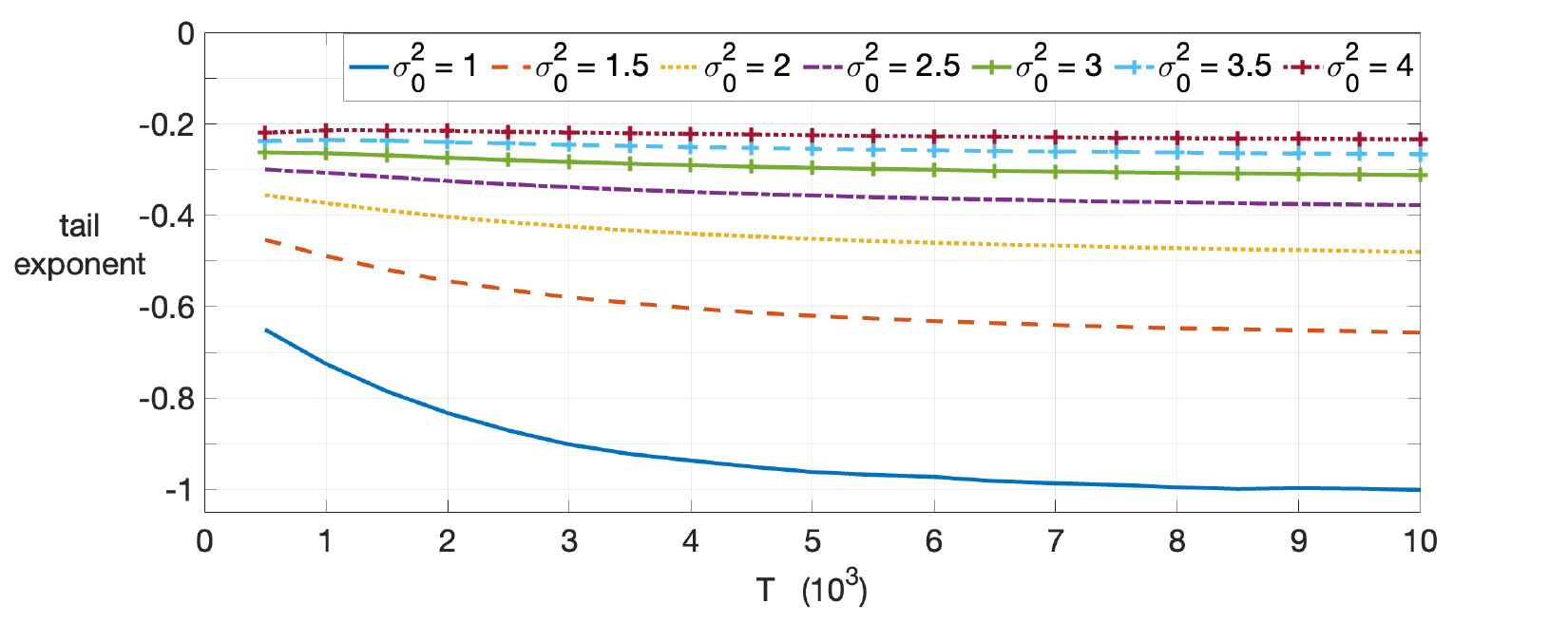}
    \caption{Plot of $\log \P_{\nu \pi}(N_2(T) \ge 0.8 T)/\log(T)$ vs $T$. Environment $\nu = (N(0.1,\sigma_0^2),N(0,\sigma_0^2))$. Algorithm $\pi$ is KL-UCB for iid unit-variance Gaussian rewards. The curves correspond to the cases $\sigma_0^2 = 1,1.5,\dots,4$, as indicated by the legend. The curves asymptote to $-1/\sigma_0^2$ in each case, which agrees with (\ref{varianceratio}) in Corollary \ref{cor1}. To generate each curve, $2 \times 10^6$ simulation runs were used.
    }
    \label{figure1}
\end{figure}

\begin{figure}
    \centering
    \hspace{7mm}
    \includegraphics[width=0.85\textwidth, trim={0cm 5cm 0cm 5.5cm}, clip]{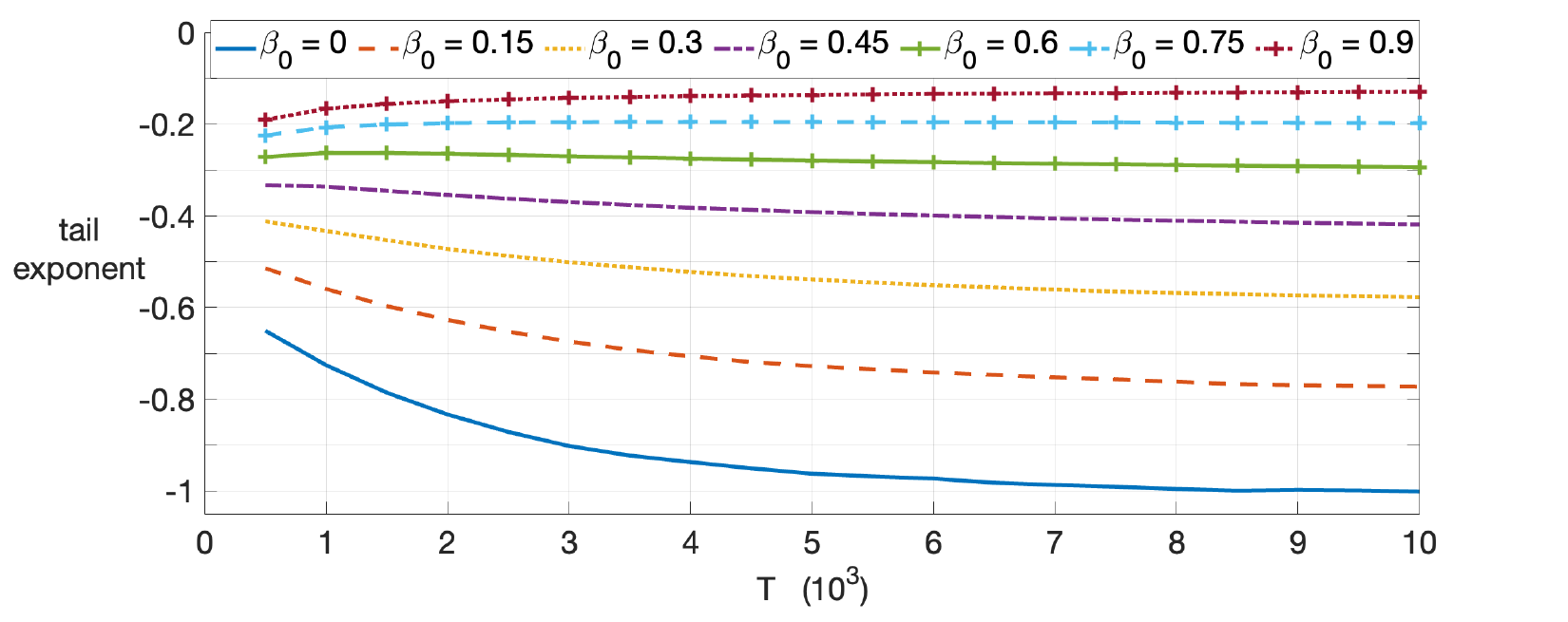}
    \caption{Plot of $\log \P_{\nu \pi}(N_2(T) \ge 0.8 T)/\log(T)$ vs $T$. Environment $\nu$ consists of two Gaussian AR(1) processes with common AR coefficient $\beta_0$, and equilibrium distributions $(N(0.1,1),N(0,1))$. Algorithm $\pi$ is KL-UCB for iid unit-variance Gaussian rewards. The curves correspond to the cases $\beta_0 = 0,0.15,\dots,0.9$, as indicated by the legend. The curves approximately asymptote to $-(1-\beta_0)/(1+\beta_0)$, which agrees with the lower bound in Corollary \ref{cor2} and (\ref{arvarianceratio2}). To generate each curve, $2 \times 10^6$ simulation runs were used.
    }
    \label{figure2}
\end{figure}

In Figure \ref{figure1}, we examine the validity of Theorem \ref{thm1} and Corollary \ref{cor1}.
For all curves but the dark blue one, the variance of the Gaussian KL-UCB algorithm is set smaller than that of the actual Gaussian reward distributions.
In Figure \ref{figure2}, we examine the validity of Corollary \ref{cor2}.
For all curves but the dark blue one, the Gaussian KL-UCB algorithm does not take into account the AR(1) serial dependence structure of the rewards, even though the algorithm is perfectly matched to the marginal distributions of the rewards.
In both Figures \ref{figure1} and \ref{figure2}, the regret tail probabilities in mis-specified cases correspond to regret distribution tails that are heavier than truncated Cauchy. 

\begin{figure}
\centering
\begin{tabular}{c}
  \includegraphics[width=0.75\linewidth, trim={0cm 7cm 0cm 7.5cm}, clip]{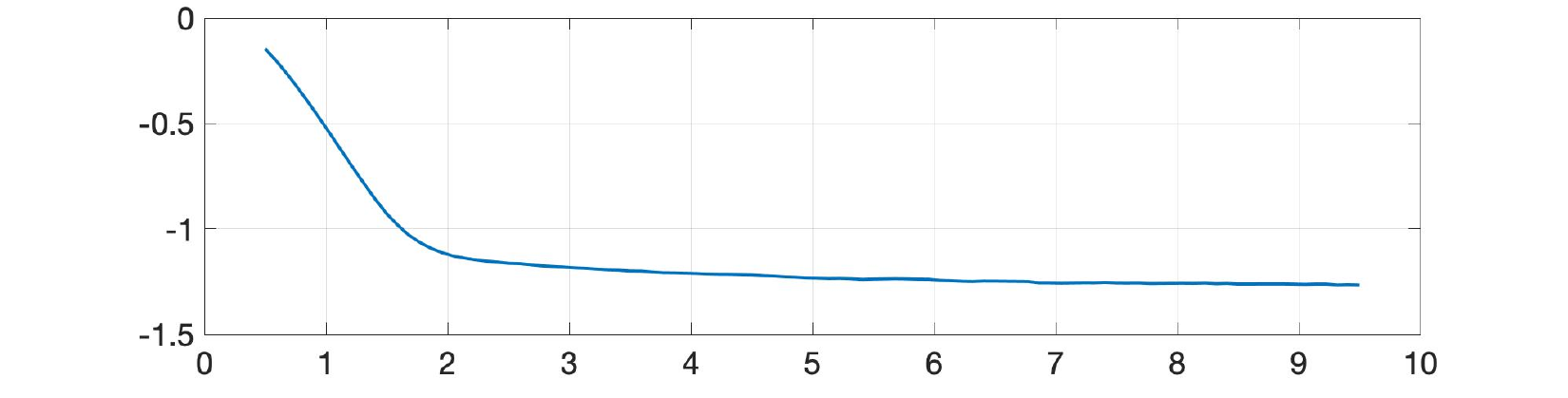} \\
  \includegraphics[width=0.75\linewidth, trim={0cm 7cm 0cm 7.5cm}, clip]{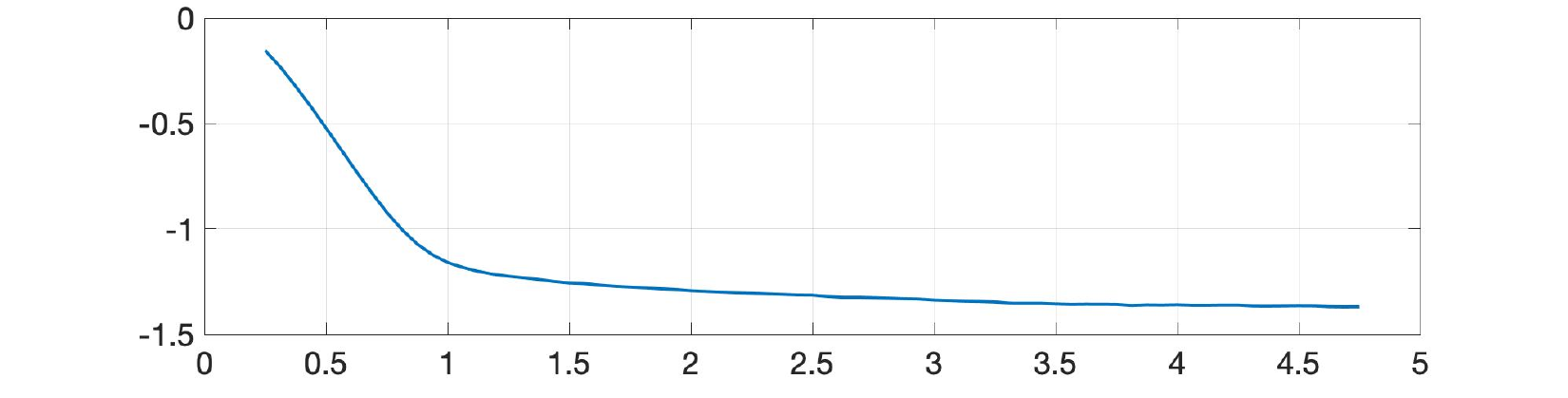} \\
  \includegraphics[width=0.75\linewidth, trim={0cm 6.5cm 0cm 7cm}, clip]{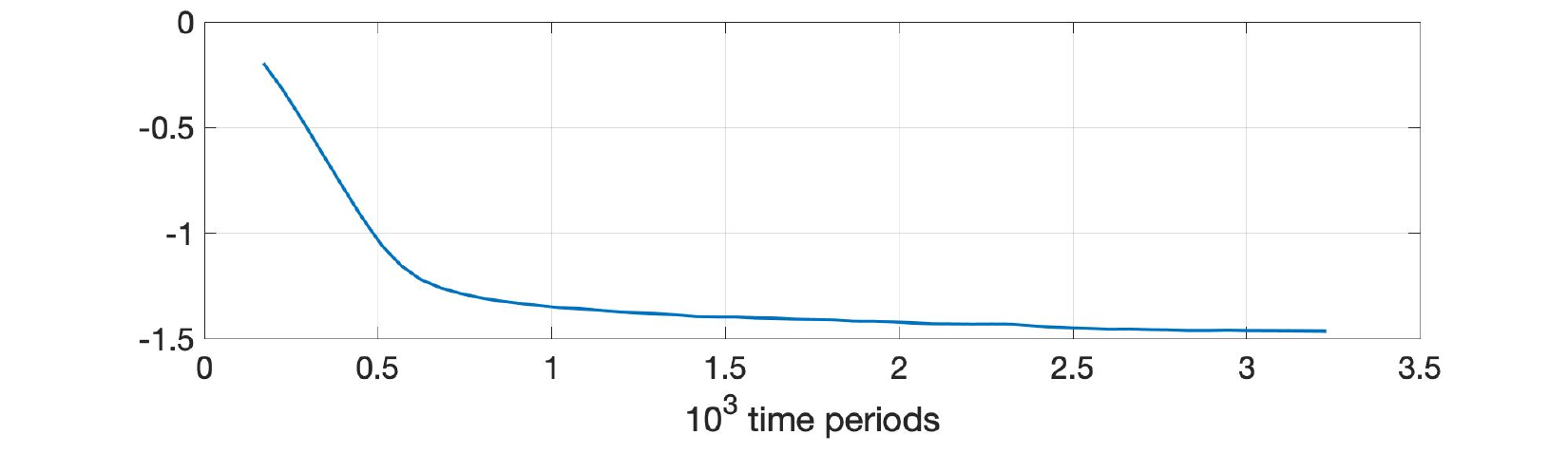} \\  
\end{tabular}
\caption{Plot of $\log \P_{\nu \pi}(N_2(T) > x)/\log(x)$ vs $x$ for $x \in [0.05 T, 0.95 T]$ (with time horizon $T$ fixed). Environment $\nu = (\text{Ber}(q),\text{Ber}(0.4))$. Algorithm $\pi$ is KL-UCB for iid Bernoulli rewards. Top: $q = 0.475$, $T = 10^4$; Middle: $q = 0.5$, $T = 5 \times 10^3$; Bottom: $q = 0.525$, $T = 3.4 \times 10^3$. Each curve asymptotes to $\lim_{z \downarrow 0} d_P(z,q)/d_P(z,0.4)$ (with values $-1.26$ (top), $-1.36$ (middle), $-1.46$ (bottom)), as specified by Theorem \ref{generalupperbound} and (\ref{klequivalence}). 
To generate each curve, $8 \times 10^6$ simulation runs were used.
}
\label{figure3}
\end{figure}

In Figure \ref{figure3}, we verify that when the arms are iid Bernoulli, KL-UCB produces regret distribution tails which are strictly lighter than truncated Cauchy, as predicted by Theorem \ref{generalupperbound}.

\begin{figure}
    \centering
    \hspace{7mm}
    \includegraphics[width=0.85\textwidth, trim={0cm 5.5cm 0cm 6cm}, clip]{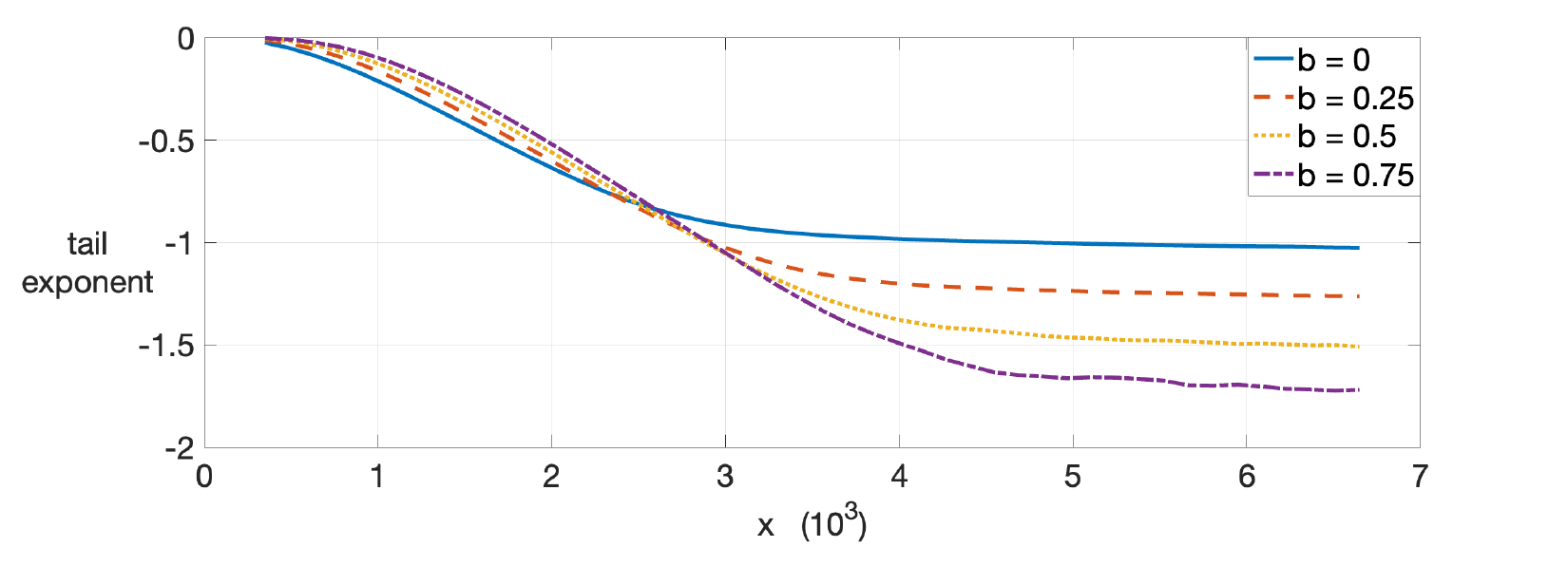}
    \caption{Plot of $\log \P_{\nu \pi}(N_2(T) > x)/\log(x)$ vs $x$ for $x \in [0.05 T, 0.95 T]$, with fixed time horizon $T = 7 \times 10^3$. Environment $\nu = (N(0.1,1),N(0,1))$. $\pi$ is Algorithm \ref{alg1} with KL divergence $d_P$ between unit-variance Gaussian distributions, and $f(t) = (1+b)\log(t)$ (to aim for a regret tail exponent of $-(1+b)$). 
    The curves correspond to the cases $b = 0,0.25,0.5,0.75$, as indicated by the legend. 
    As predicted by (\ref{psitail1}) in Proposition \ref{prop5}, the curves asymptote to $-1$, $-1.25$, $-1.5$, $-1.75$. To generate each curve, $4 \times 10^7$ simulation runs were used.
    }
    \label{figure4}
\end{figure}

In Figure \ref{figure4}, we demonstrate the trade-off established in (\ref{psitail1}) of Proposition \ref{prop5} between the amount of UCB exploration and the resulting exponent of the regret distribution tail, with $f(t) = (1+b)\log(t)$ and $b \ge 0$ in Algorithm \ref{alg1}.

\begin{figure}
    \centering
    \hspace{7mm}
    \includegraphics[width=0.85\textwidth, trim={0cm 5.5cm 0cm 6cm}, clip]{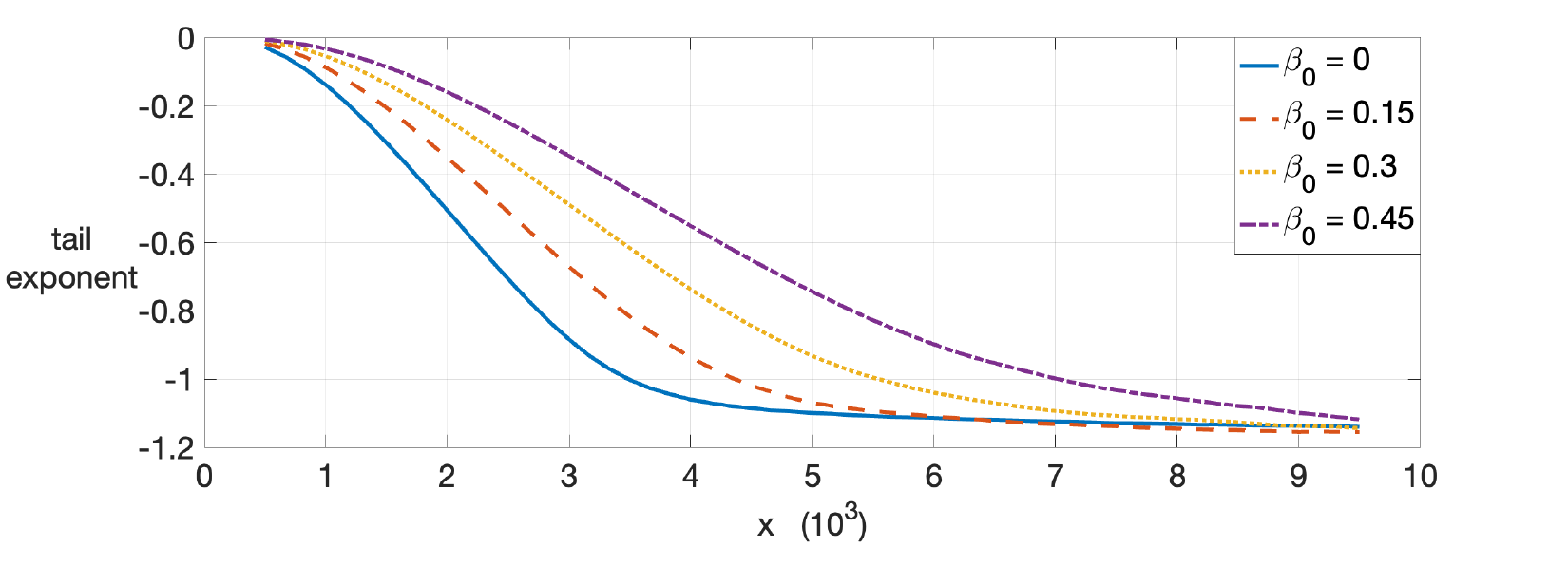}
    \caption{Plot of $\log \P_{\nu \pi}(N_2(T) > x)/\log(x)$ vs $x$ for $x \in [0.05 T, 0.95 T]$, with fixed time horizon $T = 10^4$. Environment $\nu$ consists of two Gaussian AR(1) processes with common AR coefficient $\beta_0$, and equilibrium distributions $(N(0.1,1),N(0,1))$. $\pi$ is Algorithm \ref{alg1} with KL divergence $d_P$ between unit-variance Gaussian distributions, and $f(t) = (1+b)\log(t)$ with $1+b = 1.1 \cdot \frac{1+\beta_0}{1-\beta_0}$ (to aim for a regret tail exponent of $\approx -1.1$ in each case of $\beta_0$). The curves correspond to the cases $\beta_0 = 0,0.15,0.3,0.45$, as indicated by the legend.
    All curves asymptote to (slightly less than) $-1.1$, as desired. To generate each curve, $4 \times 10^7$ simulation runs were used.
    }
    \label{figure5}
\end{figure}

In Figure \ref{figure5}, we demonstrate that the poor regret tail properties resulting from mis-specification of the serial dependence structure of the rewards can be overcome by aiming for a lighter regret tail using Algorithm \ref{alg1}.
Here, we use the same AR(1) setup that is illustrated in Figure \ref{figure2}.
As discussed in the first paragraph of Remark \ref{remark6}, here we do not have upper bounds on regret tail probabilities (only lower bounds in (\ref{arvarianceratio2})), and thus there are no provable robustness guarantees.
However, we show empirically in Figure \ref{figure5} that aiming for a lighter regret tail still provides robustness to mis-specification in this setting.
The $\frac{1+\beta_0}{1-\beta_0}$ factor in Figure \ref{figure5} is taken from the lower bound in (\ref{arvarianceratio2}), which we essentially confirm to be tight here.

\section{Acknowledgments}
We thank the Area Editor, Ramandeep Randhawa, the Associate Editor, and two Reviewers for their valuable suggestions, which significantly improved the paper.
We also thank Achal Bassamboo and Nalin Shani for a close reading of some of the technical arguments, which led to more streamlined arguments.
We are indebted to Professor Tze Leung Lai, who sadly passed away in May 2023, for paving the way in numerous fundamental areas of research, extending well beyond the multi-armed bandit problem studied here.
Peter Glynn is also grateful to have had Tze Leung as a colleague and friend for many years.




\begin{APPENDICES}

\section{Proof of Theorem \ref{thm1}} \label{thm1_proof}

\proof{Proof of Theorem \ref{thm1}.}
Without loss of generality, suppose that $\mu_1 > \mu_2 > \dots > \mu_K$ (i.e., $r(i) = i$ for all $i \in [K]$) in the environment $\nu = (P^{\mu_1},P^{\mu_2},\dots,P^{\mu_K})$.
We first show (\ref{conclusion0}) and (\ref{conclusion1}) for the second-best arm $i=2$.
Consider the alternative environment $\widetilde{\nu} = (P^{\widetilde{\mu}_1},P^{\mu_2},\dots,P^{\mu_K})$, where $\widetilde{\mu}_1 < \mu_2$, and $\mu_2,\dots,\mu_K$ are the same mean values from the original environment $\nu$.
(Arm 2 is the best arm in $\widetilde{\nu}$.
Later in the proof, we will consider different values for $\widetilde{\mu}_1$, subject to $\widetilde{\mu}_1 < \mu_2$ and $\widetilde{\mu}_1 \in \mathcal{I}_P$.)
Let $\delta > 0$, and define the events:
\begin{align*}
& \mathcal{A}_T = \left\{ \abs{ \frac{N_1(T)}{\log(T)} - \frac{1}{d_P(\widetilde{\mu}_1,\mu_2)} } \le \delta \right\} \cap \left\{ \abs{ \frac{N_j(T)}{\log(T)} - \frac{1}{d_P(\mu_j,\mu_2)} } \le \delta, \; \forall \; j \ge 3 \right\} \\
& \mathcal{B}_T = \left\{\abs{\frac{1}{N_1(T)} \sum_{t=1}^{N_1(T)} \log \frac{dP^{\mu_1}}{dP^{\widetilde{\mu}_1}}(X_1(t)) + d_P(\widetilde{\mu}_1,\mu_1)} \le \delta \right\}.
\end{align*}
By a change of measure from $\nu$ to $\widetilde{\nu}$,
\begin{align}
\P_{\nu \pi}(N_2(T) > (1-\gamma)T) & = \E_{\widetilde{\nu} \pi}\left[ \indic{N_2(T) > (1-\gamma)T} \prod_{t=1}^{N_1(T)} \frac{dP^{\mu_1}}{dP^{\widetilde{\mu}_1}}(X_1(t)) \right] \label{begin} \\ 
& \ge \E_{\widetilde{\nu} \pi}\left[ \indic{\mathcal{A}_T,\mathcal{B}_T} \exp\left( \frac{1}{N_1(T)} \sum_{t=1}^{N_1(T)} \log \frac{dP^{\mu_1}}{dP^{\widetilde{\mu}_1}}(X_1(t)) \cdot N_1(T) \right) \right] \label{smallerevents} \\
& \ge \P_{\widetilde{\nu} \pi}(\mathcal{A}_T,\mathcal{B}_T) \cdot \exp\left( -\left(d_P(\widetilde{\mu}_1,\mu_1) + \delta \right) \left(\frac{1}{d_P(\widetilde{\mu}_1,\mu_2)} + \delta \right) \log(T) \right). \label{lln2}
\end{align}
where (\ref{smallerevents}) follows from $\{N_2(T) > (1-\gamma)T\} \supset \mathcal{A}_T$ for sufficiently large $T$, and (\ref{lln2}) follows from lower bounds using $\mathcal{A}_T$ and $\mathcal{B}_T$.
From (\ref{lln2}), taking logs and dividing by $\log(T)$,
\begin{align}
\frac{\log \P_{\nu \pi}(N_2(T) > (1-\gamma)T)}{\log(T)} \ge \frac{\log \P_{\widetilde{\nu} \pi}(\mathcal{A}_T,\mathcal{B}_T)}{\log(T)} -\left(d_P(\widetilde{\mu}_1,\mu_1) + \delta \right) \left(\frac{1}{d_P(\widetilde{\mu}_1,\mu_2)} + \delta \right). \label{normalized}
\end{align}
Using Lemma \ref{lemma2} together with the WLLN for sample means, we have $\lim_{T \to \infty} \P_{\widetilde{\nu} \pi}(\mathcal{A}_T,\mathcal{B}_T) = 1$.
So the first term on the right side of (\ref{normalized}) is negligible as $T \to \infty$, and upon sending $\delta \downarrow 0$ and optimizing with respect to $\widetilde{\mu}_1$, we have
\begin{align}
\liminf_{T \to \infty} \frac{\log \P_{\nu \pi}(N_2(T) > (1-\gamma)T)}{\log(T)} & \ge - \inf_{\widetilde{\mu}_1 \in \mathcal{I}_P \, : \, \widetilde{\mu}_1 < \mu_2} \frac{d_P(\widetilde{\mu}_1,\mu_1)}{d_P(\widetilde{\mu}_1,\mu_2)}. \label{optimizedklratio}
\end{align}
Then, the conclusion (\ref{conclusion0}) with the infimum over $B_\gamma(T) = [\log^{1+\gamma}(T),(1-\gamma)T]$ follows from part (i) of Lemma \ref{lem2} (provided after the current proof) with the choice $g(t) = \log^{1+\gamma}(t)$.

We now establish (\ref{conclusion1}).
Let $\gamma \in (0,1)$.
In the context of Lemma \ref{lem2}, take $g(t) = t^\gamma$.
Because $P$ is discrimination equivalent, the right side of (\ref{optimizedklratio}) is equal to $-1$, which establishes part (i) of Lemma \ref{lem2}.
Also, since $\pi$ is $\mathcal{M}_P$-optimized, using Markov's inequality,
\begin{align*}
    \limsup_{T \to \infty} \frac{\log \P_{\nu \pi}(N_2(T) > T^\gamma)}{\log(T^\gamma)} \le -1,
\end{align*}
which establishes part (ii) of Lemma \ref{lem2}.
Then, the desired uniform convergence in (\ref{conclusion1}) for $x \in [T^\gamma,(1-\gamma)T]$ follows from part (iii) of Lemma \ref{lem2}.

We now show (\ref{conclusion0}) for any sub-optimal arm $i \ge 3$.
Consider the alternative environment $\widetilde{\nu} = (P^{\widetilde{\mu}_1},\dots,P^{\widetilde{\mu}_{i-1}},P^{\mu_i},\dots,P^{\mu_K})$, where $\widetilde{\mu}_j < \mu_i$ for all $j \le i-1$, and $\mu_i,\dots,\mu_K$ are the same mean values from the original environment $\nu$. (Arm $i$ is now the best arm in $\widetilde{\nu}$.)
We change the events $\mathcal{A}_T$ and $\mathcal{B}_T$ to:
\begin{align*}
\mathcal{A}_T & = \left\{ \abs{ \frac{N_j(T)}{\log(T)} - \frac{1}{d_P(\widetilde{\mu}_j,\mu_i)} } \le \delta, \; \forall \; j \le i-1 \right\} \cap \left\{ \abs{ \frac{N_j(T)}{\log(T)} - \frac{1}{d_P(\mu_j,\mu_i)} } \le \delta, \; \forall \; j \ge i+1 \right\} \\
\mathcal{B}_T & = \left\{\abs{\frac{1}{N_j(T)} \sum_{t=1}^{N_j(T)} \log \frac{dP^{\mu_j}}{dP^{\widetilde{\mu}_j}}(X_j(t)) + d_P(\widetilde{\mu}_j,\mu_j)} \le \delta, \; \forall \; j \le i-1 \right\}.
\end{align*}
To obtain (\ref{conclusion0}) for sub-optimal arm $i \ge 3$, we can then run through arguments analogous to those in (\ref{begin})-(\ref{optimizedklratio}).
Here, the change of measure from $\nu$ to $\widetilde{\nu}$ involves the product of $i-1$ likelihood ratios corresponding to the arms $1,\dots,i-1$.
Each of the parameter values $\widetilde{\mu}_1,\dots,\widetilde{\mu}_{i-1}$ can be optimized separately (subject to $\widetilde{\mu}_j < \mu_i$ and $\widetilde{\mu}_j \in \mathcal{I}_P$ for all $j \le i-1$) to yield the desired conclusion.
\halmos
\endproof

\vspace{4mm} 

We now introduce Lemma \ref{lem2}, which provides a unified way to establish \textit{uniform convergence} for the regret tail characterizations developed throughout the paper.

\begin{lemma} \label{lem2}
    Let $\nu$ be any bandit environment, and let $i$ be a sub-optimal arm in $\nu$.
    
    (i) Suppose for some $c_i(\nu) > 0$ and some $\gamma \in (0,1)$,
    \begin{align}
        \liminf_{T \to \infty} \frac{\log P_{\nu \pi}(N_i(T) > (1-\gamma)T)}{\log(T)} \ge -c_i(\nu). \label{hypothesis1}
    \end{align}
    Then,
    \begin{align}
        \liminf_{T \to \infty} \inf_{x \in B_\gamma(T)} \frac{\log P_{\nu \pi}(N_i(T) > x)}{\log(x)} \ge -c_i(\nu), \label{result1}
    \end{align}
    with $B_\gamma(T) = [g(T), (1-\gamma)T]$, and any strictly increasing function $g : (1,\infty) \to (0,\infty)$ such that $\lim_{t \to \infty} g(t)/\log(t) = \infty$ and $g(t) = o(t)$.
    
    (ii) Suppose,
    \begin{align}
        \limsup_{T \to \infty} \frac{\log P_{\nu \pi}(N_i(T) > g(T))}{\log(g(T))} \le -c_i(\nu). \label{hypothesis2}
    \end{align}
    Then, 
    \begin{align}
    \limsup_{T \to \infty} \sup_{x \in B_\gamma(T)} \frac{\log P_{\nu \pi}(N_i(T) > x)}{\log(x)} \le -c_i(\nu). \label{result2}
    \end{align}
    
    (iii) If both (\ref{hypothesis1}) and (\ref{hypothesis2}) hold, then together (\ref{result1}) and (\ref{result2}) yield:
    \begin{align*}
        \lim_{T \to \infty} \frac{\log P_{\nu \pi}(N_i(T) > x)}{\log(x)} = -c_i(\nu)
    \end{align*}
    uniformly for $x \in B_\gamma(T)$ as $T \to \infty$.
\end{lemma}

\indent \textit{Proof of Lemma \ref{lem2}.} \vspace{2mm} \\
\underline{Part (i)} \\
Since $t \mapsto N_i(t)$ is non-decreasing with $\P_{\nu \pi}$-probability one, we have for sufficiently large $T$ and all $x \in B_\gamma(T) = [g(T),(1-\gamma)T]$,
\begin{align*}
    \P_{\nu \pi}(N_i(T) > x) \ge \P_{\nu \pi}(N_i(\lceil x/(1-\gamma) \rceil) > x).
\end{align*}
So, for any $\epsilon > 0$, we have for sufficiently large $T$,
\begin{align}
    \frac{\log \P_{\nu \pi}(N_i(T) > x)}{\log(x)} & \ge \frac{\log \P_{\nu \pi}(N_i(\lceil x/(1-\gamma) \rceil) > x)}{\log(x)} \label{result_3} \\
    & \ge -c_i(\nu)(1+\epsilon), \label{result3}
\end{align}
uniformly for all $x \in B_\gamma(T)$, where (\ref{result3}) follows from the convergence result in (\ref{hypothesis1}). 
Then, (\ref{result1}) is established by taking the infimum over $x \in B_\gamma(T)$ on the left side of (\ref{result_3}), sending $T \to \infty$, and then $\epsilon \downarrow 0$. \vspace{2mm} \\
\noindent \underline{Part (ii)} \\
The function $g$ is strictly increasing, so it has an inverse $g^{-1}$ (defined on the range of $g$), which is also strictly increasing.
Since $t \mapsto N_i(t)$ is non-decreasing with $\P_{\nu \pi}$-probability one, we have for sufficiently large $T$ and all $x \in B_\gamma(T) = [g(T),(1-\gamma)T]$,
\begin{align*}
    \P_{\nu \pi}(N_i(\lfloor g^{-1}(x) \rfloor) > x) \ge \P_{\nu \pi}(N_i(T) > x).
\end{align*}
So, for any $\epsilon \in (0,1)$, we have for sufficiently large $T$,
\begin{align}
    -c_i(\nu)(1-\epsilon) & \ge \frac{\log \P_{\nu \pi}(N_i(\lfloor g^{-1}(x) \rfloor) > x)}{\log(x)} \label{result4} \\
    & \ge \frac{\log \P_{\nu \pi}(N_i(T) > x)}{\log(x)}, \label{result_4}
\end{align}
uniformly for all $x \in B_\gamma(T)$, where (\ref{result4}) follows from the convergence result in (\ref{hypothesis2}).
Then, (\ref{result2}) is established by taking the supremum over $x \in B_\gamma(T)$ on the right side of (\ref{result_4}), sending $T \to \infty$, and then $\epsilon \downarrow 0$.
\halmos
\endproof

\section{Proof of Theorem \ref{generalupperbound}} \label{generalupperbound_proof}

Define for arm $i$ the UCB index at time $t$, given that arm $i$ has been played $n$ times:
\begin{align*}
\widetilde{U}_i(n,t) = \sup \left\{z \in \mathcal{I}_P : d_P(\widehat{\mu}_i(\tau_i(n)),z) \le \frac{f(t)}{n} \right\},
\end{align*}
where $\tau_i(n)$ is the time of the $n$-th play of arm $i$.
The ``exploration'' function $f(t)$ is a design choice.
For KL-UCB (as introduced in Algorithm 2 of \cite{cappe_etal2013}), choices include $f(t) = \log(t)$ (as in Section 7 of \cite{cappe_etal2013}), $f(t) = \log(t) + 3 \log\log(t)$ (as in Theorem 1 of \cite{cappe_etal2013}), and $f(t) = \log(1+t\log^2(t))$ (as in Theorem 10.6 of \cite{lattimore_etal2020}).
We will leave the particular form for $f(t)$ unspecified in developing the upper bound part of this proof.
We will actually establish a more general upper bound part.
Specifically, we consider any increasing function $f : (1,\infty) \to (0,\infty)$ such that $\liminf_{t \to \infty} f(t)/\log(t) \ge 1$ and $f(t) = o(t^\lambda)$ for some $\lambda \in (0,1)$. \vspace{2mm} \\
\indent \textit{Proof of Theorem \ref{generalupperbound}.} 
Without loss of generality, suppose that $\mu_1 > \mu_2 > \dots > \mu_K$ (i.e., $r(i) = i$ for all $i \in [K]$) in the environment $\nu = (P^{\mu_1},P^{\mu_2},\dots,P^{\mu_K})$. \\
\underline{Upper Bound} \\
We first consider the sub-optimal arm $i = 2$.
Let $x_T = \floor{f^{1+\gamma}(T)}$ with any fixed $\gamma \in (0,(1/\lambda)-1)$ (with $\lambda \in (0,1)$ as specified above).
Also, let $\delta \in (0,\mu_1 - \mu_2)$.
We have the following bounds:
\begin{align}
\P_{\nu \pi} \left( N_2(T) > x_T \right) & \le \P_{\nu \pi}\Bigl( \exists \; t \in (\tau_2(x_T),T] \;\;\; \text{s.t.} \;\;\; \widetilde{U}_1(N_1(t-1),t-1) \le \widetilde{U}_2(N_2(t-1),t-1) \Bigr) \label{step3} \\ 
& \le \P_{\nu \pi}\Bigl( \exists \; t \in (x_T,T] \;\;\; \text{s.t.} \;\;\; \widetilde{U}_1(N_1(t-1),x_T) \le \widetilde{U}_2(x_T,T) \Bigr) \nonumber \\ 
& \le \P_{\nu \pi}\Bigl( \exists \; t \in (x_T,T] \;\;\; \text{s.t.} \;\;\; \widetilde{U}_1(N_1(t-1),x_T) \le \mu_2 + \delta \Bigr) \label{term6} \\ 
& \quad + \P_{\nu \pi}\Bigl( \widetilde{U}_2(x_T,T) > \mu_2 + \delta \Bigr). \label{term7}
\end{align}
Note that (\ref{step3}) holds because $N_2(T) > x_T$ is the event of interest, and so after the $x_T$-th play of arm $2$ at time $\tau_2(x_T)$, there must be at least one more time period in which arm $2$ is played.
In that period, the UCB index of arm $2$ must be greater than that of arm $1$ (and that of all other arms).

For the term in (\ref{term6}), we have
\begin{align}
(\ref{term6}) & \le \sum_{m=1}^\infty \P_{\nu \pi}\Bigl( \widetilde{U}_1(m,x_T) \le \mu_2 + \delta \Bigr) \label{continue0} \\
& = \sum_{m=1}^\infty \P_{\nu \pi}\left( d_P(\widehat{\mu}_1(\tau_1(m)), \mu_2 + \delta) \ge \frac{f(x_T)}{m}, \;\;\; \widehat{\mu}_1(\tau_1(m)) < \mu_2 + \delta \right) \nonumber \\
& = \sum_{m=1}^\infty \P_{\nu \pi}\left( \frac{1}{m} \sum_{l=1}^m X_1(l) \le y_m^* \right) \nonumber \\
& \le \sum_{m=1}^\infty \exp\left( - m \cdot d_P(y_m^*,\mu_1) \right), \label{term3}
\end{align}
where for each $m$, $y_m^*$ is the unique solution to $d_P(y_m^*,\mu_2 + \delta) = f(x_T)/m$ and $y_m^* < \mu_2+\delta$, and we have used a Chernoff bound in (\ref{term3}).
We define
\begin{align}
s_T = \frac{2 f(x_T)}{d_P(\mu_2+\delta,\mu_1)} \cdot \inf_{z < \mu_2 + \delta} \frac{d_P(z,\mu_1)}{d_P(z,\mu_2 + \delta)}, \nonumber
\end{align}
and so for $m \ge s_T$,
\begin{align}
\frac{d_P(\mu_2+\delta,\mu_1)}{2} \ge \frac{f(x_T)}{m} \cdot \inf_{z < \mu_2 + \delta} \frac{d_P(z,\mu_1)}{d_P(z,\mu_2 + \delta)}. \nonumber
\end{align}
Since $y_m^* < \mu_2+\delta$, we have $d_P(y_m^*,\mu_1) \ge d_P(\mu_2+\delta,\mu_1)$, and so for $m \ge s_T$,
\begin{align}
d_P(y_m^*,\mu_1) \ge \frac{f(x_T)}{m} \cdot \inf_{z < \mu_2 + \delta} \frac{d_P(z,\mu_1)}{d_P(z,\mu_2 + \delta)} + \frac{d_P(\mu_2+\delta,\mu_1)}{2}. \label{specialupperbound}
\end{align} 
Splitting the sum in (\ref{term3}) into two pieces at $s_T$, we have
\begin{align}
(\ref{term3}) & = \sum_{m=1}^{\floor{s_T}} \exp\left( - m \cdot d_P(y_m^*,\mu_2 + \delta) \cdot \frac{d_P(y_m^*,\mu_1)}{d_P(y_m^*,\mu_2 + \delta)} \right) \nonumber \\
& \quad + \sum_{m=\floor{s_T}+1}^\infty \exp\left( - m \cdot d_P(y_m^*,\mu_1) \right) \nonumber \\
& \le \sum_{m=1}^{\floor{s_T}} \exp\left( -m \cdot \frac{f(x_T)}{m} \cdot \inf_{z < \mu_2 + \delta} \frac{d_P(z,\mu_1)}{d_P(z,\mu_2+\delta)} \right) \label{term4} \\
& \quad + \sum_{m = \floor{s_T}+1}^\infty \exp\left( -m \cdot \left( \frac{f(x_T)}{m} \cdot \inf_{z < \mu_2 + \delta} \frac{d_P(z,\mu_1)}{d_P(z,\mu_2+\delta)} + \frac{d_P(\mu_2+\delta,\mu_1)}{2} \right) \right) \label{term5} \\
& = \exp\left(- f(x_T) \cdot \underset{z < \mu_2 + \delta}{\inf} \frac{d_P(z,\mu_1)}{d_P(z,\mu_2+\delta)}\right) \left( \floor{s_T} + \sum_{m=\floor{s_T}+1}^\infty \exp\left( -m \cdot \frac{d_P(\mu_2+\delta,\mu_1)}{2} \right) \right). \label{term6bound}
\end{align}
In (\ref{term4}), we use the fact that $d_P(y_m^*,\mu_2 + \delta) = f(x_T)/m$.
In (\ref{term5}), we use (\ref{specialupperbound}) (for $m \ge s_T$).

For the term in (\ref{term7}), we have for sufficiently large $T$,
\begin{align}
\abs{\widetilde{U}_2(x_T,T) - \widehat{\mu}_2(\tau_2(x_T))} < \frac{\delta}{2}. \nonumber
\end{align}
So, for sufficiently large $T$,
\begin{align}
(\ref{term7}) & \le \P_{\nu \pi}\left( \frac{1}{x_T} \sum_{l=1}^{x_T} X_2(l) > \mu_2 + \frac{\delta}{2} \right) \nonumber \\
& \le \exp\Bigl( - x_T \cdot d_P(\mu_2 + \delta/2, \mu_2) \Bigr), \label{term7bound}
\end{align}
where (\ref{term7bound}) follows from a Chernoff bound.

Using (\ref{term6}) and (\ref{term6bound}) together with (\ref{term7}) and (\ref{term7bound}), we have
\begin{align}
\limsup_{T \to \infty} \frac{\log \P_{\nu \pi} \left( N_2(T) > x_T \right)}{f(x_T)} \le - \inf_{z < \mu_2 + \delta} \frac{d_P(z,\mu_1)}{d_P(z,\mu_2+\delta)}. \label{ub1}
\end{align}
From the argument included separately in Appendix \ref{sketchproofs},
\begin{align}
\lim_{\delta \downarrow 0} \inf_{z < \mu_2 + \delta} \frac{d_P(z,\mu_1)}{d_P(z,\mu_2+\delta)} = \inf_{z < \mu_2} \frac{d_P(z,\mu_1)}{d_P(z,\mu_2)}. \label{deltatozero}
\end{align}
Thus, we have established that for sub-optimal arm $i = 2$,
\begin{align*}
\limsup_{T \to \infty} \frac{\log \P_{\nu \pi}\left( N_2(T) > x_T \right)}{f(x_T)} \le - \inf_{z < \mu_2} \frac{d_P(z,\mu_1)}{d_P(z,\mu_2)}.
\end{align*}

We now consider any sub-optimal arm $i \ge 3$.
Let $\delta \in (0,\mu_{i-1} - \mu_i)$.
In parallel to (\ref{term6}) and (\ref{term7}), we have
\begin{align}
\P_{\nu \pi} \left( N_i(T) > x_T \right) & \le \P_{\nu \pi}\Bigl( \exists \; t \in (x_T,T] \;\;\; \text{s.t.} \;\;\; \max_{1 \le j \le i-1} \widetilde{U}_j(N_j(t-1),x_T) \le \mu_i + \delta \Bigr) \label{term6_otherarms} \\ 
& \quad + \P_{\nu \pi}\Bigl( \widetilde{U}_i(x_T,T) > \mu_i + \delta \Bigr). \label{term7_otherarms}
\end{align}
We can bound (\ref{term6_otherarms}) via:
\begin{align}
    (\ref{term6_otherarms}) & \le \P_{\nu \pi}\Bigl( \forall \; 1 \le j \le i-1, \;\;\; \exists \; m_j \in \mathbb{Z}_+ \;\;\; \text{s.t.} \;\;\; \widetilde{U}_j(m_j,x_T) \le \mu_i + \delta \Bigr) \nonumber \\
    & \le \prod_{j=1}^{i-1} \sum_{m=1}^\infty \P_{\nu \pi}\Bigl( \widetilde{U}_j(m,x_T) \le \mu_i + \delta \Bigr), \label{prod1}
\end{align}
where (\ref{prod1}) follows from the independence of the rewards from different arms.
We can then upper bound each term of the product in (\ref{prod1}) in the same way that we upper bounded (\ref{continue0}).
We can upper bound (\ref{term7_otherarms}) in the same way that we upper bounded (\ref{term7}), and thus show that (\ref{term7_otherarms}) is asymptotically negligible.
Following the rest of the argument above (which was for the case $i = 2$), we obtain for any sub-optimal arm $i \ge 3$:
\begin{align}
\limsup_{T \to \infty} \frac{\log \P_{\nu \pi}\left( N_i(T) > x_T \right)}{f(x_T)} \le - \sum_{j=1}^{i-1} \inf_{z < \mu_i} \frac{d_P(z,\mu_j)}{d_P(z,\mu_i)}, \nonumber
\end{align}
where the sum on the right side results from taking log of the product in (\ref{prod1}).
\underline{Lower Bound and Final Result} \\
Consider any sub-optimal arm $i \ge 2$.
In the context of Lemma \ref{lem2}, take $g(t) = \log^{1+\gamma}(t)$ with any $\gamma > 0$.
From proof of Theorem \ref{thm1}, we have the lower bound:
\begin{align}
\liminf_{T \to \infty} \frac{\log \P_{\nu \pi} \left( N_i(T) > (1-\gamma)T \right)}{\log(T)} \ge - \sum_{j=1}^{i-1} \inf_{z < \mu_i} \frac{d_P(z,\mu_j)}{d_P(z,\mu_i)}, \label{lb2}
\end{align}
which establishes part (i) of Lemma \ref{lem2}.

In the above proof of the upper bound part, choose increasing $f : (1,\infty) \to (0,\infty)$ to satisfy $\lim_{t \to \infty} f(t)/\log(t) = 1$, with $x_T = \lfloor g(T) \rfloor$.
Then, the above proof of the upper bound part yields for any sub-optimal arm $i \ge 2$:
\begin{align}
\limsup_{T \to \infty} \frac{\log \P_{\nu \pi}\left( N_i(T) > g(T) \right)}{\log(g(T))} \le - \sum_{j=1}^{i-1} \inf_{z < \mu_i} \frac{d_P(z,\mu_j)}{d_P(z,\mu_i)}, \label{ub2}
\end{align}
which establishes part (ii) of Lemma \ref{lem2}.

The desired uniform convergence result in (\ref{klucbtailexponent}) then follows from part (iii) of Lemma \ref{lem2}.

\halmos
\endproof

\section{Proof of Theorem \ref{gartnerellis}} \label{gartnerellis_proof}

\proof{Proof of Theorem \ref{gartnerellis}.}
 Without loss of generality, suppose that the long-run average rewards (in the sense of (\ref{slln1})) for arms $1,2,\dots,K$ within the environment $\nu$ satisfy $\overbar{\Lambda}_1'(0) > \overbar{\Lambda}_2'(0) > \dots > \overbar{\Lambda}_K'(0)$ (i.e., $r(i) = i$ for all $i \in [K]$).
Consider any sub-optimal arm $i \ge 2$.
Let $\widetilde{\nu}$ be an alternative environment where the reward distribution structure remains the same for arms $i,i+1,\dots,K$.
However, for arms $j \le i-1$ in the environment $\widetilde{\nu}$, let the distribution of $\sum_{t=1}^n X_j(t)$ for each $n \ge 1$ be
\begin{align}
    Q_j^n(dx; \theta_j) = \exp\left(\theta_j \cdot x - n \overbar{\Lambda}_j^n(\theta_j) \right) Q_j^n(dx), \nonumber
\end{align}
where $Q_j^n(dx)$ is the original distribution for $\sum_{t=1}^n X_j(t)$ in the environment $\nu$.
Moreover, let $\theta_j \in \overbar{\Theta}_j$ such that $\overbar{\Lambda}_j'(\theta_j) < \overbar{\Lambda}_i'(0)$ (note that $\theta_j < 0$).
(If this is not possible, then the infimum on the right side of (\ref{gartnerellisratio}) is empty, and the lower bound is $-\infty$.) 
So, in the environment $\widetilde{\nu}$, arm $i$ yields the greatest long-run average rewards compared to all other arms.

Let $\delta > 0$, and define the events:
\begin{align*}
& \mathcal{A}_T = \left\{ \abs{ \frac{N_j(T)}{\log(T)} - \frac{1}{d_P(\overbar{\Lambda}_j'(\theta_j),\overbar{\Lambda}_i'(0))} } \le \delta, \; \forall \; j \le i-1 \right\} \\
& \qquad \;\; \cap \left\{ \abs{ \frac{N_j(T)}{\log(T)} - \frac{1}{d_P(\overbar{\Lambda}_j'(0),\overbar{\Lambda}_i'(0))} } \le \delta, \; \forall \; j \ge i+1 \right\} \\
& \mathcal{B}_T = \left\{\abs{\widehat{\mu}_j(T) - \overbar{\Lambda}_j'(\theta_j))} \le \delta, \; \forall \; j \le i-1 \right\}.
\end{align*}
Following steps analogous to (\ref{begin})-(\ref{lln2}) in the proof of Theorem \ref{thm1},
\begin{align}
& \P_{\nu \pi}(N_i(T) > (1-\gamma)T) \nonumber \\
& = \E_{\widetilde{\nu} \pi}\left[ \indic{N_i(T) > (1-\gamma)T} \exp\left( \sum_{j=1}^{i-1} \left( - \theta_j \cdot \sum_{t=1}^{N_j(T)} X_j(t) + N_j(T) \cdot \overbar{\Lambda}_j^{N_j(T)}(\theta_j) \right) \right)\right] \label{lastcom1} \\
& \ge \E_{\widetilde{\nu} \pi}\left[ \indic{\mathcal{A}_T,\mathcal{B}_T} \exp\left( \sum_{j=1}^{i-1} \Bigl(- \theta_j \cdot \widehat{\mu}_j(T) + \overbar{\Lambda}_j^{N_j(T)}(\theta_j) \Bigr) N_j(T) \right) \right] \label{lastcom2} \\
& \ge \E_{\widetilde{\nu} \pi}\left[ \indic{\mathcal{A}_T,\mathcal{B}_T} \exp\left( \sum_{j=1}^{i-1} \Bigl(- \theta_j \cdot \left(\overbar{\Lambda}_j'(\theta_j) - \delta \right) + \overbar{\Lambda}_j(\theta_j) - \delta \Bigr) N_j(T) \right) \right] \label{gartnerellis1} \\
& = \E_{\widetilde{\nu} \pi}\left[ \indic{\mathcal{A}_T,\mathcal{B}_T} \exp\left( - \sum_{j=1}^{i-1} \Bigl( \overbar{\Lambda}_j^*(\overbar{\Lambda}_j'(\theta_j)) + \delta(1-\theta_j) \Bigr) N_j(T) \right) \right] \label{gartnerellis2} \\
& \ge \P_{\widetilde{\nu} \pi}(\mathcal{A}_T,\mathcal{B}_T) \cdot \exp\left( - \sum_{j=1}^{i-1} \Bigl( \overbar{\Lambda}_j^*(\overbar{\Lambda}_j'(\theta_j)) + \delta(1-\theta_j) \Bigr) \left(\frac{1}{d_P(\overbar{\Lambda}_j'(\theta_j),\overbar{\Lambda}_i'(0))} + \delta \right) \log(T) \right). \label{gartnerellis3}
\end{align}
In (\ref{lastcom1}), we have performed a change-of-measure from environment $\nu$ to $\widetilde{\nu}$.
In (\ref{lastcom2}), we use the fact that $\{N_i(T) > (1-\gamma)T\} \supset \mathcal{A}_T$ for sufficiently large $T$.
We have used the event $\mathcal{B}_T$ in (\ref{gartnerellis1}), and the relevant identity for the convex conjugates $\overbar{\Lambda}_j^*$ in (\ref{gartnerellis2}).
We have used the event $\mathcal{A}_T$ in (\ref{gartnerellis3}).
We also note that $\lim_{T \to \infty} \P_{\widetilde{\nu} \pi}(\mathcal{A}_T,\mathcal{B}_T) = 1$.
In environment $\widetilde{\nu}$, $\lim_{T \to \infty} \P_{\widetilde{\nu} \pi}(\mathcal{A}_T) = 1$ is due to the $\mathcal{M}_P$-pathwise convergence property of the algorithm $\pi$, as in (\ref{samplepathconsistent}).
And $\lim_{T \to \infty} \P_{\widetilde{\nu} \pi}(\mathcal{B}_T) = 1$ is due to the same result for $\mathcal{A}_T$, together with the sample mean WLLN that comes from Assumptions \ref{A3}-\ref{A4} (using the upper bound part of the G\"{a}rtner-Ellis Theorem; for details, see Lemma 3.2.5 of \cite{bucklew_2004}). 
From (\ref{gartnerellis3}), taking logs and dividing by $\log(T)$, and sending $T \to \infty$ followed by $\delta \downarrow 0$, we obtain:
\begin{align}
    \liminf_{T \to \infty} \frac{\log \P_{\nu \pi}(N_i(T) > (1-\gamma)T)}{\log(T)} \ge - \sum_{j=1}^{i-1} \frac{\overbar{\Lambda}_j^*(\overbar{\Lambda}_j'(\theta_j))}{d_P(\overbar{\Lambda}_j'(\theta_j),\overbar{\Lambda}_i'(0))}. \nonumber
\end{align}
This holds for any $\theta_j \in \overbar{\Theta}_j$, $j \le i-1$ such that $\overbar{\Lambda}_j'(\theta_j) < \overbar{\Lambda}_i'(0)$.
Under Assumptions \ref{A3}-\ref{A4}, each $\overbar{\Lambda}_j'$ is an invertible mapping between $\overbar{\Theta}_j$ and $\overline{\mathcal{I}}_j$ (see Theorem 26.5 of \cite{rockafellar_1970}).
Thus,
\begin{align}
    \liminf_{T \to \infty} \frac{\log \P_{\nu \pi}(N_i(T) > (1-\gamma)T)}{\log(T)} \ge - \sum_{j=1}^{i-1} \inf_{z \in \overline{\mathcal{I}}_j \, : \, z < \overbar{\Lambda}_i'(0)} \frac{\overbar{\Lambda}_j^*(z)}{d_P(z,\overbar{\Lambda}_i'(0))}. \nonumber
\end{align}
The conclusion in (\ref{gartnerellisratio}), with the infimum over $B_\gamma(T) = [\log^{1+\gamma}(T),(1-\gamma)T]$, follows from part (i) of Lemma \ref{lem2} with $g(t) = \log^{1+\gamma}(t)$.
\halmos
\endproof

\section{Proof of Theorem \ref{thm0}} \label{tradeoff_proof}

In the proof below, for any distributions $Q$ and $Q'$, we use $dQ/dQ'$ to denote the Radon-Nikodym derivative of the absolutely continuous part of $Q$ with respect to $Q'$, in accordance with the Lebesgue decomposition of $Q$ with respect to $Q'$ (see Theorem 6.10 of \cite{rudin_1987} for a precise statement), and we write $Q \ll Q'$ if $Q$ is absolutely continuous with respect to $Q'$.

\proof{Proof of Theorem \ref{thm0}.}
Suppose there is an environment $\widetilde{\nu} = (\widetilde{P}_1,P_2,\dots,P_K) \in \mathcal{M}^K$ for which (\ref{generalinproblowerbound}) is false.
Without loss of generality, let arm $2$ be optimal (i.e., $\mu(P_2) = \mu_*(\widetilde{\nu})$), and suppose for sub-optimal arm 1 there exists $\epsilon \in (0,1)$ and a sequence of deterministic times $T_n \uparrow \infty$ such that for all $n$,
\begin{align}
\P_{\widetilde{\nu} \pi}\left( \frac{N_1(T_n)}{f(T_n)} \le \frac{1 - \epsilon}{D_{\text{inf}}(\widetilde{P}_1,\mu(P_2),\mathcal{M})} \right) \ge \epsilon. \label{contradiction}
\end{align}
Denote the event in (\ref{contradiction}) by $\mathcal{A}_n$.
Consider any $P_1 \in \mathcal{M}$ such that $\widetilde{P}_1 \ll P_1$, $\mu(P_1) > \mu(P_2)$, and
\begin{align}
\frac{D(\widetilde{P}_1 \: \lVert \: P_1)}{D_{\text{inf}}(\widetilde{P}_1,\mu(P_2),\mathcal{M})} \le 1 + \epsilon. \label{theta1choice}
\end{align}
(Such $P_1$ exists or else $D_{\text{inf}}(\widetilde{P}_1,\mu(P_2),\mathcal{M}) = \infty$ and (\ref{generalinproblowerbound}) would hold trivially for $\widetilde{\nu}$.)
Let $\nu = (P_1,P_2,\dots,P_K) \in \mathcal{M}^K$ so that arm 1 is now optimal, with $P_2,\dots,P_K$ the same as in $\widetilde{\nu}$.
Let $\delta > 0$, and define the events:
\begin{align*}
\mathcal{B}_n & = \biggl\{ \biggl| \frac{1}{N_1(T_n)} \sum_{t=1}^{N_1(T_n)} \log \frac{dP_1}{d\widetilde{P}_1}(X_1(t)) + D(\widetilde{P}_1 \: \lVert \: P_1) \biggr| \le \delta \biggr\} \\
\mathcal{C}_n & = \{\exists \; i \neq 1 : N_i(T_n) > T_n/K\}.
\end{align*}
By a change of measure from $\nu$ to $\widetilde{\nu}$ (with an inequality due to the possibility that $P_1 \not\ll \widetilde{P}_1$),
\begin{align}
\P_{\nu \pi}(\mathcal{C}_n) & \ge \E_{\widetilde{\nu} \pi}\left[ \indic{\mathcal{C}_n} \prod_{t=1}^{N_1(T_n)} \frac{dP_1}{d\widetilde{P}_1}(X_1(t)) \right] \label{changeofmeasure} \\
& \ge \E_{\widetilde{\nu} \pi}\left[ \indic{\mathcal{A}_n,\mathcal{B}_n} \exp\left( \frac{1}{N_1(T_n)} \sum_{t=1}^{N_1(T_n)} \log \frac{dP_1}{d\widetilde{P}_1}(X_1(t)) \cdot N_1(T_n) \right) \right] \label{changeofmeasure0} \\
& \ge \P_{\widetilde{\nu} \pi}(\mathcal{A}_n,\mathcal{B}_n) \cdot \exp\left( -\left(D(\widetilde{P}_1 \: \lVert \: P_1) + \delta \right) \cdot \frac{1-\epsilon}{D_{\text{inf}}(\widetilde{P}_1,\mu(P_2),\mathcal{M})} f(T_n) \right), \label{changeofmeasure00}
\end{align}
where (\ref{changeofmeasure0}) follows from $\mathcal{C}_n \supset \mathcal{A}_n$ for large $n$ since $f(t) = o(t)$, and (\ref{changeofmeasure00}) follows from lower bounds using $\mathcal{A}_n$ and $\mathcal{B}_n$.
By Lemma \ref{lem0} (see below) and the WLLN for sample means, $\lim_{n \to \infty} \P_{\widetilde{\nu} \pi}(\mathcal{B}_n) = 1$.
So from (\ref{contradiction}), $\liminf_{n \to \infty} \P_{\widetilde{\nu} \pi}(\mathcal{A}_n,\mathcal{B}_n) \ge \epsilon$.
From (\ref{changeofmeasure00}), taking logs and dividing by $f(T_n)$, sending $n \to \infty$ followed by $\delta \downarrow 0$, and then applying (\ref{theta1choice}), we obtain:
\begin{align}
\liminf_{n \to \infty} \frac{\log \P_{\nu \pi}(\mathcal{C}_n)}{f(T_n)} \ge \liminf_{n \to \infty} \frac{\log \P_{\widetilde{\nu} \pi}(\mathcal{A}_n,\mathcal{B}_n)}{f(T_n)} -(1-\epsilon) \frac{D(\widetilde{P}_1 \: \lVert \: P_1)}{D_{\text{inf}}(\widetilde{P}_1,\mu(P_2),\mathcal{M})} \ge -(1-\epsilon^2). \label{heavytail}
\end{align}
Noting that $(K-1) \cdot \max_{i \ne 1} \P_{\nu \pi}(N_i(T_n) > T_n/K) \ge \P_{\nu \pi}(\mathcal{C}_n)$, we obtain:
\begin{align*}
\liminf_{n \to \infty} \frac{\max_{i \ne 1} \log \P_{\nu \pi}(N_i(T_n) > T_n/K)}{f(T_n)} \ge -(1-\epsilon^2).
\end{align*}
Since $\epsilon \in (0,1)$, this violates (\ref{tailconsistency}) for some sub-optimal arm $i \ne 1$ (under environment $\nu$), and thus (\ref{contradiction}) cannot be true.
\halmos
\endproof

The proof of Lemma \ref{lem0} is a simplification of the proof of Theorem \ref{thm0}.

\begin{lemma} \label{lem0}
Under the assumptions of Theorem \ref{thm0}, for any environment $\nu = (P_1,\dots,P_K) \in \mathcal{M}^K$ and each sub-optimal arm $i$, we have $N_i(T) \to \infty$ in $\P_{\nu \pi}$-probability as $T \to \infty$.
\end{lemma}

\proof{Proof of Lemma \ref{lem0}.}
Suppose the conclusion is false for some environment $\widetilde{\nu} = (\widetilde{P}_1,P_2,\dots,P_K) \in \mathcal{M}^K$.
Without loss of generality, suppose arm 1 is sub-optimal in $\widetilde{\nu}$ and there exists $m > 0$, $\epsilon > 0$ and a deterministic sequence of times $T_n \uparrow \infty$ such that for all $n$,
\begin{align}
\P_{\widetilde{\nu} \pi}\left( N_1(T_n) \le m \right) \ge \epsilon. \label{contradiction0}
\end{align}
Denote the event in (\ref{contradiction0}) by $\mathcal{A}_n'$. 
Consider another environment $\nu = (P_1,P_2,\dots,P_K) \in \mathcal{M}^K$ where arm 1 is optimal (with all other arms being the same as in $\widetilde{\nu}$).
Pick $L > 0$ large enough so that
\begin{align}
\P_{\widetilde{\nu} \pi}\left( \forall \; l = 1,\dots,m : \frac{1}{l} \sum_{t=1}^l \log \frac{dP_1}{d\widetilde{P}_1}(X_1(t)) \ge - L \right) \ge 1-\epsilon/2. \label{bigevent}
\end{align}
Define
\begin{align*}
\mathcal{B}_n' & = \biggl\{ \frac{1}{N_1(T_n)} \sum_{t=1}^{N_1(T_n)} \log \frac{dP_1}{d\widetilde{P}_1}(X_1(t)) \ge - L \biggr\}.
\end{align*}
Following the same steps from (\ref{changeofmeasure})-(\ref{changeofmeasure00}) but with $\mathcal{A}_n'$, $\mathcal{B}_n'$ in the place of $\mathcal{A}_n$, $\mathcal{B}_n$, respectively,
\begin{align*}
\P_{\nu \pi}(\exists \; i \neq 1 : N_i(T_n) > T_n/K) 
\ge \P_{\widetilde{\nu} \pi}(\mathcal{A}_n',\mathcal{B}_n') \cdot \exp\left( - L m \right).
\end{align*}
By (\ref{contradiction0})-(\ref{bigevent}), we have $\P_{\widetilde{\nu} \pi}(\mathcal{A}_n',\mathcal{B}_n') \ge \epsilon/2$ for all $n$.
Like previously with (\ref{heavytail}), this violates (\ref{tailconsistency}) for some sub-optimal arm $i \ne 1$ (under environment $\nu$), and so (\ref{contradiction0}) cannot be true.
\halmos
\endproof

\bibliographystyle{informs2014} 
\bibliography{references} 



\ECSwitch


\ECHead{Online Supplement for ``The Fragility of Optimized Bandit Algorithms''}


\section{Proofs for Section \ref{generalchar}} \label{generalcharproofs}

For the proofs in Appendix \ref{generalcharproofs}, we will work with the natural parameterization of the exponential family in (\ref{model}):
\begin{align}
    P_\theta(dx) = \exp\left(\theta \cdot x - \Lambda_P(\theta)\right) P(dx), \quad \theta \in \Theta_P. \label{alt_model}
\end{align}
Then, the KL divergence between distributions $P_{\theta}$ and $P_{\theta_0}$ has the expression:
\begin{align}
    D(P_{\theta} \: \lVert \: P_{\theta_0}) = \Lambda_P(\theta_0) - \Lambda_P(\theta) - \Lambda_P'(\theta) \cdot (\theta_0 - \theta). \label{alt_bregman}
\end{align}

\proof{Proof of Lemma \ref{lem1}.}
First of all, the definition of discrimination equivalence, as expressed in (\ref{klratio}) for the exponential family with base distribution $P$ parameterized by mean (as in (\ref{model})), is equivalent to the following statement for the same exponential family with natural parameterization (as in (\ref{alt_model})).
For any $\theta_1,\theta_2 \in \Theta_P$ with $\theta_1 > \theta_2$,
\begin{align}
    \inf_{\theta \in \Theta_P \, : \, \theta < \theta_2} \frac{D(P_{\theta} \: \lVert \: P_{\theta_1})}{D(P_{\theta} \: \lVert \: P_{\theta_2})} = 1. \label{alt_klratio}
\end{align}
Also, we recall that on $\Theta_P$, $\Lambda_P$ is strictly convex and $\Lambda_P'$ is strictly increasing.

We first show the forward direction, that (\ref{alt_klratio}) implies (\ref{equivalentcondition}).
Suppose $\inf \Theta_P > -\infty$.
Note that (\ref{alt_klratio}) implies that for any fixed $\theta_0 > \inf \Theta_P$,
\begin{align}
\lim_{\theta \downarrow \inf \Theta_P} D(P_{\theta} \: \lVert \: P_{\theta_0}) = \lim_{\theta \downarrow \inf \Theta_P} \Lambda_P(\theta_0) - \Lambda_P(\theta) - \Lambda_P'(\theta) \cdot (\theta_0 - \theta) = \infty. \label{alt_bregman_limit}
\end{align}
Because $\inf \Theta_P > -\infty$ and $\Lambda_P$ is strictly convex, we must have 
\begin{align}
    \lim_{\theta \downarrow \inf \Theta_P} \Lambda_P(\theta) > -\infty, \label{impliciation0}
\end{align} 
and so (\ref{alt_bregman_limit}) implies that 
\begin{align}
\lim_{\theta \downarrow \inf \Theta_P} \Lambda_P'(\theta) = -\infty. \label{implication2}
\end{align}
Then, taking $\theta_0$ arbitrarily close to $\inf \Theta_P$ in (\ref{alt_bregman_limit}), we have for any $\epsilon > 0$:
\begin{align}
\lim_{\theta \downarrow \inf \Theta_P} \Lambda_P(\theta) + \epsilon \Lambda_P'(\theta) = -\infty. \label{implication1}
\end{align}
So, (\ref{impliciation0}), (\ref{implication2}) and (\ref{implication1}) imply that
\begin{align*}
\lim_{\theta \downarrow \inf \Theta_P} \frac{\Lambda_P(\theta)}{\Lambda_P'(\theta)} = 0. 
\end{align*}
So, for any $\theta_1,\theta_2$ fixed with $\theta_1 > \theta_2 > \inf \Theta_P$, we have
\begin{align*}
\lim_{\theta \downarrow \inf \Theta_P} \frac{D(P_{\theta} \: \lVert \: P_{\theta_1})}{D(P_{\theta} \: \lVert \: P_{\theta_2})} = \lim_{\theta \downarrow \inf \Theta_P} \frac{\Lambda_P'(\theta) \cdot (\theta_1 - \theta)}{\Lambda_P'(\theta) \cdot (\theta_2 - \theta)} = \frac{\theta_1 - \inf \Theta_P}{\theta_2 - \inf \Theta_P} > 1, 
\end{align*}
which contradicts (\ref{alt_klratio}) if $\inf \Theta_P > -\infty$.
Hence, it must be that $\inf \Theta_P = -\infty$.

Given $\inf \Theta_P = -\infty$, now suppose that 
\begin{align}
\liminf_{\theta \to -\infty} \bigl(\theta \Lambda_P'(\theta) - \Lambda_P(\theta)\bigr) < \infty. \label{finiteA2}
\end{align}
Consider the two possible cases:
\begin{enumerate}
\centering
    \item $\lim_{\theta \to -\infty} \Lambda_P'(\theta) = -\infty$
    \item $\lim_{\theta \to -\infty} \abs{\Lambda_P'(\theta)} < \infty$.
\end{enumerate}
(Since on $\Theta_P$, $\Lambda_P$ is strictly convex and $\Lambda_P'$ is strictly increasing, we cannot have $\lim_{\theta \to -\infty} \Lambda_P'(\theta) = \infty$.)
In the first case, (\ref{alt_klratio}) cannot hold because (\ref{alt_bregman}) and (\ref{finiteA2}) prevent (\ref{alt_bregman_limit}) for $\theta_0 < 0$.
In the second case, (\ref{alt_klratio}) cannot hold because (\ref{alt_bregman}) and (\ref{finiteA2}) imply that $\liminf_{\theta \to -\infty} D(P_{\theta} \: \lVert \: P_{\theta_0}) < \infty$ for any $\theta_0 \in \Theta_P$, thus preventing (\ref{alt_bregman_limit}).
So, it must be that
\begin{align*}
\lim_{\theta \to -\infty} \bigl( \theta \Lambda_P'(\theta) - \Lambda_P(\theta) \bigr) = \infty.
\end{align*}
Thus, the forward direction is established.

We now show the reverse direction, that (\ref{equivalentcondition}) implies (\ref{alt_klratio}).
Starting with (\ref{equivalentcondition}), there are again two possible cases:
\begin{enumerate}
\centering
    \item $\lim_{\theta \to -\infty} \Lambda_P'(\theta) = -\infty$
    \item $\lim_{\theta \to -\infty} \abs{\Lambda_P'(\theta)} < \infty$.
\end{enumerate}
In the first case, using the fact that $\lim_{\theta \to -\infty} D(P_{\theta} \: \lVert \: P_{\theta_0}) > 0$ for arbitrarily negative values of $\theta_0$, together with the identity in (\ref{alt_bregman}), we conclude that
\begin{align}
    \lim_{\theta \to -\infty} \frac{\Lambda_P'(\theta)}{\theta \Lambda_P'(\theta) - \Lambda_P(\theta)} = 0. \label{vanishingratio}
\end{align}
Then, (\ref{alt_bregman}), (\ref{equivalentcondition}) and (\ref{vanishingratio}) imply that
\begin{align}
    \lim_{\theta \to -\infty} \frac{D(P_{\theta} \: \lVert \: P_{\theta_1})}{D(P_{\theta} \: \lVert \: P_{\theta_2})} = \lim_{\theta \to -\infty} \frac{\theta \Lambda_P'(\theta) - \Lambda_P(\theta)}{\theta \Lambda_P'(\theta) - \Lambda_P(\theta)} = 1. \label{directconclusion}
\end{align}
In the second case, (\ref{equivalentcondition}) directly implies (\ref{vanishingratio}), which again leads to (\ref{directconclusion}).
Thus, the reverse direction is established.
\halmos
\endproof

\vspace{5mm}

\proof{Proof of Proposition \ref{prop3}.}
Since $\inf \Theta_P = -\infty$ and the support of the distributions is unbounded to the left (i.e., there is always positive probability mass to the left of any point on the real line), as we send $\theta$ to $-\infty$, the mean $\mu(P_{\theta}) = \Lambda_P'(\theta)$ must also go to $-\infty$. 
By the definition of the convex conjugate $\Lambda_P^*$, we have for any $\theta \in \Theta_P$,
\begin{align*}
\Lambda_P^*(z) \ge \theta \cdot z - \Lambda_P(\theta),
\end{align*}
which implies for $\theta < 0$ that
\begin{align*}
\lim_{z \to -\infty} \Lambda_P^*(z) = \infty.
\end{align*}
Also, note that for any $\theta \in \Theta_P$,
\begin{align*}
\Lambda_P^*(\Lambda_P'(\theta)) = \theta \cdot \Lambda_P'(\theta) - \Lambda_P(\theta).
\end{align*}
So, the desired result follows from the fact that $\lim_{\theta \to -\infty} \Lambda_P'(\theta) = -\infty$, together with Lemma \ref{lem1}.
\halmos
\endproof

\vspace{5mm}

\proof{Proof of Proposition \ref{prop4}.}
Let $X$ be a random variable with distribution $P$.
We first address the case in which $P$ assigns zero probability mass to the (finite) infimum of its support, which we denote by $L$.
For $l > L$, we have by the definition of convex conjugation:
\begin{align}
\Lambda_P^*(l) & = \sup_{\theta \in \Theta_P} \bigl(\theta \cdot l - \log \E[\exp(\theta X)]\bigr) \nonumber \\
& = - \log\left( \inf_{\theta \in \Theta_P} \E[\exp(\theta(X - l))] \right). \label{conjugate}
\end{align}
For any $\theta \in \Theta_P$ and $l > L$,
\begin{align}
0 \le \E[\exp(\theta(X - l))] \le \exp(\abs{\theta(l - L)}) \cdot \E[\exp(\theta(X - L))]. \nonumber
\end{align}
Therefore,
\begin{align}
0 \le \inf_{\theta \in \Theta_P} \E[\exp(\theta(X - l))] & \le \exp(\abs{-(l-L)^{-1}(l - L)}) \cdot \E[\exp(-(l-L)^{-1}(X - L))] \nonumber \\
& = \exp(1) \cdot \E[\exp(-(l-L)^{-1}(X - L))], \nonumber
\end{align}
and since $X > L$ with probability one, we have by the Bounded Convergence Theorem, $\lim_{l \downarrow L} \E[\exp(-(l-L)^{-1}(X - L))] = 0$.
So,
\begin{align*}
\lim_{l \downarrow L} \inf_{\theta \in \Theta_P} \E[\exp(\theta(X - l))] = 0,
\end{align*}
which, by (\ref{conjugate}), translates into
\begin{align*}
\lim_{l \downarrow L} \Lambda_P^*(l) = \infty.
\end{align*}
Since
\begin{align*}
\lim_{\theta \to -\infty} \Lambda_P'(\theta) = L,
\end{align*}
we have
\begin{align}
\lim_{\theta \to -\infty} \Lambda_P^*(\Lambda_P'(\theta)) = \infty, \label{continuousunboundedness}
\end{align}
which is the equivalent representation for (\ref{equivalentcondition}).

For the case in which $P$ places strictly positive mass on $L$, fix some $\eta \in (0,1)$ such that $\P(X \ge L + \eta) \ge \eta$.
Then, taking $m = (2/\eta)\log(1/\eta)$ and $l \in (L,L + \eta/2)$, we have
\begin{align}
\inf_{\theta \ge m} \E[\exp(\theta(X - l))] & = \inf_{\theta \ge m} \Bigl\{ \E[\exp(\theta(X - l)) ; X \ge l] + \E[\exp(\theta(X - l)) ; X < l] \Bigr\} \nonumber \\
& \ge \exp(m (L + \eta - l)) \P(X \ge L + \eta) \nonumber \\
& \ge \exp\left(m \frac{\eta}{2}\right) \eta \nonumber \\
& = 1 = \E[\exp(0 \cdot (X - l))]. \nonumber
\end{align}
So, it suffices to take the infimum over $\theta < m$:
\begin{align*}
\inf_{\theta \in \Theta_P} \E[\exp(\theta(X - l))] & = \inf_{\theta < m} \E[\exp(\theta(X - l))] \\
& \ge \inf_{\theta < m} \E[\exp(\theta(X - l)) ; X = L] \\
& = \inf_{\theta < m} \exp(\theta(L - l)) \cdot \P( X = L ) \\
& = \exp(m(L - l)) \cdot \P( X = L ).
\end{align*}
Therefore,
\begin{align*}
\liminf_{l \downarrow L} \inf_{\theta \in \Theta_P} \E[\exp(\theta(X - l))] \ge \P( X = L ),
\end{align*}
which, by (\ref{conjugate}), translates into
\begin{align*}
\limsup_{l \downarrow L} \Lambda_P^*(l) \le - \log \P( X = L ) < \infty,
\end{align*}
since $\P( X = L ) \in (0,1)$ by assumption.
So, although
\begin{align*}
\lim_{\theta \to -\infty} \Lambda_P'(\theta) = L,
\end{align*}
unlike in the case of continuous distributions, where we ended up with (\ref{continuousunboundedness}), here we have
\begin{align*}
\limsup_{\theta \to -\infty} \Lambda_P^*(\Lambda_P'(\theta)) < \infty. 
\end{align*} 
\halmos
\endproof

\textit{Derivations for Examples \ref{ex0}-\ref{ex3}.}

\noindent \underline{Example \ref{ex0}} \\
From (\ref{klequivalence}), we have for $z_2 < z_1 \in \mathbb{R}$,
\begin{align*}
\inf_{z \in \mathcal{I}_P \, : \, z < z_2} \frac{d_P(z,z_1)}{d_P(z,z_2)} = \lim_{z \downarrow -\infty} \frac{(z-z_1)^2}{(z-z_2)^2} = 1.
\end{align*}
So, $P$ is discrimination equivalent.

\noindent \underline{Example \ref{ex1}} \\
The CGF is $\theta \mapsto \log((e^\theta - 1)/\theta)$.
Let $\theta_1 = \theta_P(z_1)$, $\theta_2 = \theta_P(z_2)$ be the tilting parameters corresponding to the means $z_1$, $z_2$, with $0 < z_2 < z_1 < 1$.
Note that the mean $z \downarrow 0$ corresponds to $\theta_P(z) \downarrow -\infty$.
From (\ref{klequivalence}) and using (\ref{bregman}), we have
\begin{align*}
\inf_{z \in \mathcal{I}_P \, : \, z < z_2} \frac{d_P(z,z_1)}{d_P(z,z_2)} & = \lim_{\theta \to -\infty}  \frac{\log\left( \frac{e^{\theta_1} - 1}{\theta_1} \right) - \log\left( \frac{e^\theta - 1}{\theta} \right) - \frac{e^\theta (\theta - 1) + 1}{(e^\theta - 1)\theta} ( \theta_1 - \theta )}{\log\left( \frac{e^{\theta_2} - 1}{\theta_2} \right) - \log\left( \frac{e^\theta - 1}{\theta} \right) - \frac{e^\theta (\theta - 1) + 1}{(e^\theta - 1)\theta} ( \theta_2 - \theta )} \\
& = 1.
\end{align*}
So, $P$ is discrimination equivalent.

\noindent \underline{Example \ref{ex2}} \\
From (\ref{klequivalence}), we have for $0 < z_2 < z_1 < 1$,
\begin{align*}
\inf_{z \in \mathcal{I}_P \, : \, z < z_2} \frac{d_P(z,z_1)}{d_P(z,z_2)} & = \lim_{z \downarrow 0} \frac{z \log\left( \frac{z}{z_1} \right) + (1-z) \log\left( \frac{1-z}{1-z_1} \right)}{z \log\left( \frac{z}{z_2} \right) + (1-z) \log\left( \frac{1-z}{1-z_2} \right)} \\
& = \frac{\log(1-z_1)}{\log(1-z_2)}.
\end{align*}
So, $P$ is not discrimination equivalent.

\noindent \underline{Example \ref{ex3}} \\
The density of $P^z$ is $x \mapsto (\theta_P(z) + 1) e^{(\theta_P(z) + 1)x} \indic{x \le 0}$, where $\theta_P(z) \in (-1,\infty)$ is the tilting parameter corresponding to mean $z \in \mathbb{R}$.
The KL-divergence has the form:
\begin{align*}
d_P(z,z') & = \int_{-\infty}^0 (\theta_P(z) + 1) e^{(\theta_P(z) + 1)x} \left( \log\left( \frac{\theta_P(z) + 1}{\theta_P(z') + 1} \right) + (\theta_P(z) - \theta_P(z'))x \right) dx \\
& = - \frac{\theta_P(z) - \theta_P(z')}{\theta_P(z) + 1} + \log\left( \frac{\theta_P(z) + 1}{\theta_P(z') + 1} \right).
\end{align*}
Let $\theta_1 = \theta_P(z_1)$, $\theta_2 = \theta_P(z_2)$ be the tilting parameters corresponding to the means $z_1$, $z_2$, with $z_2 < z_1 < 0$.
Note that the mean $z \downarrow -\infty$ corresponds to $\theta_P(z) \downarrow -1$.
From (\ref{klequivalence}), we have
\begin{align*}
\inf_{z \in \mathcal{I}_P \, : \, z < z_2} \frac{d_P(z,z_1)}{d_P(z,z_2)} & = \lim_{\theta \downarrow -1} \frac{ - \frac{\theta - \theta_1}{\theta + 1} + \log\left( \frac{\theta + 1}{\theta_1 + 1} \right) }{ - \frac{\theta - \theta_2}{\theta + 1} + \log\left( \frac{\theta + 1}{\theta_2 + 1} \right)} \\
& = \frac{\theta_1 - \theta + (\theta+1)\left( \log(\theta + 1) - \log(\theta_1+1) \right)}{\theta_2 - \theta + (\theta+1)\left( \log(\theta+1) - \log(\theta_2+1) \right)} \\
& = \frac{\theta_1 + 1}{\theta_2 + 1} \\
& = \frac{z_2}{z_1}.
\end{align*}
So, $P$ is not discrimination equivalent.

\section{Proofs for Sections \ref{sketch}-\ref{upperbound}} \label{sketchproofs}

\proof{Proof of Proposition \ref{prop6}.}

Let $i$ be any sub-optimal arm.
From the lower bounds in (\ref{optimizedklratio}) in the proof of Theorem \ref{thm1}, there exists $a > 0$ such that for all $x \in [\log^{1+\gamma}(T), (1-\gamma)T]$ and $T$ sufficiently large,
\begin{align*}
    T^{-a} & \le \P_{\nu \pi}(N_i(T) > (1-\gamma)T) \\
    & \le \P_{\nu \pi}(N_i(T) > x).
\end{align*}
Thus,
\begin{align*}
    0 & \le \P_{\nu \pi}\left( \abs{\widehat{\mu}_i(T) - \mu_i} > \epsilon \mid N_i(T) > x \right) \\
    & \le \frac{\P_{\nu \pi}\left( \abs{\widehat{\mu}_i(T) - \mu_i} > \epsilon, \; N_i(T) > \log^{1+\gamma}(T) \right)}{\P_{\nu \pi}\left(N_i(T) > (1-\gamma)T\right)} \\
    & \le \frac{1}{\P_{\nu \pi}\left(N_i(T) > (1-\gamma)T\right)} \sum_{k = \lceil \log^{1+\gamma}(T) \rceil}^\infty \P_{\nu \pi}\left( \abs{\widehat{\mu}_i(T) - \mu_i} > \epsilon, \; N_i(T) = k \right) \\
    & \le T^a \left( \frac{\exp\left( - \log^{1+\gamma}(T) \cdot d_P(\mu_i + \epsilon,\mu_i) \right)}{1 - \exp\left(-d_P(\mu_i + \epsilon,\mu_i) \right)} + \frac{\exp\left( - \log^{1+\gamma}(T) \cdot d_P(\mu_i - \epsilon,\mu_i) \right)}{1 - \exp\left(-d_P(\mu_i - \epsilon,\mu_i) \right)} \right),
\end{align*}
where the last inequality follows from a Chernoff bound.
So, $\P_{\nu \pi}\left( \abs{\widehat{\mu}_i(T) - \mu_i} > \epsilon \mid N_i(T) > x \right) \to 0$ uniformly for $x \in [\log^{1+\gamma}(T), (1-\gamma)T]$ as $T \to \infty$, which yields the desired result.
\halmos
\endproof

\proof{Verification of (\ref{deltatozero}) in Proof of Theorem \ref{generalupperbound}.}
With the natural parameterization of an exponential family $P_\theta$, $\theta \in \Theta_P$, as in (\ref{alt_model}), with KL divergence as in (\ref{alt_bregman}), we have:
\begin{align*}
    \frac{d}{d\theta} D(P_{\theta} \: \lVert \: P_{\theta_0}) = - \Lambda_P''(\theta)(\theta_0 - \theta).
\end{align*}
Denote $\theta_1 := \theta_P(\mu_1)$ and $\theta_2 := \theta_P(\mu_2)$ (with $\theta_P(\cdot)$ as defined in the parameterization by mean in (\ref{model})), so that $\theta_2 < \theta_1$.
Let $\epsilon > 0$ such that $\theta_2 + \epsilon < \theta_1$.
Then,
\begin{align}
    \frac{d}{d\theta} \frac{D(P_{\theta} \: \lVert \: P_{\theta_1})}{D(P_{\theta} \: \lVert \: P_{\theta_2 + \epsilon})} = \frac{\Lambda_P''(\theta)}{D(P_{\theta} \: \lVert \: P_{\theta_2+\epsilon})^2} \left( \underbrace{D(P_{\theta} \: \lVert \: P_{\theta_1})(\theta_2+\epsilon-\theta) - D(P_{\theta} \: \lVert \: P_{\theta_2+\epsilon})(\theta_1-\theta)}_{:=\xi(\theta)} \right). \nonumber
\end{align}
Note that $\xi(\theta_2+\epsilon) = 0$ and $\xi'(\theta) = D(P_{\theta} \: \lVert \: P_{\theta_2+\epsilon}) - D(P_{\theta} \: \lVert \: P_{\theta_1})$ for $\theta < \theta_2+\epsilon$.
So, $\xi'(\theta) < 0$, and thus $\xi(\theta) > 0$ for $\theta < \theta_2+\epsilon$.
From this, together with the fact that $\Lambda_P''(\theta) \ge 0$ for all $\theta$, we conclude that $\theta \mapsto D(P_{\theta} \: \lVert \: P_{\theta_1})/D(P_{\theta} \: \lVert \: P_{\theta_2+\epsilon})$ is monotone increasing for $\theta < \theta_2+\epsilon$.

Let $\delta > 0$ such that $\mu_2 + \delta < \mu_1$.
Since $z \mapsto \theta_P(z)$ is monotone increasing, $z \mapsto d_P(z,\mu_1)/d_P(z,\mu_2+\delta)$ must also be monotone increasing for $z < \mu_2+\delta$.
So for any $\delta > 0$,
\begin{align}
    \inf_{z < \mu_2+\delta} \frac{d_P(z,\mu_1)}{d_P(z,\mu_2+\delta)} = \inf_{z < \mu_2} \frac{d_P(z,\mu_1)}{d_P(z,\mu_2+\delta)}. \nonumber
\end{align}
Since for $z < \mu_2$, $\delta \mapsto d_P(z,\mu_1)/d_P(z,\mu_2+\delta)$ is monotone decreasing, it must also be that $\delta \mapsto \inf_{z < \mu_2} d_P(z,\mu_1)/d_P(z,\mu_2+\delta)$ is monotone decreasing.
Therefore,
\begin{align}
    \lim_{\delta \downarrow 0} \inf_{z < \mu_2} \frac{d_P(z,\mu_1)}{d_P(z,\mu_2+\delta)} = \sup_{\delta > 0} \inf_{z < \mu_2} \frac{d_P(z,\mu_1)}{d_P(z,\mu_2+\delta)}. \nonumber
\end{align}
Finally, since both $z \mapsto d_P(z,\mu_1)/d_P(z,\mu_2+\delta)$ and $\delta \mapsto d_P(z,\mu_1)/d_P(z,\mu_2+\delta)$ are monotone, and thus are both quasi-convex and quasi-concave, Sion's Minimax Theorem yields:
\begin{align}
    \sup_{\delta > 0} \inf_{z < \mu_2} \frac{d_P(z,\mu_1)}{d_P(z,\mu_2+\delta)} = \inf_{z < \mu_2} \sup_{\delta > 0} \frac{d_P(z,\mu_1)}{d_P(z,\mu_2+\delta)} = \inf_{z < \mu_2} \frac{d_P(z,\mu_1)}{d_P(z,\mu_2)}. \nonumber
\end{align}
\halmos
\endproof

\section{Proofs for Section \ref{mis-specification}} \label{mis-specificationproofs}

\proof{Proof of Proposition \ref{wrongdistribution}.}

The proof follows the approach we have taken to establish previous results.
We use Lemma \ref{lem2} with $g(t) = \log^{1+\gamma}(t)$.
In the context of part (i) of Lemma \ref{lem2}, the proof of the lower bound part for (\ref{wrongdistributionratio}) follows from Theorem \ref{gartnerellis} (which uses Lemma \ref{stronglaw}).
In the context of part (ii) of Lemma \ref{lem2}, the upper bound part for (\ref{wrongdistributionratio}) can be established using the same arguments in the proof of Theorem \ref{generalupperbound}; see Appendix \ref{generalupperbound_proof}.
The uniform convergence result then follows from part (iii) of Lemma \ref{lem2}.

Without loss of generality, suppose that $\mu(Q_1) > \mu(Q_2) > \dots > \mu(Q_K)$ (i.e., $r(i) = i$ for all $i \in [K]$).
Here, the only thing that needs to be checked is the analog of (\ref{deltatozero}):
\begin{align}
    \lim_{\delta \downarrow 0} \inf_{z < \mu(Q_{2}) + \delta} \frac{d_{Q_{1}}(z, \mu(Q_{1}))}{d_P(z,\mu(Q_{2}) + \delta)} = \inf_{z < \mu(Q_{2})} \frac{d_{Q_{1}}(z, \mu(Q_{1}))}{d_P(z,\mu(Q_{2}))}. \label{deltatozero_general}
\end{align}
(Below, we check (\ref{deltatozero_general}) for $Q_1$ and $Q_2$. The same arguments apply for the other combinations of $Q_i$, $i \ge 3$ and $Q_j$, $j \le i-1$.)
First, there exists a fixed $\eta > 0$ (depending on $Q_{1}$ and $Q_{2}$) such that for all $\delta > 0$ sufficiently small, we have both:
\begin{align}
    \inf_{z < \mu(Q_{2}) + \delta}  \frac{d_{Q_{1}}(z,\mu(Q_{1}))}{d_P(z,\mu(Q_{2}) + \delta)} & = \inf_{z < \mu(Q_{2}) - \eta} \frac{d_{Q_{1}}(z, \mu(Q_{1}))}{d_P(z,\mu(Q_{2}) + \delta)}, \label{actualinf1} \\
    \inf_{z < \mu(Q_{2})} \frac{d_{Q_{1}}(z, \mu(Q_{1}))}{d_P(z,\mu(Q_{2}))} & = \inf_{z < \mu(Q_{2}) - \eta} \frac{d_{Q_{1}}(z, \mu(Q_{1}))}{d_P(z,\mu(Q_{2}))}. \label{actualinf2}
\end{align}
Note that
\begin{align*}
    z \mapsto \frac{d_P(z,\mu(Q_{2}))}{d_P(z,\mu(Q_{2}) + \delta)}
\end{align*}
is monotone decreasing for $z < \mu(Q_{2})$, which we deduce from the verification of (\ref{deltatozero}) in proof of Theorem \ref{generalupperbound} in Appendix \ref{sketchproofs}.
Also, we have:
\begin{align}
    \lim_{\delta \downarrow 0} \sup_{z < \mu(Q_{2}) - \eta} \frac{d_P(z,\mu(Q_{2}))}{d_P(z,\mu(Q_{2}) + \delta)} & = \sup_{\delta > 0} \sup_{z < \mu(Q_{2}) - \eta} \frac{d_P(z,\mu(Q_{2}))}{d_P(z,\mu(Q_{2}) + \delta)} = 1, \label{intermediate1} \\
    \lim_{\delta \downarrow 0} \frac{d_P(\mu(Q_{2}) - \eta,\mu(Q_{2}))}{d_P(\mu(Q_{2}) - \eta,\mu(Q_{2}) + \delta)} & = 1. \label{intermediate2}
\end{align}
The monotonicity property, together with (\ref{intermediate1})-(\ref{intermediate2}), imply uniform convergence for $z < \mu(P_{2}) - \eta$:
\begin{align}
    \lim_{\delta \downarrow 0} \sup_{z < \mu(Q_{2}) - \eta} \abs{\frac{d_P(z,\mu(Q_{2}))}{d_P(z,\mu(Q_{2}) + \delta)} - 1} = 0. \label{actualinf3}
\end{align}
For any $\epsilon \in (0,1)$, using (\ref{actualinf3}), we have for sufficiently small $\delta > 0$:
\begin{align}
    (1-\epsilon) \inf_{z < \mu(Q_{2}) - \eta} \frac{d_{Q_{1}}(z, \mu(Q_{1}))}{d_P(z,\mu(Q_{2}))} & \le \inf_{z < \mu(Q_{2}) - \eta} \frac{d_{Q_{1}}(z, \mu(Q_{1}))}{d_P(z,\mu(Q_{2}) + \delta)} \cdot \frac{d_P(z,\mu(Q_{2}))}{d_P(z,\mu(Q_{2}))} \label{intermediate3} \\
    & \le (1+\epsilon) \inf_{z < \mu(Q_{2}) - \eta} \frac{d_{Q_{1}}(z, \mu(Q_{1}))}{d_P(z,\mu(Q_{2}))}. \label{intermediate4}
\end{align}
In (\ref{intermediate3})-(\ref{intermediate4}), sending $\delta \downarrow 0$, followed by $\epsilon \downarrow 0$, and then using (\ref{actualinf1})-(\ref{actualinf2}), we obtain (\ref{deltatozero_general}).
\halmos
\endproof

\vspace{5mm}

\proof{Proof of Corollary \ref{cor1}.}

Let $\nu$ consist of two Gaussian reward distributions with variance $\sigma_0^2$, and $\mu_1$ and $\mu_2$ as the means for arms 1 and 2, respectively.
Without loss of generality, suppose that $\mu_1 > \mu_2$ (i.e., $r(i) = i$ for $i=1,2$).
We again use Lemma \ref{lem2} with $g(t) = \log^{1+\gamma}(t)$.
In the context of part (i) of Lemma \ref{lem2}, the proof of the lower bound part for (\ref{wrongdistributionratio}) follows from Theorem \ref{gartnerellis} (which uses Lemma \ref{stronglaw}).
In the context of part (ii) of Lemma \ref{lem2}, the upper bound part:
\begin{align}
    \limsup_{T \to \infty} \frac{\log \P_{\nu \pi}(N_2(T) > \log^{1+\gamma}(T))}{\log(\log^{1+\gamma}(T))} \le - \frac{\sigma^2}{\sigma_0^2}, \label{TSUCBupperbound}
\end{align}
actually follows from the proof of the upper bound part of Theorem \ref{generalupperbound}.
In the Gaussian setting, the proof is substantially simpler, and so for future reference, we provide it below.
The uniform convergence result then follows from part (iii) of Lemma \ref{lem2}.
\halmos
\endproof

\vspace{5mm}

\proof{Verification of (\ref{TSUCBupperbound}) in Proof of Corollary \ref{cor1}.}
Let $x_T = \floor{\log^{1+\gamma}(T)}$ with fixed $\gamma \in (0,1)$.
Let $\Delta = \mu_1 - \mu_2 > 0$.
As in the proof of Theorem \ref{generalupperbound}, we have:
\begin{align}
& \P_{\nu \pi} \left( N_2(T) > x_T \right) \nonumber \\
& \le \P_{\nu \pi}\left( \exists \; t \in (\tau_2(x_T),T] \;\;\; \text{s.t.} \;\;\; \widehat{\mu}_1(t-1) + \sqrt{\frac{2\sigma^2 \log(t-1)}{N_1(t-1)}} \le \widehat{\mu}_2(t-1) + \sqrt{\frac{2\sigma^2 \log(t-1)}{N_2(t-1)}} \right) \nonumber \\ 
& \le \P_{\nu \pi}\left( \exists \; t \in (x_T,T] \;\;\; \text{s.t.} \;\;\; \widehat{\mu}_1(t-1) + \sqrt{\frac{2\sigma^2 \log(x_T)}{N_1(t-1)}} \le \widehat{\mu}_2(\tau_2(x_T)) + \sqrt{\frac{2\sigma^2 \log(T)}{x_T}} \right) \nonumber \\ 
& \le \P_{\nu \pi}\left( \exists \; t \in (x_T,T] \;\;\; \text{s.t.} \;\;\; \widehat{\mu}_1(t-1) + \sqrt{\frac{2\sigma^2 \log(x_T)}{N_1(t-1)}} \le \mu_2 + \frac{\Delta}{2} \right) \label{term1} \\ 
& \quad + \P_{\nu \pi}\left( \widehat{\mu}_2(\tau_2(x_T)) + \sqrt{\frac{2\sigma^2 \log(T)}{x_T}} > \mu_2 + \frac{\Delta}{2} \right). \label{term2}
\end{align}

For the term in (\ref{term1}), we have
\begin{align}
(\ref{term1}) & = \P_{\nu \pi}\left( \exists \; t \in (x_T,T] \;\;\; \text{s.t.} \;\;\; \widehat{\mu}_1(t-1) + \sqrt{\frac{2\sigma^2 \log(x_T)}{N_1(t-1)}} \le \mu_1 - \frac{\Delta}{2} \right) \nonumber \\
& \le \sum_{m=1}^\infty \P_{\nu \pi}\left( \frac{1}{m} \sum_{l=1}^m X_1(l) \le \mu_1 - \sqrt{\frac{2\sigma^2 \log(x_T)}{m}} - \frac{\Delta}{2} \right) \label{unionbound} \\
& \le \sum_{m=1}^\infty \exp\left( -\frac{m}{2 \sigma_0^2} \left( \sqrt{\frac{2\sigma^2 \log(x_T)}{m}} + \frac{\Delta}{2} \right)^2\right) \label{ldbound} \\
& = x_T^{-\sigma^2/\sigma_0^2} \cdot \sum_{m=1}^\infty \exp\left( -\frac{\sqrt{m \sigma^2 \log(x_T)}\Delta}{\sqrt{2}\sigma_0^2} - \frac{m \Delta^2}{8 \sigma_0^2} \right) \nonumber \\
& \le x_T^{-\sigma^2/\sigma_0^2} \cdot \sum_{m=1}^\infty \exp\left( -\frac{\sqrt{m} \sigma \Delta}{\sqrt{2}\sigma_0^2} - \frac{m \Delta^2}{8 \sigma_0^2} \right) \qquad \quad (\text{for } T \ge 16), \label{term1bound}
\end{align}
where to obtain (\ref{unionbound}), we have used a union bound over all possible values of $N_1(t)$, $t \ge 1$, and (\ref{ldbound}) follows from a Chernoff bound.

For the term in (\ref{term2}), we have for sufficiently large $T$,
\begin{align}
\sqrt{\frac{2\sigma^2 \log(T)}{x_T}} < \frac{\Delta}{4}. \nonumber
\end{align}
So, for sufficiently large $T$,
\begin{align}
(\ref{term2}) & \le \P_{\nu \pi}\left( \frac{1}{x_T} \sum_{t=1}^{x_T} X_2(t) > \mu_2 + \frac{\Delta}{4} \right) \nonumber \\
& \le \exp\left( - x_T \cdot \frac{\Delta^2}{32\sigma_0^2} \right), \label{term2bound}
\end{align}
where (\ref{term2bound}) follows from a Chernoff bound.

Putting together (\ref{term1}), (\ref{term1bound}) and (\ref{term2}), (\ref{term2bound}), we have established the desired result:
\begin{align}
\limsup_{T \to \infty} \frac{\log \P_{\nu \pi}\left( N_2(T) > x_T \right)}{\log(x_T)} \le -\frac{\sigma^2}{\sigma_0^2}. \nonumber
\end{align}
\halmos
\endproof

\section{Proofs for Section \ref{lowerbound}} \label{lowerboundproofs}

\proof{Proof of Lemma \ref{stronglaw}.}

This proof is an extension and simplification of Propositions 7-8 of \cite{cowan_etal2019}.

We restrict our attention to sample paths $\omega$ belonging to
\begin{align}
    \left\{ \omega : \lim_{n \to \infty} \frac{1}{n} \sum_{t=1}^n X_i(t) = \mu_i, \; i \in [K] \right\}. \label{ergodicity}
\end{align}
Without loss of generality, suppose that arm $1$ is the unique optimal arm, i.e., $\mu_1 > \max_{i \ge 2} \mu_i$.
The UCB index for arm $i$ at time $t+1$ is:
\begin{align}
U_i(t) = \sup \left\{z \in \mathcal{I}_P : d_P(\widehat{\mu}_i(t),z) \le \frac{f(t)}{N_i(t)} \right\}, \label{klucbef}
\end{align}
where, as defined previously, $\widehat{\mu}_i(t) = \frac{1}{N_i(t)} \sum_{s=1}^{N_i(t)} X_i(s)$.
As discussed in the proof of Theorem \ref{generalupperbound}, $f(t)$ is a design choice.
For KL-UCB, choices include $f(t) = \log(t)$, $f(t) = \log(t) + 3 \log\log(t)$ and $f(t) = \log(1+t\log^2(t))$.
We will leave the particular form for $f(t)$ unspecified in developing this proof.
This proof holds for any regularly varying and increasing function $f : (1,\infty) \to (0,\infty)$ satisfying $\lim_{t \to \infty} f(t) = \infty$ and $f(t) = o(t)$.

We begin with the upper bound part of the proof.
Consider sub-optimal arm $i \ge 2$, and let $\delta \in (0,(\mu_1 - \mu_i)/2)$.
We have
\begin{align}
    N_i(T) & = 1 + \sum_{t=K}^{T-1} \indic{A(t+1) = i, \; U_i(t) \ge \mu_1 - \delta, \; \widehat{\mu}_i(t) \le \mu_i + \delta } \label{sum1} \\
    & \quad \;\;\; + \sum_{t=K}^{T-1} \indic{A(t+1) = i, \; U_i(t) \ge \mu_1 - \delta, \; \widehat{\mu}_i(t) > \mu_i + \delta } \label{sum2} \\ 
    & \quad \;\;\; + \sum_{t=K}^{T-1} \indic{A(t+1) = i, \; U_i(t) < \mu_1 - \delta }, \label{sum3}
\end{align}
where $A(t)$ is the arm played by the algorithm at time $t$.

The first sum is upper bounded via:
\begin{align}
    (\ref{sum1}) & \le \sum_{t=K}^{T-1} \indic{A(t+1) = i, \; d_P(\mu_i + \delta, \mu_1 - \delta) \le \frac{f(t)}{N_i(t)} } \label{sum4} \\ 
    & \le \sum_{t=K}^{T-1} \indic{A(t+1) = i, \; N_i(t) \le \frac{f(T)}{d_P(\mu_i + \delta, \mu_1 - \delta)} } \nonumber \\
    & \le \frac{f(T)}{d_P(\mu_i + \delta, \mu_1 - \delta)} + 1. \label{sum5}
\end{align}
The bound in (\ref{sum4}) holds due to the events $U_i(t) \ge \mu_1 - \delta$ and $\widehat{\mu}_i(t) \le \mu_i + \delta$ and the definition of the index in (\ref{klucbef}).

The second sum is upper bounded via:
\begin{align}
    (\ref{sum2}) \le \sum_{t=K}^{\infty} \indic{A(t+1) = i, \; \widehat{\mu}_i(t) > \mu_i + \delta }. \label{sum6}
\end{align}
On sample paths in (\ref{ergodicity}), the indicators on the right side of (\ref{sum6}) can equal $1$ for only finitely many $t$.
(For each $1$ in the sum, arm $i$ is played an additional time, and an additional sample is incorporated into $\widehat{\mu}_i(t)$.)

The third sum is upper bounded via:
\begin{align}
    (\ref{sum3}) & \le \sum_{t=K}^{\infty} \indic{A(t+1) = i, \; U_1(t) \le U_i(t) < \mu_1 - \delta } \nonumber \\
    & \le \sum_{t=K}^{\infty} \indic{ U_1(t) < \mu_1 - \delta }. \label{sum7}
\end{align}
On sample paths in (\ref{ergodicity}), the indicators on the right side of (\ref{sum7}) can equal $1$ only for finitely many $t$.
(As $t \to \infty$, either $N_1(t)$ increases to infinity or remains bounded uniformly in $t$.
In the first case, $\widehat{\mu}_1(t) \to \mu_1$, and so for $t$ sufficiently large, $U_1(t) \ge \widehat{\mu}_1(t) > \mu_1 - \delta/2$.
In the second case, $f(t)$ in (\ref{klucbef}) increases without bound, and so $U_1(t)$ also increases without bound, with $U_1(t) > \mu_1$ for all $t$ sufficiently large.)

Putting together (\ref{sum5})-(\ref{sum7}), and sending $T \to \infty$ followed by $\delta \downarrow 0$, we have for each sub-optimal arm $i \ge 2$,
\begin{align}
    \limsup_{T \to \infty} \frac{N_i(T)}{f(T)} \le \frac{1}{d_P(\mu_i,\mu_1)}. \label{suboptimallimsup}
\end{align}
Therefore, for the optimal arm $1$,
\begin{align}
    \lim_{T \to \infty} \frac{N_1(T)}{T} = 1, \label{optimalconvergence}
\end{align}
which, by the form of the index in (\ref{klucbef}), then implies:
\begin{align}
    \lim_{t \to \infty} U_1(t) = \mu_1. \label{optimallim}
\end{align}
Then, (\ref{optimalconvergence}) and (\ref{optimallim}) imply that for each sub-optimal arm $i \ge 2$,
\begin{align}
    \lim_{T \to \infty} N_i(T) = \infty. \label{infiniteplays}
\end{align}
(If (\ref{infiniteplays}) is not true for some sub-optimal arm $j$, then since the term $f(t)$ grows without bound in the index (\ref{klucbef}), we would eventually have $U_j(t) > \mu_1 + \epsilon > U_1(t)$ for some $\epsilon > 0$ and all $t$ sufficiently large.
This would prevent arm $1$ from being played for all $t$ sufficiently large, thereby contradicting (\ref{optimalconvergence}).)

We now develop the lower bound parts of the proof.
As defined previously, for any positive integer $m$, we use $\tau_1(m)$ to denote the time of the $m$-th play of arm $1$. 
So, for each sub-optimal arm $i \ge 2$,
\begin{align}
    U_1(\tau_1(m) - 1) > U_i(\tau_1(m) - 1). \label{taum}
\end{align}
Let $\delta > 0$. 
We have for $m$ sufficiently large,
\begin{align}
    \max_{t \in [\tau_1(m),\tau_1(m+1)]} \frac{f(t)}{N_i(t)} & \le \frac{f(\tau_1(m+1))}{N_i(\tau_1(m)-1)} \nonumber \\
    & = \frac{f(\tau_1(m+1))}{f(\tau_1(m)-1)} \frac{f(\tau_1(m)-1)}{N_i(\tau_1(m)-1)} \nonumber \\
    & \le (1+\delta) \frac{f(\tau_1(m)-1)}{N_i(\tau_1(m)-1)} \label{loglog} \\ 
    & \le (1+\delta) d_P(\mu_i-\delta,U_i(\tau_1(m)-1)) \label{d1} \\
    & \le (1+\delta) d_P(\mu_i-\delta,U_1(\tau_1(m)-1)) \label{d2} \\
    & \le (1+\delta) d_P(\mu_i-\delta,\mu_1 + \delta). \label{d3}
\end{align}
Note that (\ref{loglog}) is due to (\ref{optimalconvergence}), (\ref{d1}) is due to $\lim_{t \to  \infty} \widehat{\mu}_i(t) = \mu_i$ for each sub-optimal arm $i \ge 2$ and the form of the index in (\ref{klucbef}), (\ref{d2}) is due to (\ref{taum}), and (\ref{d3}) is due to (\ref{optimallim}).
From (\ref{d3}), we obtain
\begin{align*}
    \liminf_{T \to \infty} \frac{N_i(T)}{f(T)} \ge \frac{1}{d_P(\mu_i,\mu_1)}, 
\end{align*}
which together with (\ref{suboptimallimsup}), completes the proof.
\halmos
\endproof

\section{Proofs for Section \ref{simple}} \label{simpleproofs}

\proof{Proof of (\ref{psitail1}) in Proposition \ref{prop5}.}
Without loss of generality, suppose that $\mu_1 > \mu_2 > \dots > \mu_K$ (i.e., $r(i) = i$ for all $1 \le i \le K$) for the environment $\nu$.
To simplify notation, for the right side of (\ref{psitail1}), we use the shorthand:
\begin{align*}
    c_i(\nu) := \sum_{j=1}^{i-1} \inf_{z \in \mathcal{I}_P \, : \, z < \mu_i} \frac{d_P(z,\mu_j)}{d_P(z,\mu_i)}.
\end{align*}
Recall that we consider any (fixed) $\gamma \in (0,(1/\lambda) - 1)$. 
Let $B_\gamma(T) = [f^{1+\gamma}(T),(1-\gamma)T]$.
To establish (\ref{psitail1}), we will show for any sub-optimal arm $i$:
\begin{align}
    \liminf_{T \to \infty} \inf_{x \in B_\gamma(T)} \frac{\log \P_{\nu \pi}(N_i(T) > x)}{f(x)} \ge - c_i(\nu), \label{result11} \\
    \limsup_{T \to \infty} \sup_{x \in B_\gamma(T)} \frac{\log \P_{\nu \pi}(N_i(T) > x)}{f(x)} \le - c_i(\nu). \label{result22}
\end{align}
This is analogous to the approach taken in Lemma \ref{lem2}, where parts (i) and (ii) were used to establish part (iii). 
However, here we must handle any regularly varying function $f$ satisfying $\liminf_{t \to \infty} f(t)/\log(t) \ge 1$ and $f(t) = o(t^\lambda)$ for some $\lambda \in (0,1)$ (instead of just $x \mapsto \log(x)$, as in Lemma \ref{lem2}). \vspace{2mm} \\
\noindent \underline{Proof of (\ref{result11})} \\
Let $\gamma' \in (0,\gamma)$.
We follow the proof of Theorem \ref{thm1} (in Appendix \ref{thm1_proof}) with three changes.
First, we use $\gamma'$ instead of $\gamma$.
Second, we replace $\log(T)$ by $f(T)$ everywhere.
Third, in place of the WLLN provided by Lemma \ref{lemma2}, we use the WLLN derived from the SLLN of Lemma \ref{stronglaw} (see Appendix \ref{lowerboundproofs}), which continues to hold for regularly varying and strictly increasing functions $f$ satisfying $\liminf_{t \to \infty} f(t)/\log(t) \ge 1$ and $f(t) = o(t^\lambda)$ for some $\lambda \in (0,1)$.
Running through the proof of Theorem \ref{thm1} with these three changes, instead of obtaining (\ref{optimizedklratio}), we obtain:
\begin{align*}
\liminf_{T \to \infty} \frac{\log \P_{\nu \pi}(N_i(T) > (1-\gamma')T)}{f(T)} 
& \ge - c_i(\nu).
\end{align*}
Since $f$ is regularly varying, there exists $a \in (0,1)$ such that
\begin{align*}
    \lim_{T \to \infty} \frac{f((1-\gamma')T)}{f(T)} = (1-\gamma')^a.
\end{align*}
Therefore, 
\begin{align}
\liminf_{T \to \infty} \frac{\log \P_{\nu \pi}(N_i(T) > (1-\gamma')T)}{f((1-\gamma')T)} \ge - c_i(\nu) (1-\gamma')^{-a}. \label{result33}
\end{align}

Since $t \mapsto N_i(t)$ is non-decreasing with $\P_{\nu \pi}$-probability one, we have for sufficiently large $T$ and all $x \in [f^{1+\gamma}(T),(1-\gamma')T]$,
\begin{align*}
    \P_{\nu \pi}(N_i(T) > x) \ge \P_{\nu \pi}(N_i(\lceil x/(1-\gamma') \rceil) > x).
\end{align*}
So, for any $\epsilon > 0$, we have for sufficiently large $T$,
\begin{align}
    \frac{\log \P_{\nu \pi}(N_i(T) > x)}{f(x)} & \ge \frac{\log \P_{\nu \pi}(N_i(\lceil x/(1-\gamma') \rceil) > x)}{f(x)} \label{result_44} \\
    & \ge -c_i(\nu) (1-\gamma')^{-a} (1+\epsilon), \label{result44}
\end{align}
uniformly for all $x \in B_\gamma(T) = [f^{1+\gamma}(T),(1-\gamma)T] \subset [f^{1+\gamma}(T),(1-\gamma')T]$, where (\ref{result44}) follows from the convergence result in (\ref{result33}).
Then, (\ref{result11}) is established by taking the infimum over $x \in B_\gamma(T)$ on the left side of (\ref{result_44}), sending $T \to \infty$, and then $\epsilon \downarrow 0$ and $\gamma' \downarrow 0$. \vspace{2mm} \\
\noindent \underline{Proof of (\ref{result22})} \\
Let $g(t) = f^{1+\gamma}(t)$.
Using the upper bound part of the proof of Theorem \ref{generalupperbound} (in Appendix \ref{generalupperbound_proof}), we obtain:
\begin{align}
\limsup_{T \to \infty} \frac{\log \P_{\nu \pi}\left( N_i(T) > g(T) \right)}{f(g(T))} \le - c_i(\nu). \label{result55}
\end{align}
The function $g$ is strictly increasing, so it has an inverse $g^{-1}$ (defined on the range of $g$), which is also strictly increasing.
Since $t \mapsto N_i(t)$ is non-decreasing with $\P_{\nu \pi}$-probability one, we have for sufficiently large $T$ and all $x \in B_\gamma(T) = [g(T),(1-\gamma)T]$,
\begin{align*}
    \P_{\nu \pi}(N_i(\lfloor g^{-1}(x) \rfloor) > x) \ge \P_{\nu \pi}(N_i(T) > x).
\end{align*}
Thus, for any $\epsilon \in (0,1)$, we have for sufficiently large $T$,
\begin{align}
    -c_i(\nu)(1-\epsilon) & \ge \frac{\log \P_{\nu \pi}(N_i(\lfloor g^{-1}(x) \rfloor) > x)}{f(x)} \label{result66} \\
    & \ge \frac{\log \P_{\nu \pi}(N_i(T) > x)}{f(x)}, \label{result_66}
\end{align}
uniformly for all $x \in B_\gamma(T)$, where (\ref{result66}) follows from the convergence result in (\ref{result55}).
Then, (\ref{result22}) is established by taking the supremum over $x \in B_\gamma(T)$ on the right side of (\ref{result_66}), sending $T \to \infty$, and then $\epsilon \downarrow 0$.
\halmos
\endproof

\vspace{4mm}

\proof{Proof of (\ref{asymptotic_expected_regret}) in Proposition \ref{prop5}.}
We adapt the proof of Theorem 10.6 (page 116) of \cite{lattimore_etal2020}, with only a few slight modifications.
Recall from the proof of Theorem \ref{generalupperbound} (in Appendix \ref{generalupperbound_proof}) that $\tau_i(n)$ denotes the time of the $n$-th play of arm $i$.
Without loss of generality, suppose arm 1 is optimal for the environment $\nu$ under consideration.
Let $i \ge 2$, and let $\epsilon_1, \epsilon_2 \in (0,(\mu_1-\mu_i)/2)$.
Define:
\begin{align}
    \xi_1 & = \min\left\{ t : \max_{1 \le s \le T} \left( \underline{d}_P(\widehat{\mu}_1(\tau_1(s)), \mu_1 - \epsilon_2) - \frac{f(t)}{s} \right) \le 0 \right\} \label{final_xi} \\
    \kappa_i & = \sum_{s=1}^T \indic{d_P(\widehat{\mu}_i(\tau_i(s)), \mu_1 - \epsilon_2) \le \frac{f(T)}{s}}, \label{final_kappa}
\end{align}
where we use the notation $\underline{d}_P(x,y) := d_P(x,y) \indic{x \le y}$.
Then,
\begin{align}
    \E_{\nu \pi}[N_i(T)] & = \E_{\nu \pi}\left[ \sum_{t=1}^T \indic{A(t) = i} \right] \nonumber \\
    & \le \E_{\nu \pi}[\xi_1] + \E_{\nu \pi}\left[ \sum_{t = \xi_1 + 1}^T \indic{A(t) = i} \right] \nonumber \\
    & \le \E_{\nu \pi}[\xi_1] + \E_{\nu \pi}\left[ \sum_{t=1}^T \indic{A(t) = i, \, d_P(\widehat{\mu}_i(t-1), \mu_1-\epsilon_2) \le \frac{f(t)}{N_i(t-1)}} \right] \label{final_bound_1} \\
    & \le \E_{\nu \pi}[\xi_1] + \E_{\nu \pi}[\kappa_i] \label{final_bound_2} \\
    & \le C(\mu_1,\epsilon_2) + \frac{f(T)}{d_P(\mu_i + \epsilon_1, \mu_1 - \epsilon_2)} + C'(\mu_i,\epsilon_1). \label{final_bound_3}
\end{align}
Note that (\ref{final_bound_1}) follows from the observation that for time periods after $\xi_1$, the UCB index for arm $1$ must be at least as large as $\mu_1-\epsilon_2$, and so if arm $i$ is played, then the UCB index for arm $i$ must also be at least as large as $\mu_1-\epsilon_2$.
Also, (\ref{final_bound_2}) follows directly from the definition of $\kappa_i$ in (\ref{final_kappa}).
And (\ref{final_bound_3}) follows from Lemmas \ref{lemma_xi} and \ref{lemma_kappa}.
Therefore,
\begin{align}
    \limsup_{T \to \infty} \frac{\E_{\nu \pi}[N_i(T)]}{f(T)} \le \frac{1}{d_P(\mu_i,\mu_1)}. \nonumber
\end{align}
The matching asymptotic lower bound
\begin{align}
    \liminf_{T \to \infty} \frac{\E_{\nu \pi}[N_i(T)]}{f(T)} \ge \frac{1}{d_P(\mu_i,\mu_1)} \nonumber
\end{align}
is directly obtained from the WLLN derived from the SLLN of Lemma \ref{stronglaw} (see Appendix \ref{lowerboundproofs}), together with Markov's inequality.
\halmos
\endproof

\vspace{4mm}

Lemmas \ref{lemma_xi} and \ref{lemma_kappa} are directly from Chapter 10 of \cite{lattimore_etal2020}, and are included here for convenience in referencing.
Lemma \ref{lemma2_klucb} is directly from the Appendix of \cite{cappe_etal2013}, and is included here also for convenience in referencing.

\begin{lemma}[Lemma 10.7 from \cite{lattimore_etal2020}] \label{lemma_xi}
    Let $X_1,\dots,X_T$ be independent random variables from $P^\mu$, and let $\overline{\mu}_s = s^{-1} \sum_{j=1}^s X_j$.
    Let $\epsilon > 0$ and define
    \begin{align}
        \xi = \min\left\{ t : \max_{1 \le s \le T} \left( \underline{d}_P(\overline{\mu}_s, \mu - \epsilon) - \frac{f(t)}{s} \right) \le 0 \right\}, \label{xi_def}
    \end{align}
    where $f(t) \ge \log(1+t\log^2(t))$ for sufficiently large $t$.
    Then, $\E[\xi] \le C(\mu,\epsilon)$, for some finite constant $C(\mu,\epsilon)$ depending on $\mu$ and $\epsilon$.
\end{lemma}

\proof{Proof of Lemma \ref{lemma_xi}.}
Define $\sigma^2_P(\mu_1,\mu_2) := \max_{z \in [l,u]} \textnormal{Var}_{X \sim P^{z}}(X)$, with $l = \min(\mu_1,\mu_2)$ and $u = \max(\mu_1,\mu_2)$.
Then, we have
\begin{align}
    \P(\xi > t) & \le \P\left(\exists 1 \le s \le T : \underline{d}_P(\overline{\mu}_s,\mu-\epsilon) > \frac{f(t)}{s}\right) \nonumber \\
    & \le \sum_{s=1}^T \P\left(\underline{d}_P(\overline{\mu}_s,\mu-\epsilon) > \frac{f(t)}{s}\right) \nonumber \\
    & = \sum_{s=1}^T \P\left(d_P(\overline{\mu}_s,\mu-\epsilon) > \frac{f(t)}{s}, \; \overline{\mu}_s < \mu-\epsilon\right) \nonumber \\
    & \le \sum_{s=1}^T \P\left(d_P(\overline{\mu}_s,\mu) - d_P(\mu-\epsilon,\mu) > \frac{f(t)}{s}, \; \overline{\mu}_s < \mu-\epsilon\right) \label{lemma_xi_pythag} \\
    & \le \sum_{s=1}^T \P\left(d_P(\overline{\mu}_s,\mu) > \frac{f(t)}{s} + \frac{\epsilon^2}{2\sigma_P^2(\mu-\epsilon,\mu)}, \; \overline{\mu}_s < \mu\right) \label{lemma_xi_pinsker} \\
    & \le \sum_{s=1}^T \exp\left(-s\left(\frac{f(t)}{s} + \frac{\epsilon^2}{2\sigma_P^2(\mu-\epsilon,\mu)}\right)\right) \label{lemma_xi_chernoff} \\
    & \le (1+t \log^2(t))^{-1} \sum_{s=1}^\infty \exp\left(-s\frac{\epsilon^2}{2\sigma_P^2(\mu-\epsilon,\mu)}\right) \label{lemma_xi_growth} \\
    & = (1+t \log^2(t))^{-1} \left(\exp\left(\frac{\epsilon^2}{2\sigma_P^2(\mu-\epsilon,\mu)}\right) - 1 \right)^{-1}. \nonumber
\end{align}
Note that (\ref{lemma_xi_pythag}) follows from Lemma \ref{lemma_kl_pythag}, (\ref{lemma_xi_pinsker}) follows from Lemma \ref{lemma2_klucb}, (\ref{lemma_xi_chernoff}) follows from a Chernoff bound.
For sufficiently large $t$, say $t \ge L$ for some $L > 0$, we have $f(t) \ge \log(1+t\log^2(t))$, which then yields (\ref{lemma_xi_growth}).
Then,
\begin{align}
    \E[\xi] & = \int_0^\infty \P(\xi > t) dt \nonumber \\
    & \le L + \left(\exp\left(\frac{\epsilon^2}{2\sigma_P^2(\mu-\epsilon,\mu)}\right) - 1 \right)^{-1} \int_L^\infty (1+t \log^2(t))^{-1} dt \label{convergent_integral} \\
    & =: C(\mu,\epsilon), \nonumber
\end{align}
where the integral on the right side of (\ref{convergent_integral}) is finite, and thus $C(\mu,\epsilon)$ (defined to be equal to the right side of (\ref{convergent_integral})) is a finite constant depending on $\mu$ and $\epsilon$.
\halmos
\endproof

\begin{lemma} [Lemma 10.8 from \cite{lattimore_etal2020}] \label{lemma_kappa}
    Let $X_1,\dots,X_T$ be independent random variables from $P^\mu$, and let $\overline{\mu}_s = s^{-1} \sum_{j=1}^s X_j$.
    Let $\Delta > 0$ and define
    \begin{align}
        \kappa = \sum_{s=1}^T \indic{d_P(\overline{\mu}_s, \mu + \Delta) \le \frac{f(T)}{s}}. \label{kappa_def}
    \end{align}
    Then, for any $\epsilon \in (0,\Delta)$,
    \begin{align}
        \E[\kappa] \le \frac{f(T)}{d_P(\mu+\epsilon,\mu+\Delta)} + C'(\mu,\epsilon),
    \end{align}
    where $C'(\mu,\epsilon)$ is a finite constant depending on $\mu$ and $\epsilon$.
\end{lemma}

\proof{Proof of Lemma \ref{lemma_kappa}.}
Let $\epsilon \in (0,\Delta)$ and $a = f(T)/d_P(\mu+\epsilon,\mu+\Delta)$.
Then,
\begin{align}
    \E[\kappa] & = \sum_{s=1}^T \P\left(d_P(\overline{\mu}_s,\mu+\Delta) \le \frac{f(T)}{s}\right) \nonumber \\
    & \le \sum_{s=1}^T \P\left(\overline{\mu}_s \ge \mu+\epsilon \text{ or } d_P(\mu+\epsilon,\mu+\Delta) \le \frac{f(T)}{s}\right) \nonumber \\
    & \le a + \sum_{s=\lceil a \rceil}^T \P(\overline{\mu}_s \ge \mu + \epsilon) \nonumber \\
    & \le a + \sum_{s=1}^T \exp(-sd_P(\mu+\epsilon,\mu)) \label{lemma_kappa_chernoff} \\
    & \le a + \sum_{s=1}^\infty \exp\left(-s \frac{\epsilon^2}{2\sigma_P^2(\mu+\epsilon,\mu)}\right), \label{lemma_kappa_pinsker} \\
    & = a + \left(\exp\left(\frac{\epsilon^2}{2\sigma_P^2(\mu+\epsilon,\mu)}\right) - 1 \right)^{-1}, \nonumber
\end{align}
where (\ref{lemma_kappa_chernoff}) follows from a Chernoff bound, and (\ref{lemma_kappa_pinsker}) follows from Lemma \ref{lemma2_klucb}.
Then, defining $C'(\mu,\epsilon) := \left(\exp(\epsilon^2/(2\sigma_P^2(\mu+\epsilon,\mu))) - 1 \right)^{-1}$, we see that $C'(\mu,\epsilon)$ is a finite constant depending on $\mu$ and $\epsilon$.
\halmos
\endproof

\begin{lemma} [Lemma 2 from Appendix of \cite{cappe_etal2013}] \label{lemma2_klucb}
    For a base distribution $P$, let $\mu_1,\mu_2 \in \mathcal{I}_P$. 
    Then,
    \begin{align}
        d_P(\mu_1,\mu_2) \ge \frac{(\mu_1-\mu_2)^2}{2 \sigma^2_P(\mu_1,\mu_2)},
    \end{align}
    where $\sigma^2_P(\mu_1,\mu_2) = \max_{z \in [l,u]} \textnormal{Var}_{X \sim P^{z}}(X)$, with $l = \min(\mu_1,\mu_2)$ and $u = \max(\mu_1,\mu_2)$.
\end{lemma}

\proof{Proof of Lemma \ref{lemma2_klucb}.}
See pages 7-8 of the Appendix of \cite{cappe_etal2013}.
\halmos
\endproof

\begin{lemma} [Triangle Inequality for Information Projection] \label{lemma_kl_pythag}
    For a base distribution $P$, let $\epsilon > 0$ and $\mu_1,\mu_2 \in \mathcal{I}_P$ such that $\mu_1 \le \mu_2 - \epsilon < \mu_2$.
    Then,
    \begin{align}
        d_P(\mu_1,\mu_2-\epsilon) + d_P(\mu_2-\epsilon,\mu_2) \le d_P(\mu_1,\mu_2). \label{kl_pythag_id_1}
    \end{align}
\end{lemma}

\proof{Proof of Lemma \ref{lemma_kl_pythag}.}
Let $\mathcal{E} = \{Q : \mu(Q) \le \mu_2 - \epsilon\}$, the set of all distributions with mean $\le \mu_2 - \epsilon$, which is a closed convex set of distributions.
By Theorem 11.6.1 (on page 367) of \cite{cover_etal2006} (alternatively, see Theorem 15.10 (on page 303) of \cite{polanskiy_etal2023}), for $Q^* = \argmin_{Q \in \mathcal{E}} D(Q \: \lVert \: P^{\mu_2})$, we have
\begin{align}
    D(P^{\mu_1} \: \lVert \: Q^*) + D(Q^* \: \lVert \: P^{\mu_2}) \le d_P(\mu_1,\mu_2). \label{kl_pythag_id_2}
\end{align}
Moreover, Theorem 15.11 (on pages 303-304) of \cite{polanskiy_etal2023} indicates that in this case, $Q^* = P^{\mu_2-\epsilon}$.
So, together with (\ref{kl_pythag_id_2}), the desired result in (\ref{kl_pythag_id_1}) is established.
\halmos
\endproof

\section{Proofs for Sections \ref{robust1}-\ref{robust2}} \label{robustnessproofs}

We first establish that distributions in $\mathcal{M}_{P,b}$ obey a Chernoff bound using the re-scaled divergence $d_P/(1+b)$ instead of their usual KL divergence. 
Recall that the enlarged family of distributions is:
\begin{align}
\mathcal{M}_{P,b} = \left\{ Q \, : \, \mu(Q) \in \mathcal{I}_P, \; \Lambda_{Q}(\theta) \le \Psi_{P,b}(\mu(Q),\theta) \;\; \forall \; \theta \in \mathbb{R} \right\}, \label{mdelta2}
\end{align}
where for any distribution $Q$ and $z \in \mathcal{I}_Q$, we define:
\begin{align}
    \Psi_{Q,b}(z,\theta) = \frac{\Lambda_{Q^z}((1+b)\theta)}{1+b}, \quad \theta \in \mathbb{R}. \label{psibz2}
\end{align}

\begin{lemma} \label{lemma_rescaled_chernoff}
    Let $Q \in \mathcal{M}_{P,b}$ and $\overline{\mu}_s = s^{-1} \sum_{j=1}^s X_j$, where $X_1,\dots,X_s$ are independent random variables from $Q$.
    Then,
    \begin{align}
        & \P\left( \overline{\mu}_s > z \right) \le \exp\left(-s \frac{d_P(z,\mu(Q))}{1+b} \right), \quad \text{for } z > \mu(Q), \label{rescaled_divergence_chernoff1} \\
        & \P\left( \overline{\mu}_s < z \right) \le \exp\left(-s \frac{d_P(z,\mu(Q))}{1+b} \right), \quad \text{for } z < \mu(Q). \label{rescaled_divergence_chernoff2}
    \end{align}
\end{lemma}

\proof{Proof of Lemma \ref{lemma_rescaled_chernoff}.}
The re-scaled divergence $d_P/(1+b)$ is directly related to the re-scaled CGF's in (\ref{psibz2}).
Specifically, $y \mapsto d_P(y,z)/(1+b)$ is the convex conjugate of $\theta \mapsto \Psi_{P,b}(z,\theta)$, as seen via:
\begin{align}
    \sup_{\theta \in \mathbb{R}} \{ \theta y - \Psi_{P,b}(z,\theta) \} = \frac{1}{1+b} \cdot \sup_{\theta \in \mathbb{R}} \left\{ \theta y - \Lambda_{P^z}(\theta) \right\} = \frac{d_P(y,z)}{1+b}. \label{psidivergence2}
\end{align}
For $Q \in \mathcal{M}_{P,b}$, using (\ref{psidivergence2}), we have
\begin{align}
    d_Q(y,\mu(Q)) & = \sup_{\theta \in \mathbb{R}} \{\theta y - \Lambda_Q(\theta)\} \nonumber \\
    & \ge \sup_{\theta \in \mathbb{R}} \{\theta y - \Psi_{P,b}(\mu(Q),\theta) \} = \frac{d_P(y,\mu(Q))}{1+b}. \nonumber
\end{align}
Therefore, for $z > \mu(Q)$
\begin{align}
    \P\left( \overline{\mu}_s > z \right) & \le \exp(-s d_Q(z,\mu(Q))) \label{usual_chernoff} \\
    & \le \exp\left(-s \frac{d_P(z,\mu(Q))}{1+b} \right), \nonumber
\end{align}
where (\ref{usual_chernoff}) follows from a Chernoff bound.
So, (\ref{rescaled_divergence_chernoff1}) is established, and (\ref{rescaled_divergence_chernoff2}) is established by the same arguments.
\halmos
\endproof

\vspace{4mm}

Next, we have the proofs for Corollary \ref{cor4} and \ref{cor5}.

\proof{Proof of Corollary \ref{cor4}.}
The above proof of (\ref{asymptotic_expected_regret}) in Proposition \ref{prop5} can be applied to the new setting in Corollary \ref{cor4}.
Recall that we now assume that $f(t) \ge (1+b) \log(1+t\log^2(t))$ for sufficiently large $t$.
(In the above proof of (\ref{asymptotic_expected_regret}), the lower bound was $\log(1+t\log^2(t))$, without the $1+b$ factor.)
In Lemma \ref{lemma_xi} and Lemma \ref{lemma_kappa}, in the definitions of $\xi$ and $\kappa$ in (\ref{xi_def}) and (\ref{kappa_def}), respectively, the following two options are equivalent upon inspection.
\begin{enumerate}
    \item Assume that $f(t) \ge (1+b) \log(1+t\log^2(t))$ for sufficiently large $t$, and keep $d_P$ unmodified.
    \item Assume that $f(t) \ge \log(1+t\log^2(t))$ for sufficiently large $t$, and replace $d_P$ by $d_P/(1+b)$.
\end{enumerate}
Instead of option 1, we use option 2.
Then, the proofs of Lemma \ref{lemma_xi} and Lemma \ref{lemma_kappa} go through exactly as before, except that now $d_P$ is replaced by $d_P/(1+b)$ in every instance.
Since the distributions of the arm rewards now belong to $\mathcal{M}_{P,b}$, the Chernoff bounds used in (\ref{lemma_xi_chernoff}) and (\ref{lemma_kappa_chernoff}) now hold with the re-scaled divergence $d_P/(1+b)$, as justified by Lemma \ref{lemma_rescaled_chernoff}.

Then, as in the above proof of (\ref{asymptotic_expected_regret}) in Proposition \ref{prop5}, we have established the asymptotic upper bound part of (\ref{psiregret2}), i.e., for sub-optimal arm $i$ in environment $\nu \in \mathcal{M}_{P,b}^K$,
\begin{align}
\limsup_{T \to \infty} \frac{\E_{\nu \pi}\left[ N_i(T) \right]}{f(T)} \le \frac{1+b}{d_P(\mu(Q_{i}),\mu_*(\nu))}, \label{equivalent_upper}
\end{align}
where $f(t) \ge \log(1+t\log^2(t))$ for sufficiently large $t$.
The matching asymptotic lower bound part
\begin{align}
    \liminf_{T \to \infty} \frac{\E_{\nu \pi}\left[ N_i(T) \right]}{f(T)} \ge \frac{1+b}{d_P(\mu(Q_{i}),\mu_*(\nu))}. \label{equivalent_lower}
\end{align}
is easily obtained from the WLLN derived from the SLLN of Lemma \ref{stronglaw} (see Appendix \ref{lowerboundproofs}), together with Markov's inequality.
Together, (\ref{equivalent_upper}) and (\ref{equivalent_lower}) yield the desired conclusion of Corollary \ref{cor4}.
\halmos
\endproof

\proof{Proof of Corollary \ref{cor5}.}
The proof of Corollary \ref{cor5} is essentially the same as that of Corollary \ref{cor4}.
The only difference is that we now must use a Chernoff bound for additive functionals of finite state space Markov chains.
Theorem 1 of \cite{moulos_etal2019} provides such a Chernoff bound that is convenient for this purpose.
(Earlier and more general results can be found in \cite{miller_1961} and \cite{kontoyiannis_etal2003}, respectively.)
Recall that in (\ref{widetildeMb}) from Section \ref{robust2}, we defined the following set of transition matrices:
\begin{align}
    \widetilde{\mathcal{M}}_{P,b} = \left\{ H \in \mathcal{S}_{\abs{S}} \, : \, \phi_H(\theta) \le \Psi_{P,b}(\phi_H'(0),\theta) \;\; \forall \; \theta \in \mathbb{R} \right\}. \nonumber
\end{align}
Using the same arguments as in Lemma \ref{lemma_rescaled_chernoff}, we can deduce that Markov chains with transition matrices $H \in \widetilde{\mathcal{M}}_{P,b}$ obey a Chernoff bound involving the re-scaled divergence $d_P/(1+b)$.
Specifically, with $\overline{\mu}_s = s^{-1} \sum_{j=1}^s X_j$ and $X_1,X_2,\dots$ evolving according to transition matrix $H$ (with any initial distribution), we have
\begin{align}
        & \P\left( \overline{\mu}_s > z \right) \le c_H \exp\left(-s \frac{d_P(z,\phi_H'(0))}{1+b} \right), \quad \text{for } z > \phi_H'(0), \nonumber \\
        & \P\left( \overline{\mu}_s < z \right) \le c_H \exp\left(-s \frac{d_P(z,\phi_H'(0))}{1+b} \right), \quad \text{for } z < \phi_H'(0), \nonumber
\end{align}
where $c_H > 0$ is a constant depending only on the transition matrix $H$ (see Theorem 1 and Proposition 1 of \cite{moulos_etal2019}).
The rest of the proof is identical to that of Corollary \ref{cor4}.
\halmos
\endproof

%
%
%

\end{APPENDICES}







\end{document}